\documentclass{article}

\usepackage{PRIMEarxiv}

\usepackage[utf8]{inputenc} 
\usepackage[T1]{fontenc}    
\usepackage{hyperref}       
\usepackage{url}            
\usepackage{booktabs}       
\usepackage{amsfonts}       
\usepackage{nicefrac}       
\usepackage{microtype}      
\usepackage{lipsum}
\usepackage{fancyhdr}       
\usepackage{graphicx}       
\usepackage{caption,subcaption}
\usepackage{amssymb}
\usepackage{amsmath}

\usepackage{pdflscape}
\usepackage{caption,subcaption}
\usepackage{hyperref}
\usepackage{siunitx}
\usepackage{rotating}
\usepackage{anyfontsize}
\usepackage{tabulary}
\usepackage{tabularx}
\usepackage{tikz}
\usepackage{changepage}
\usepackage{geometry}
\usepackage{graphicx} 
\usetikzlibrary{shapes.geometric, arrows}
\tikzstyle{startstop} = [rectangle, rounded corners, 
minimum width=3cm, 
minimum height=1cm,
text width=3cm, 
text centered, 
draw=white, 
fill=blue!30]
\tikzstyle{startstop1} = [rectangle, rounded corners, 
minimum width=3cm, 
minimum height=2cm,
text width=3cm, 
text centered, 
draw=white, 
fill=blue!30]
\tikzstyle{io} = [trapezium, 
trapezium stretches=true, 
trapezium left angle=70, 
trapezium right angle=110, 
minimum width=3cm, 
minimum height=1cm, text centered, 
draw=black, fill=blue!30]

\tikzstyle{process} = [rectangle, 
minimum width=3cm, 
minimum height=1cm, 
text centered, 
text width=3cm, 
draw=black, 
fill=orange!30]

\tikzstyle{decision} = [diamond, 
minimum width=3cm, 
minimum height=1cm, 
text centered, 
draw=black, 
fill=green!30]
\tikzstyle{arrow} = [thick,->,>=stealth]

\usepackage{amssymb}
\usepackage{amsmath}

\usepackage{pifont}
\newcommand{\xmark}{\ding{55}}
\graphicspath{{media/}}     

\title{Recent Advancements in Deep Learning Applications and Methods for Autonomous Navigation: A Comprehensive Review
}

\author{
  Arman Asgharpoor Golroudbari \\
  Department of Aerospace, \\Faculty of New Sciences \& Technologies, \\University of Tehran,\\
  Tehran, Iran \\
  a.asgharpoor@ut.ac.ir
  \And
  Mohammad Hossein Sabour \\
  Department of Aerospace, \\Faculty of New Sciences \& Technologies, \\ University of Tehran, \\
  Tehran, Iran \\
  sabourmh@ut.ac.ir
}

\begin{document}
\maketitle

\begin{abstract}
This review article is an attempt to survey all recent AI based techniques used to deal with major functions in This review paper presents a comprehensive overview of end-to-end deep learning frameworks used in the context of autonomous navigation, including obstacle detection, scene perception, path planning, and control. The paper aims to bridge the gap between autonomous navigation and deep learning by analyzing recent research studies and evaluating the implementation and testing of deep learning methods. It emphasizes the importance of navigation for mobile robots, autonomous vehicles, and unmanned aerial vehicles, while also acknowledging the challenges due to environmental complexity, uncertainty, obstacles, dynamic environments, and the need to plan paths for multiple agents. The review highlights the rapid growth of deep learning in engineering data science and its development of innovative navigation methods. It discusses recent interdisciplinary work related to this field and provides a brief perspective on the limitations, challenges, and potential areas of growth for deep learning methods in autonomous navigation. Finally, the paper summarizes the findings and practices at different stages, correlating existing and future methods, their applicability, scalability, and limitations. The review provides a valuable resource for researchers and practitioners working in the field of autonomous navigation and deep learning.
\end{abstract}

\section{Introduction}
\label{sec:Introduction}
Autonomous navigation is a critical component of robotics that has transformed numerous application domains, such as medical, industrial, space, and agricultural. By equipping robots with the ability to navigate autonomously, they can efficiently and securely move through dynamic environments without human intervention, expanding their versatility and functionality. To enhance the performance of autonomous navigation systems, researchers have been pushing the technology to its limits, employing state-of-the-art techniques and methodologies.

Given the expansive and ever-evolving nature of the literature surrounding autonomous navigation, it is imperative to conduct regular literature surveys in order to remain abreast of the latest advancements. Therefore, the primary objective of this review is to provide a comprehensive and in-depth overview of the current state-of-the-art in autonomous navigation, with a focus on catering to both experienced researchers and novices in the field. Additionally, a terminology section is included to provide clarity and understanding of the technical vocabulary utilized throughout the article.

\begin{itemize}
	\item \textbf{Autonomous Navigation}: Navigate and move through an environment without human intervention.
		
	\item \textbf{Perception}: Sense and understand the surroundings through the use of sensors.

	\item \textbf{Localization}: Determining the position within a known map or environment.

	\item \textbf{Mapping}: Creating a map of an environment using sensor data and other inputs.

	\item \textbf{Simultaneous Localization and Mapping (SLAM)}: Creating a map of an unknown environment while simultaneously localizing the robot or autonomous system within that environment.

	\item \textbf{Control}: Regulating the motion to follow a desired trajectory and achieve a specific task.
	
	\item \textbf{Obstacle Avoidance}: Navigate around obstacles in the path.

	\item \textbf{Collision Avoidance}: Using sensors and algorithms to detect potential collisions and take action to avoid them.
	
	\item \textbf{Path Planning}: Determining a safe and efficient path for an autonomous system to follow.

	\item \textbf{Motion Planning}: Determining the trajectory to reach a goal while avoiding obstacles and adhering to other constraints.

	\item \textbf{Sensor Fusion}: Combining data from multiple sensors to obtain a more accurate and comprehensive understanding of the environment.

	\item \textbf{Odometry}: Using sensory data to estimate the position and orientation by analyzing the movement over time.
	
	\item \textbf{Dead Reckoning}: Estimates the current position by using its previous position and velocity.
	
\end{itemize}

Navigation is a critical task for systems that operate in dynamic environments, such as robots, autonomous systems, and unmanned aerial vehicles \cite{britting2010inertial}. It requires the ability to perceive the surroundings, plan a path, execute it, and adapt as needed, all while avoiding obstacles and collisions to ensure safe, efficient, and accurate travel. Recent developments in deep learning have made navigation more reliable, effective, and efficient, enabling its use in a wide range of applications, including transportation, search and rescue, and delivery. Autonomous systems are becoming increasingly prevalent and can determine their actions based on the current situation. These systems have numerous uses, such as self-driving cars \cite{ni2022improved}, drones \cite{lee2021flying}, and search-and-rescue robots \cite{niroui2019deep}.

Autonomous systems fall into two broad categories \cite{muscettola2002idea}: reactive and deliberative. \textbf{Reactive systems}, also known as behavior-based systems, are designed to respond to the environment using predefined rules. These systems are typically used in robotics applications, where the environment is relatively stable, and the system has a specific task, such as pick and place. On the other hand, \textbf{deliberative systems} are designed to plan and execute a path to a destination. These systems are commonly used in transportation, where the environment is complex, and the system must navigate efficiently and safely.

The deliberative systems can be further classified into two categories \cite{haith2013model}: (1) model-based and (2) model-free systems. Model-based systems use a mathematical model of the environment to plan a path, such as using dynamic programming or graph search algorithms \cite{wolek2017model}. Model-free systems, also known as model-agnostic systems, do not rely on a model of the environment to plan a path
\cite{kontoudis2019kinodynamic}. Instead, these systems use techniques such as Reinforcement Learning \cite{zhu2021deep} or Apprenticeship Learning \cite{abbeel2008apprenticeship} to learn the optimal policy for navigation. The choice of autonomous system, whether reactive or deliberative and model-based or model-free, depends on the specific requirements and constraints of the task.

Autonomous systems require effective and successful navigation to operate in dynamic environments without human intervention and guidance. This involves integrating various technologies, including sensors, actuators, and control systems. To improve the accuracy, efficiency, and robustness of navigation algorithms, deep learning has been applied to various navigation tasks such as perception and planning. Deep learning's ability to learn complex representations from vast amounts of data is well-suited for these tasks, such as image analysis and integrating multiple sources of information, such as speech and text. However, deep learning-based navigation systems must overcome challenges such as limited data, reliability, ethical concerns, such as privacy and bias. Techniques such as transfer learning, multi-modal fusion, model uncertainty estimation, and safety-critical architectures can address these challenges. Additionally, differential privacy and fairness-aware machine learning techniques can address privacy and bias concerns, respectively.

The application of deep learning in navigation has seen a rise in recent times, as evidenced by numerous studies and surveys (refer to Tables \ref{tab:recent_articles} and \ref{tab:recent_surveys}). Although deep learning holds great promise in enhancing navigation systems, it is crucial to tackle the challenges and ethical considerations that come with its usage. Furthermore, there is a need for further exploration on how deep learning techniques can be integrated with conventional navigation methods. 

Numerous surveys have been conducted on the applications of deep learning in various navigation domains, including urban navigation\cite{el2021systematic}, visual navigation \cite{wang2022applications, o2018deep}, reinforcement learning \cite{almahamid2022autonomous, zhu2021deep}, obstacle detection \cite{badrloo2022image}, and spacecraft navigation \cite{song2022deep, turan2022autonomous}. However, there is a lack of comprehensive surveys that provide a general overview of the use of deep learning in navigation. This survey aims to fill this gap by presenting a comprehensive overview of the applications of deep learning in navigation. The paper is structured as follows: Section \ref{sec:overview} provides an overview of deep learning and its methods, while Section \ref{sec:activation_functions} discusses the different activation functions used in deep learning. Section \ref{sec:autonomous_navigation} presents an overview of navigation, autonomy, and autonomous navigation. Section \ref{sec:applications} discusses the applications of deep learning in navigation, and Section \ref{sec:deep_learning_components} delves into the various components of deep learning in autonomous navigation, such as perception, localization, mapping, planning, and control. Finally, Section \ref{sec:conclusion} concludes the paper and highlights future directions.

\begin{table*}[h]
\caption{Recent articles on deep learning for navigation}
\label{tab:recent_articles} 
\begin{tabular*}{\linewidth}{ p{13.5cm} @{\extracolsep{\fill}} l }
\toprule
\textbf{Title}  & \textbf{Application} \\ 
\midrule
Hierarchical multi-robot navigation and formation in unknown environments via deep reinforcement learning and distributed optimization \cite{chang2023hierarchical} & Multi-robot Navigation \\
HINNet: Inertial navigation with head-mounted sensors using a neural network \cite{hou2023hinnet} & Inertial Navigation \\
Multi-sensor integrated navigation/positioning systems using data fusion: From analytics-based to learning-based approaches \cite{zhuang2023multi} & Integrated Navigation \\
Study of convolutional neural network-based semantic segmentation methods on edge intelligence devices for field agricultural robot navigation line extraction \cite{yu2023study} & Visual Navigation \\
Goal-guided Transformer-enabled Reinforcement Learning for Efficient Autonomous Navigation \cite{huang2023goal} &  Autonomous Navigation \\ 
DeepNAVI: A deep learning based smartphone navigation assistant for people with visual impairments \cite{kuriakose2023deepnavi} & Visual Navigation \\ 
Monocular vision with deep neural networks for autonomous mobile robots navigation \cite{sleaman2023monocular} & Visual Navigation \\ 
URWalking: Indoor Navigation for Research and Daily Use \cite{ludwig2023urwalking} & Indoor Navigation \\ 
A Simple Self-Supervised IMU Denoising Method For Inertial Aided Navigation \cite{yuan2023simple} & Inertial Navigation \\ 
Multi-Scale Fully Convolutional Network-Based Semantic Segmentation for Mobile Robot Navigation \cite{dang2023multi} & Visual Navigation \\ 
Drone Navigation Using Octrees and Object Recognition for Intelligent Inspections \cite{martinez2023drone} & Visual Navigation  \\ 
Deep learning-enabled fusion to bridge GPS outages for INS/GPS integrated navigation \cite{liu2022deep} & Inertial Navigation \\ 
Deep learning based wireless localization for indoor navigation \cite{ayyalasomayajula2020deep} & Indoor Navigation \\ 
Efficient and robust LiDAR-based end-to-end navigation \cite{liu2021efficient} & Terrain modelling \\
End-to-End Deep Learning Framework for Real-Time Inertial Attitude Estimation using 6DoF IMU \cite{golroudbari2023end} & Inertial Navigation \\
\bottomrule
\end{tabular*}
\end{table*}

\begin{table*}[h]
\caption{Recent surveys on navigation or deep learning for navigation}
\label{tab:recent_surveys}
\begin{tabular*}{\linewidth}{l @{\extracolsep{\fill}}l p{11.5cm} }
\toprule
\textbf{Year} & \textbf{Topic} & \textbf{Highlight} \\ 
\midrule
 2017 & Unmanned Aerial Vehicles & Deep learning for UAV (navigation, motion control, and situational awareness) \cite{carrio2017review}. \\
2017 & Adaptive Navigation & AI-based planetary rovers autonomous navigation \cite{wong2017adaptive}. \\
2019 & Deep Learning Approaches & State-of-the-art deep learning approaches \cite{alom2019state}. \\
2019 & Visual Navigation & Reinforcement Learning for visual autonomous navigation \cite{ejaz2019autonomous}. \\
2020 & Deep Learning Approaches & Major architectures and applications of deep learning \cite{emmert2020introductory}. \\
2020 & Autonomous Driving & Deep learning architectures Autonomous Driving \cite{grigorescu2020survey}. \\
2020 & Inertial Sensors & Enhancement of various aspects of inertial sensors (sensor fusion, calibration, and navigation) \cite{li2020inertial}. \\
2020 & Autonomous Driving & Deep learning architectures for autonomous driving \cite{ni2020survey}. \\
2020 & Perception & Localization and Mapping \cite{chen2020survey} \\
2021 & Convolutional Neural Network & Applications of CNNs \cite{li2021survey}. \\
2021 & Indoor Navigation & Machine learning for indoor localization \cite{roy2021survey}. \\
2021 & Assistive navigation &  Outdoor and urban navigation for people with visual impairments \cite{el2021systematic}. \\
2022 & Visual Navigation & Agricultural robot navigation \cite{wang2022applications}. \\
2022 & Spacecraft Navigation & Deep learning for spacecraft dynamics control, guidance and navigation \cite{silvestrini2022deep}. \\
2022 & Visual Navigation & Unmanned underwater vehicles \cite{qin2022survey}. \\
2022 & Unmanned Aerial Vehicle & Reinforcement Learning for UAVs \cite{almahamid2022autonomous}. \\
2022 & Autonomous Driving & Visual obstacle detection for unmanned ground vehicles \cite{badrloo2022image} \\
2022 & Autonomous Driving & Deep Learning for Perception \cite{wen2022deep} \\
2022 & Spacecraft Navigation& Autonomous relative navigation for orbital applications \cite{song2022deep} \\
2022 & Learning-based methods for perception \cite{tang2022perception} \\
\bottomrule
\end{tabular*}
\end{table*}

\section{A brief overview of deep learning \label{sec:overview}}
In this section we provide a brief overview of deep learning for a more detailed discussion, the reader is referred to \cite{goodfellow2016deep}. We begin by defining machine learning and deep learning, followed by a brief history of deep learning. We then discuss the different types of deep learning algorithms and their applications. 

Machine learning is a field of artificial intelligence focused on developing algorithms that learn and improve from experience without being explicitly programmed. These algorithms identify patterns and make predictions based on data inputs, improving their performance over time with more data. Deep learning is a specialized type of machine learning based on Artificial Neural Networks (ANNs), inspired by the structure of the human brain. Deep learning algorithms consist of multiple layers of ANNs that are trained to automatically extract and learn relevant features from large amounts of data. This makes them well-suited for tasks such as image and speech recognition, natural language processing, and autonomous navigation. In summary, machine learning is a broad field that includes deep learning, with deep learning being a specialized type of machine learning based on ANNs \cite{goodfellow2016deep}.

Deep learning has its roots in the 1940s and 1950s with the introduction of the concept of artificial neurons by Warren McCulloch and Walter Pitts \cite{mcculloch1943logical} which was later extended by Frank Rosenblatt \cite{rosenblatt1958perceptron} to include a learning mechanism. In 1957, Rosenblatt introduced the perceptron \cite{rosenblatt1958perceptron}, which is single-layer artificial neural network that is used for binary classification and composed of a set of artificial neurons, each of which is connected to an input and has a weight associated with it. The output is determined by the weighted sum of the inputs, which is then passed through an activation function. The weights of the perceptron are adjusted during training to minimize the error between the output and the desired output. However, the field did not gain significant traction until the late 1980s and early 1990s, when the backpropagation algorithm \cite{rumelhart1986learning} was introduced for training ANNs. 

In the early 2000s, deep learning began to be applied to tasks such as image \cite{lecun1998gradient} and speech recognition \cite{hinton2012deep}. The success of these early applications led to increased interest in the field and further developments. In the late 2000s and early 2010s, deep learning began to gain significant attention due to the availability of large amounts of data and the development of more powerful hardware for training.

AlexNet \cite{krizhevsky2017imagenet}, a deep convolutional neural network trained on a large dataset of images, achieved state-of-the-art results in the ImageNet visual recognition challenge in 2012, which marked a turning point in deep learning.

Since then, deep learning has been used in a wide range of applications, including image recognition, natural language processing \cite{otter2020survey}, speech recognition \cite{deng2014ensemble}, self-driving cars \cite{rao2018deep}, healthcare \cite{esteva2019guide}, computational biology \cite{angermueller2016deep}, and gaming \cite{justesen2019deep}. The field has continued to evolve and improve, with the development of new architectures and techniques such as Generative Adversarial Networks (GANs) \cite{aggarwal2021generative}, and attention mechanisms \cite{vaswani2017attention}.

Deep learning has emerged as a highly effective tool for solving complex tasks, and it is currently being widely used across various industries and research fields. The increasing availability of large data sets and powerful computing resources has enabled more extensive use of deep learning, and the field's future looks very promising, with even more breakthroughs and advancements expected to come. Some remarkable examples of recent advancements in deep learning include the development of the GPT-4 language model \cite{openai2023gpt}, which has shown impressive results in natural language processing tasks, as well as AlphaFold \cite{ruff2021alphafold}, which can predict the complex process of protein folding. DeepMind's AlphaGo \cite{silver2016mastering}, which achieved groundbreaking results in the ancient Chinese board game, is also a notable example of deep learning's potential.

There are several common structures of deep neural networks, each of which is suited for different types of tasks and applications. Some of the most common structures will be discussed in the following sections.

\subsection{Artificial Neuron}
\begin{figure}
\centering
	\includegraphics[width=0.5\textwidth]{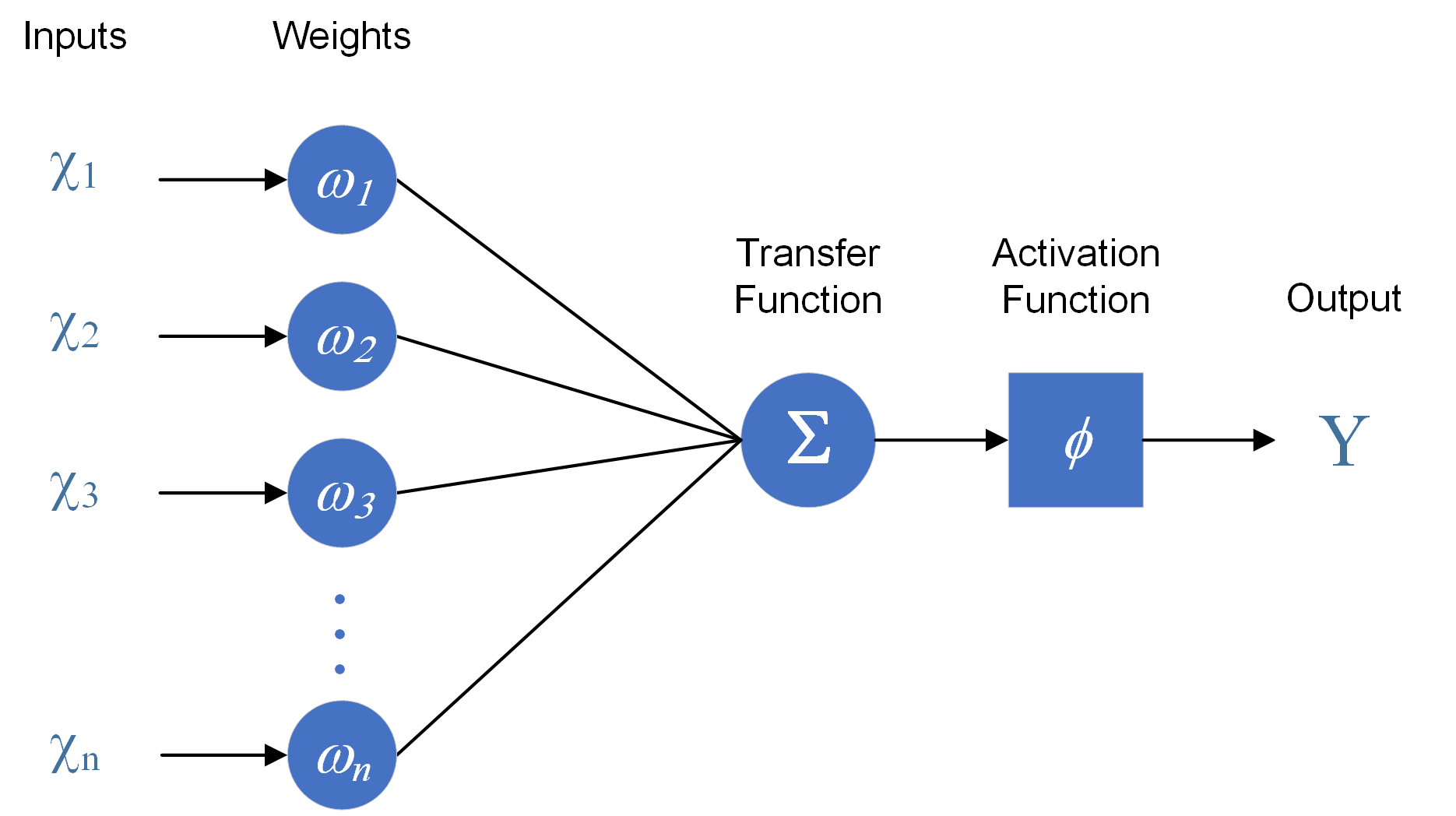}
	\caption{Structure of an artificial neuron}
	\label{fig:neuron}
\end{figure}

An artificial neuron is a mathematical function that is used to model the behavior of biological neurons. It was first introduced by Warren McCulloch and Walter Pitts in 1943. The idea of artificial neurons is that an input signal weighted by a weight $w$ with a bias $b$ is passed through an activation function $f$ to produce an output signal $y$ \cite{schmidhuber2015deep}. Figure ~\ref{fig:neuron} and Equation ~\ref{eq:neuron} shows the structure of an artificial neuron.

\begin{equation} 
    y = f(w \cdot x + b) 
	\label{eq:neuron}
\end{equation}

where the activation function $f$ performs a non-linear transformation on the weighted sum of the inputs. 

\subsection{Deep Feedforward Neural Networks} 
\begin{figure}
\centering
	\includegraphics[width=\linewidth]{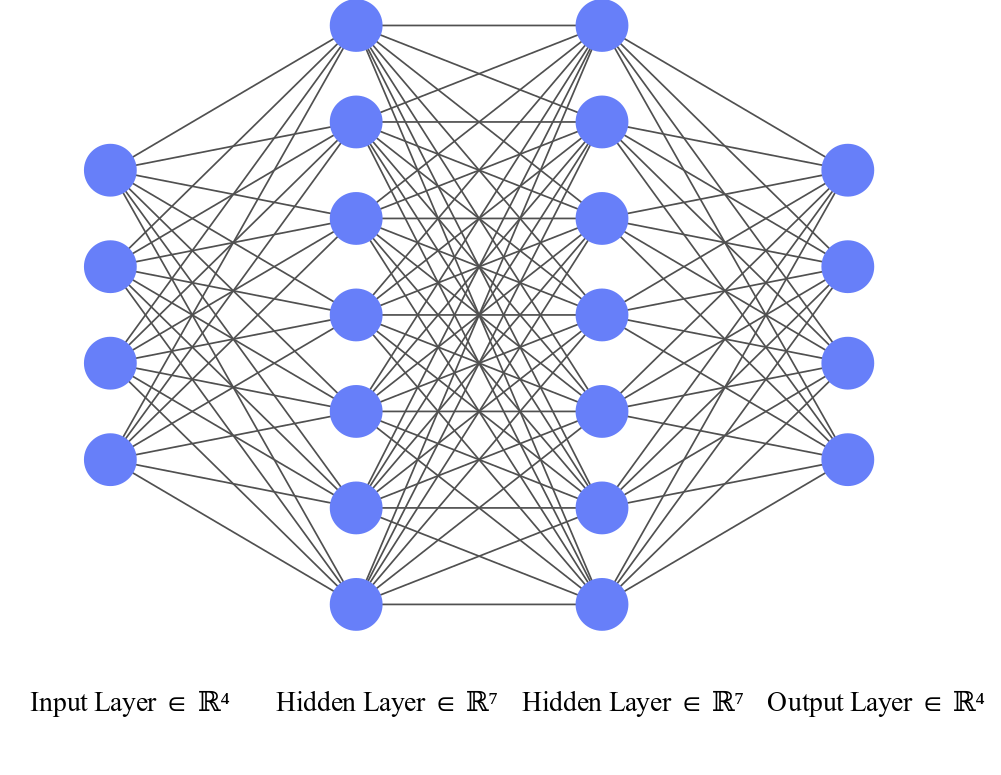}
	\caption{Structure of a deep feedforward neural network}
	\label{fig:feedforward}
\end{figure}

A deep feedforward neural network is composed of multiple layers of artificial neurons, each of which is fully connected to the next layer. This type of neural network is also known as a multilayer perceptron (MLP) \cite{goodfellow2016deep}. All data is passed through the network in a forward direction. The input is passed through the first layer, which is then passed through the second layer, and so on until the output of the network is produced. The output is determined by the output of the last layer. Figure ~\ref{fig:feedforward} and Equation ~\ref{eq:feedforward} shows the structure of a deep feedforward neural network. 

\begin{equation}
	\label{eq:feedforward}
\begin{aligned}
&\text{Input:} \quad \mathbf{x} \in \mathbb{R}^n \\
&\text{Hidden layers:} \quad \mathbf{h}_1, \mathbf{h}_2, \dots, \mathbf{h}_L \\
&\text{Weights:} \quad \mathbf{w}_1, \mathbf{w}_2, \dots, \mathbf{w}_L \\
&\text{Biases:} \quad \mathbf{b}_1, \mathbf{b}_2, \dots, \mathbf{b}_L \\
&\text{Activation function:} \quad f \\
&\text{Output:} \\
&\quad \mathbf{h}_1 = f(\mathbf{w}_1 \cdot \mathbf{x} + \mathbf{b}_1) \\
&\quad \mathbf{h}_i = f(\mathbf{w}i \cdot \mathbf{h}{i-1} + \mathbf{b}_i), \quad i = 2, \dots, L \\
&\quad \mathbf{y} = \mathbf{w}L \cdot \mathbf{h}{L-1} + \mathbf{b}_L \\
&\quad \text{Output:} \quad \mathbf{y} \in \mathbb{R}^m \\
\end{aligned}
\end{equation}

\subsection{Autoencoder}
An autoencoder is designed to learn a compressed representation of input data and composed of two main parts: (1) encoder and (2) decoder. The encoder takes in the input data and learns to compress it into a lower-dimensional representation, typically called a latent representation or bottleneck. The decoder then takes this compressed representation and learns to reconstruct the original input data from it.
The goal is to learn a compressed representation that captures the most important features of the input data while discarding the unnecessary or redundant information. They can be used for a variety of tasks such as dimensionality reduction, data denoising, and generative modeling. 
There are different variations of autoencoders such as convolutional autoencoder, variational autoencoder, and more. Figure ~\ref{fig:autoencoder} shows the structure of a standard feedforward autoencoder:

\begin{figure}
\centering
	\includegraphics[width=0.5\textwidth]{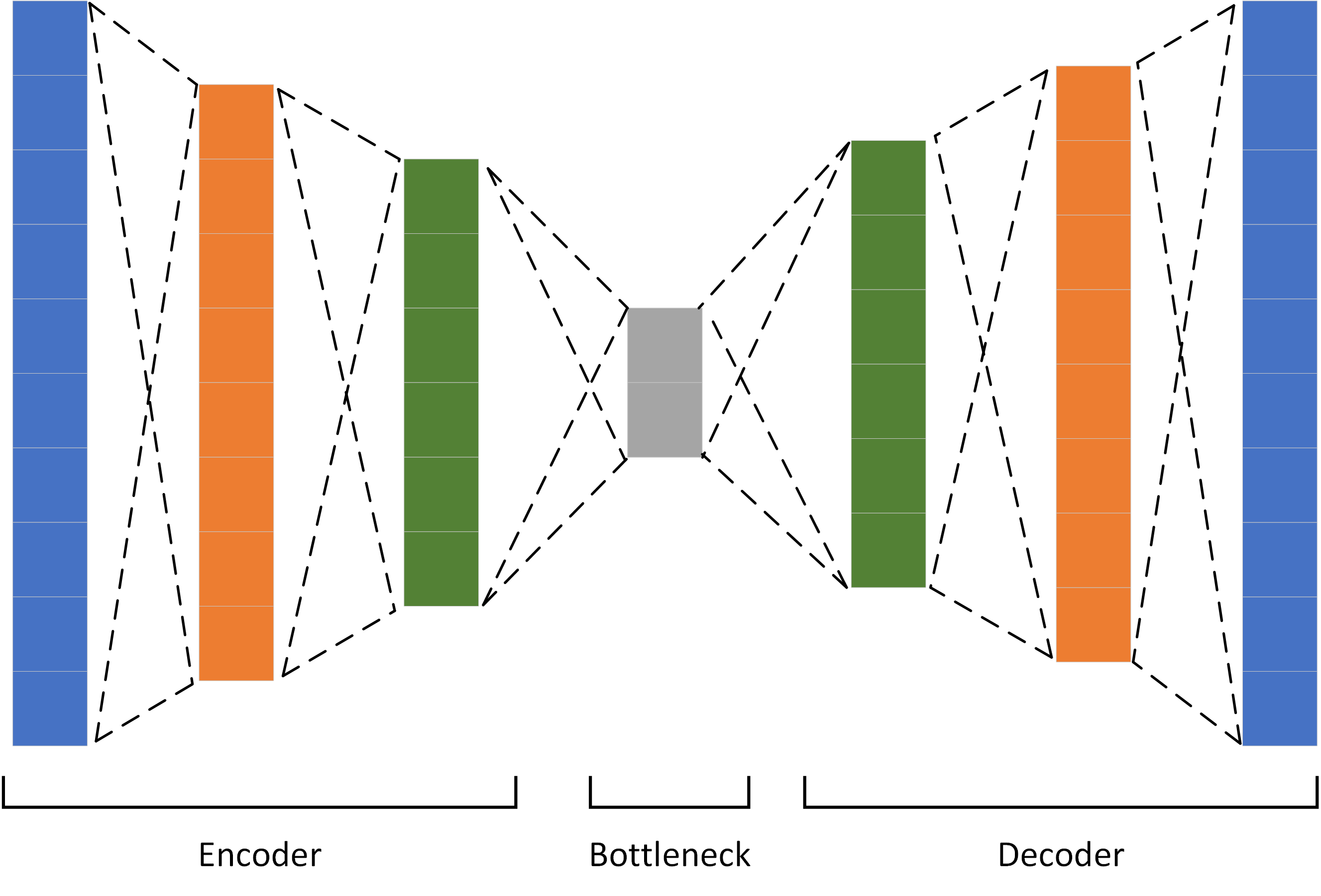}
	\caption{Structure of an autoencoder}
	\label{fig:autoencoder}
\end{figure}

\subsection{Convolutional Neural Network (CNN)}
A CNN used convolution operation to extract features from data. This operation involves combining two sets of data, such as an image and a filter, in order to extract meaningful features from the input. By using this technique, CNNs can identify patterns in complex datasets that would otherwise be difficult or impossible for traditional algorithms to detect. CNNs are particularly useful for tasks such as image recognition, object detection, and image segmentation.  It is composed of multiple layers which are arranged in a hierarchical structure, with the lower layers extracting simple features such as edges, and the higher layers extracting more complex features such as patterns and objects. Figure ~\ref{fig:cnn} shows the structure of a CNN:

\begin{figure}
\centering
	
	\includegraphics[width=0.5\textwidth]{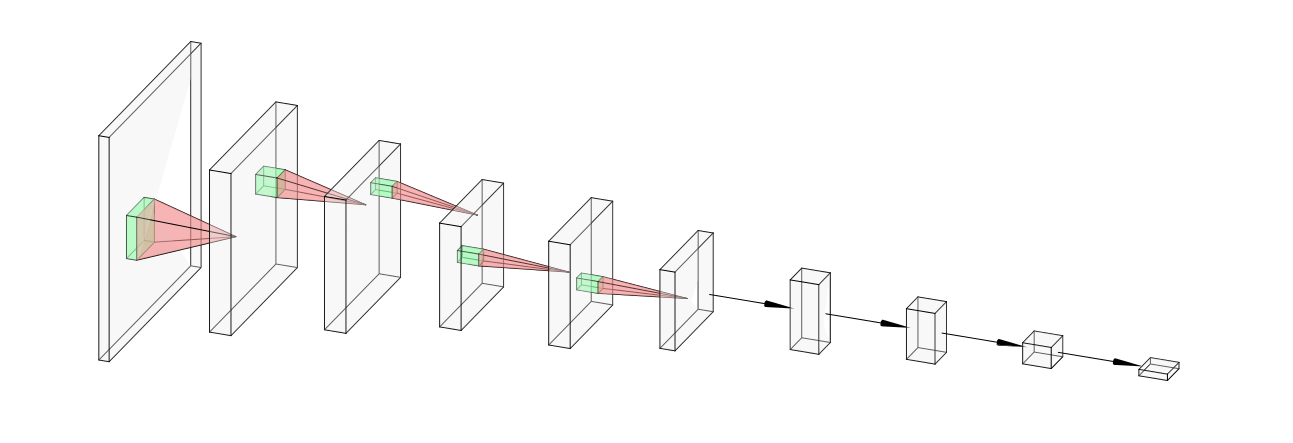}
	\caption{Structure of a CNN}
	\label{fig:cnn}
\end{figure}

\subsection{Deep Belief Networks (DBN)} 
Deep belief network was first introduced in 2006 by Hinton et al. \cite{hinton2006fast} which is composed of multiple layers of hidden units, where each layer is a Restricted Boltzmann Machine (RBM). RBMs are shallow, two-layer neural networks where the visible units are connected to the hidden units, but not to each other. In a DBN, the RBMs are stacked on top of each other to form a deep network. The equation for an RBM can be represented as follows:

\begin{equation}
\begin{aligned}
P(v,h) = \frac{1}{Z} \exp (-E(v,h)) 
\end{aligned}
\end{equation}

\begin{equation}
\begin{aligned}
E(v,h) = -\sum_{i}^{m} b_i v_i - \sum_{j}^{n} c_j h_j - \sum_{i}^{m} \sum_{j}^{n} v_i w_{ij} h_j \
\end{aligned}
\end{equation}

where $v$ and $h$ are the visible and hidden layer activations respectively, $b$ and $c$ are the biases of the visible and hidden layer, $w$ is the weight matrix between the visible and hidden layer, and $Z$ is the normalizing constant.

DBNs can be trained using an unsupervised learning algorithm, where the input data is used to learn the generative model. DBNs have been applied in several domains, including computer vision, speech recognition, natural language processing, and bioinformatics, for tasks such as image classification, feature extraction, and language recognition.

\subsection{Recurrent Neural Network (RNN)}
\begin{figure}
	\centering
		
		\includegraphics[width=0.48\textwidth]{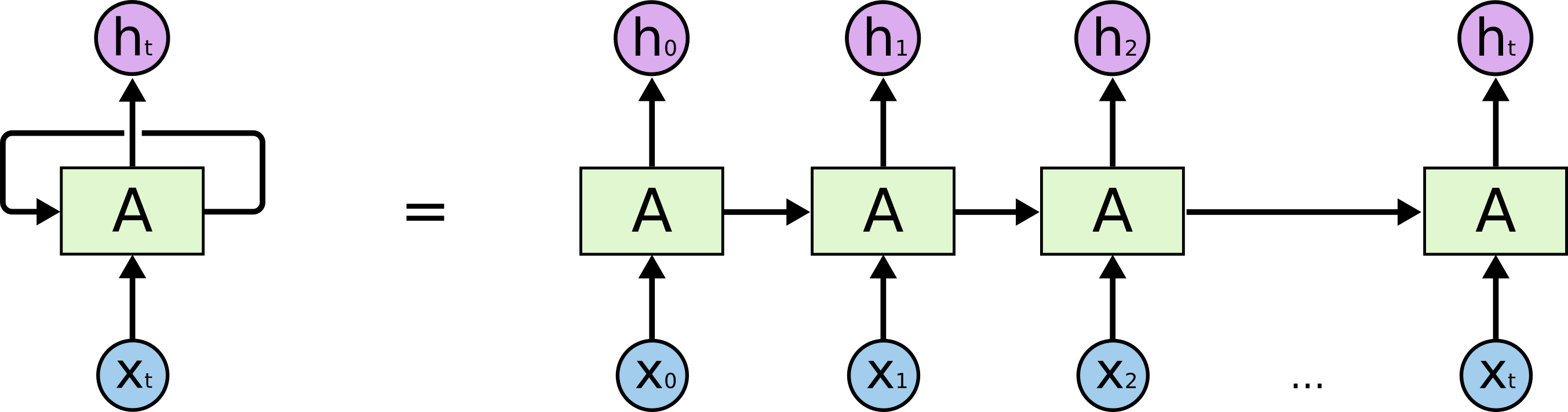}
		\caption{A simple RNN}
		\label{fig:rnn}
	\end{figure}
	
Recurrent neural networks have gained popularity in recent years due to their ability to process sequential data, such as time series, speech, and text. They are characterized by the presence of recurrent connections, which allow information to flow through the network over multiple time steps. This allows RNNs to maintain a hidden state that can be updated at each time step, allowing them to remember previous input and use that information to inform their current output \cite{goodfellow2016deep}.

The most basic form of an RNN is the simple recurrent neural network (SRN), which has a single hidden layer with recurrent connections. The SRN's hidden state at time $t$, $h_t$, is a function of the current input $x_t$ and the previous hidden state $h_t-1$. The output $y_t$ is then computed based on the hidden state $h_t$. The basic equation of a Recurrent Neural Network can be represented as follows:
\begin{equation} 
	\label{eq:recurrent}
	\begin{aligned}
		&\text{Input:} \quad x \in \mathbb{R}^n \\
		&\text{Output:} \quad y \in \mathbb{R}^m \\
		&\text{Hidden state:} \quad h \in \mathbb{R}^k \\
		&\text{Weights:} \quad W(hh), W(hx) \\
		&\text{Biases:} \quad b(h) \\
		&\text{Activation function:} \quad f \\
		&\text{Output:} \quad y = f(W(yh)h + b(y))
	\end{aligned}
\end{equation}
where $h$ is the hidden state, x is the input, $W(hh)$, $W(hx)$ and $b(h)$ are the weights and bias for the hidden state, $W(yh)$ and $b(y)$ are the weights and bias for the output, and $f$ is the activation function. 
The hidden state at the current time step is a function of the previous hidden state and the current input, with the weights and bias being learned during training. The output of the RNN at each time step is a function of the current hidden state, with the weights and bias for the output also being learned during training. The hidden state at time step $t$ is defined as follows:

\begin{equation} 
	h(t) = f(W(hh)h(t-1) + W(hx)x(t) + b(h))
\end{equation}

where $h(t-1)$ is the hidden state at the previous time step, $x(t)$ is the input at time step t, and $f$ is the activation function. The output of the RNN at each time step is a function of the current hidden state, with the weights and bias for the output also being learned during training. 

Long short-term memory (LSTM) and gated recurrent units (GRUs) are two variations of RNNs that have been developed to address the problem of vanishing gradients which occur when the gradients of the parameters of the network become very small during the backpropagation process, making it difficult to train the network \cite{hochreiter1998vanishing}. LSTM and GRUs both introduce gating mechanisms that allow the network to selectively choose which information to forget or remember at each time step, thus addressing the vanishing gradients problem.

Another variant of RNNs is the bidirectional RNN (BRNN), which allows the network to take into account future context in addition to past context. In a BRNN, two RNNs are used to processes the input in the forward direction and reverse direction. The output of the two RNNs is then concatenated and used as the final output. 

RNNs have been used in a variety of applications, including natural language processing, speech recognition, and time series prediction and used for tasks such as language translation, text summarization, sentiment analysis, and forecast future values based on past values.

One of the main benefits of RNNs is their ability to handle sequential data, which allows them to model temporal dependencies and patterns. However, RNNs also have some limitations. One limitation is that they can be sensitive to the order of input data, which can be a problem when dealing with permuted data. Additionally, RNNs can be computationally expensive and may require a large amount of memory to store the hidden states. Figure \ref{fig:rnn} shows a diagram of a simple RNN.

\subsection{Gate Recurrent Unit (GRU)}
GRU is introduced by Cho et al. in 2014 \cite{cho2014learning}. The main difference between a traditional RNN and a GRU is that a GRU has two gates, called the update gate and the reset gate, which control the flow of information between the previous hidden state and the current hidden state.

The update gate controls how much of the previous hidden state should be kept, and the reset gate controls how much of the previous hidden state should be forgotten. This allows a GRU to effectively handle long-term dependencies in the input sequence, as it can selectively choose which information to keep and which to discard.

The mathematical equation for the hidden state update in a GRU is as follows:

\begin{equation}
	h_t = z_t*h_{t-1} + (1-z)*\tilde{h}_t
\end{equation}
where $h_t$ is the hidden state at time $t$, $h_{t-1}$ is the hidden state at time $t-1$, $z_t$ is the update gate, and $\tilde{h}_t$ is the candidate hidden state.

The update gate $z_t$  is computed using the following equation:

\begin{equation}
	z_t = \sigma(W_zx_t + U_zh_{t-1} + b_z)
\end{equation}
where $W_z$ is the weight matrix for the input, $U_z$ is the weight matrix for the previous hidden state, $x_t$ is the input at time $t$, and $b_z$ is the bias term for the update gate.

The candidate hidden state $\tilde{h}_t$ is computed using the following equation:

\begin{equation}
	\tilde{h}_t = tanh(W_hx_t + U_h(h_{t-1} \circ r_t) + b_h)
\end{equation}

where $W_h$ is the weight matrix for the input, $U_h$ is the weight matrix for the previous hidden state, $r_t$ is the reset gate, and $b_h$ is the bias term for the candidate hidden state. $\circ$ is the element-wise multiplication operator.

The reset gate $r_t$ is computed using the following equation:
\begin{equation}
	r_t = \sigma(W_rx_t + U_rh_{t-1} + b_r)
\end{equation}
where $W_r$ is the weight matrix for the input, $U_r$ is the weight matrix for the previous hidden state, $x_t$ is the input at time $t$, and $b_r$ is the bias term for the reset gate.

\begin{figure}
\centering
	\includegraphics[width=0.5\textwidth]{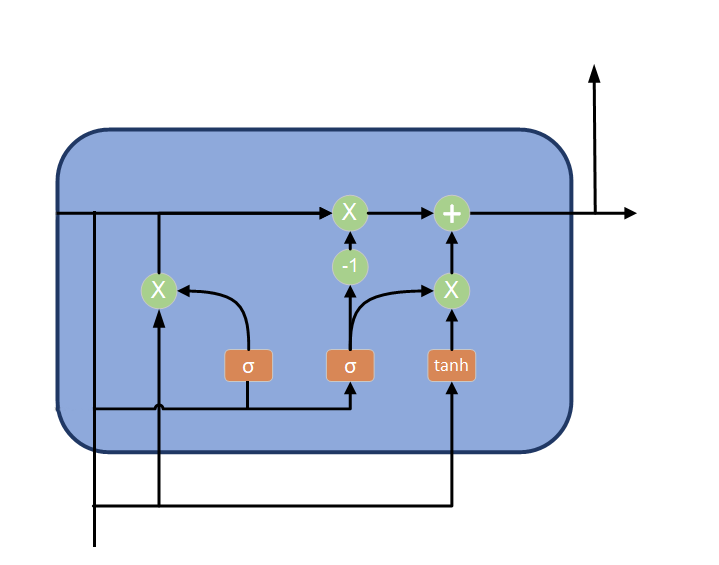}
	\caption{A GRU}
	\label{fig:gru}
\end{figure}

The output of the GRU at time $t$ is calculated by using the following equation:

\begin{equation}
	y_t = \sigma(W_yh_t + b_y)
\end{equation}

where $W_y$ is the weight matrix for the hidden state and by is the bias term for the output. Figure \ref{fig:gru} shows a diagram of a GRU.

\subsection{Long Short-Term Memory (LSTM)}
An LSTM is introduced in 1997 by S. Hochreiter and J. Schmidhuber \cite{hochreiter1997long}. It has a memory cell that is used to retain information from previous inputs and use it to process the current input. LSTMs are consists of three gates, an input gate, an output gate, and a forget gate. The input gate is used to decide which values from the current input should be added to the memory cell. The forget gate is used to decide which values from the memory cell should be removed. The output gate is used to decide which values from the memory cell should be used to produce the output. LSTMs are particularly useful for tasks such as language modeling, machine translation, and speech recognition. The main advantage of LSTMs over other RNNs is that they can learn long-term dependencies. Its equation is as follows:

Input Gate:
\begin{equation}
i_t = \sigma(W_{xi}x_t + W_{hi}h_{t-1} + b_i)
\end{equation}

Forget Gate:
\begin{equation}
f_t = \sigma(W_{xf}x_t + W_{hf}h_{t-1} + b_f)
\end{equation}

Output Gate:
\begin{equation}
o_t = \sigma(W_{xo}x_t + W_{ho}h_{t-1} + b_o)
\end{equation}

Cell State:
\begin{equation}
\tilde{C}_t = \tanh(W_{xc}x_t + W_{hc}h_{t-1} + b_c)
\end{equation}

\begin{figure}
\centering
	
	\includegraphics[width=0.5\textwidth]{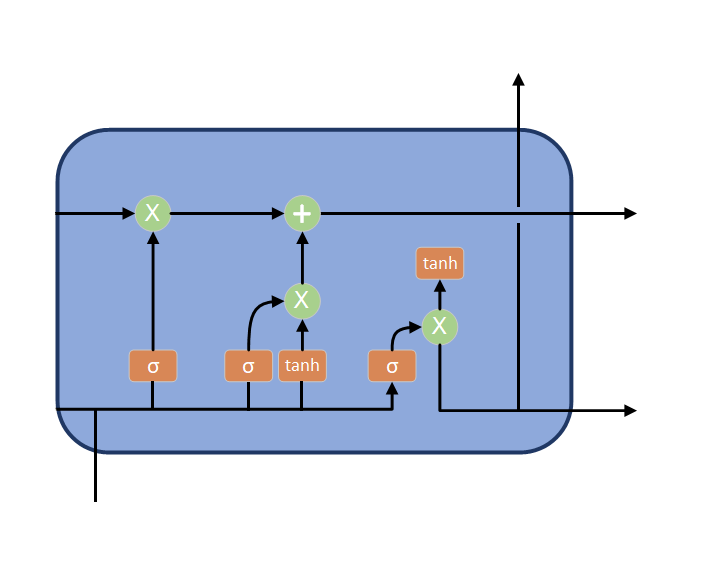}
	\caption{An LSTM}
	\label{fig:lstm}
\end{figure}

Figure \ref{fig:lstm} shows a diagram of an LSTM.

\subsection{Transformer}

Transformer is a neural network architecture introduced by Vaswani et al. in 2017 \cite{vaswani2017attention}. It is a type of encoder-decoder architecture that utilizes self-attention mechanisms to process sequential data.

The transformer architecture consists of an encoder and a decoder, each of which is made up of multiple layers. The encoder takes in a sequence of input data, such as a sentence or a document, and produces a set of hidden representations called the "keys" and "values". The decoder then takes in the keys and values and generates a new sequence of output data, such as a translated sentence or a summary of the input.

One of the key features of the this architecture is its use of self-attention mechanisms which allows the model to weigh the importance of different parts of the input sequence when generating the output. This is done by computing a set of attention weights for each element in the input sequence, which are used to weigh the contribution of each element to the final output.

Another important feature is the use of multi-head attention. Multi-head attention allows the model to attend to different parts of the input sequence in parallel, rather than serially. This allows the model to learn more complex relationships between different parts of the input sequence.

This architecture has been used in a wide range of natural language processing tasks, such as machine translation, text summarization, and language modeling. It has also been adapted for other types of data, such as images and time series. It has proven to be very effective at these tasks and has become a popular choice among researchers and practitioners.

\subsection{Generative Adversarial Network (GAN)}
The GAN was first introduced in 2014 by Goodfellow and his colleagues \cite{goodfellow2020generative}. It is designed to generate new, previously unseen examples from a given dataset, such as images, text, or audio. GANs consist of two main components: a generator network and a discriminator network.

The generator network is responsible for creating new examples that are similar to the examples in the given dataset which takes a random noise input and maps it to a sample from the target distribution. The generator is typically a deep neural network with multiple layers, such as a fully connected or convolutional neural network.

The discriminator network is responsible for distinguishing between the examples generated by the generator and the examples from the real dataset. It takes an example as input and produces a probability value indicating the likelihood that the example is real. The discriminator is also typically a deep neural network with multiple layers.

The two networks are trained together in a two-player minimax game, where the generator is trying to produce examples that can fool the discriminator into thinking they are real, while the discriminator is trying to correctly identify which examples are real and which are fake. As the training progresses, the generator becomes better at producing realistic examples, and the discriminator becomes better at identifying them.

The goal of training is to reach a point where the generator can produce examples that are indistinguishable from real examples, and the discriminator is unable to make a clear distinction between the two. At this point, the generator can be used to generate new examples from the target distribution, such as new images that look like photographs of faces, or new audio samples that sound like speech.

GANs have been used in a wide range of applications, such as image synthesis, image-to-image translation, video synthesis, and text-to-speech synthesis. They have also been used for unsupervised learning tasks, such as anomaly detection and feature learning.

It's worth noting that GANs are considered one of the most difficult models to train and stabilize due to the adversarial nature of the training process and the potential for the generator and discriminator to get stuck in equilibrium where neither network makes progress. Therefore, multiple techniques have been proposed to stabilize the training process like Wasserstein GAN \cite{arjovsky2017wasserstein}, Improved WGAN \cite{gulrajani2017improved}, and BEGAN \cite{berthelot2017began}.

Table \ref{tab:deep_learning} summarizes the main architectures discussed in this section and their main properties and applications.

\begin{table*}[h]
\caption{Deep Learning Architectures, Properties, and Applications. \label{tab:deep_learning}}
\begin{tabularx}{\textwidth}{p{2cm}p{8.cm}p{8.3cm}}
    \toprule

\textbf{Arch} & \textbf{Properties} & \textbf{Applications} \\
\midrule
CNN & 
Uses convolutional layers \newline
Proper for extracting features from images
&
Obstacle detection \cite{he2022improved}\newline
Object classification \cite{aftf2019indoor}\newline
Lane detection \cite{wang2020cnn} \newline
Traffic sign detection \cite{li2022traffic} \newline
Outdoor Navigation \cite{ kouris2018learning} \newline
Indoor Navigation \cite{gong2021deepnav} \\
\hline
RNN &
Uses recurrent layers \newline
Memory and store hidden states \newline
Extracts features from sequential data \newline 
Learn temporal dependencies &
Attitude Estimation \cite{chumuang2022feature} \newline
Denoising \cite{brossard2020denoising} \newline
Dead Reckoning \cite{topini2020lstm} \newline
Inertial Navigation \cite{tong2019cascade} \\
\hline
Transformer & 
Uses self-attention mechanisms  \newline
Uses multi-head attention & 
Visual Navigation \cite{du2021vtnet} \newline
Pedestrian Navigation \cite{yu2020spatio} \\
\hline
GAN &
Uses a generator network \newline 
Uses a discriminator network & 
Trajectory Prediction \cite{roy2019vehicle} \newline
Obstacle Detection \cite{dimas2019obstacle} \newline
Path Planning \cite{mohammadi2018path} \\
\bottomrule
\end{tabularx}
\end{table*}

\section{Activation Functions \label{sec:activation_functions}}
Activation functions introduce non-linearity into the network. Without non-linearity, a neural network would be limited to linear operations and would not be able to learn complex representations \cite{goodfellow2016deep}.

One of the earliest activation functions used in neural networks is the step function, invented by W. McCulloch and W. Pitts in 1943 \cite{mcculloch1943logical}. The step function outputs a value of 1 if the input is greater than a certain threshold and 0 otherwise. This function is not used in modern neural networks, as it is not differentiable and therefore not useful for backpropagation.

Another early activation function is the sigmoid function, invented by Frank Rosenblatt in 1958 \cite{rosenblatt1958perceptron} which maps the input to a value between 0 and 1, making it useful for binary classification tasks. However, it is not recommended for use in modern neural networks, as it can saturate and produce vanishing gradients. It takes an input between $-\infty$ and $+\infty$ and outputs a value between 0 and 1. This can be considered as a likelihood or probability of the input being in a certain class.
It is defined as follows:

\begin{equation}
    f(x) = \frac{1}{1 + e^{-x}}
\end{equation}

A more recent activation function is the rectified linear unit (ReLU) function, invented by Fukushima \cite{fukushima1975cognitron} in 1975. The ReLU function outputs the input if it is positive and 0 otherwise, making it computationally efficient and effective in avoiding the vanishing gradients problem. It is defined as follows:

\begin{equation}
    f(x) = \max(0, x)
\end{equation}

Leaky ReLU is an extension of the ReLU function, which introduces small negative slope in the negative part, to overcome the dying neurons problem \cite{maas2013rectifier}. The difference between Leaky ReLU and ReLU is that Leaky ReLU is how to handle negative values. Instead of setting them to zero, Leaky ReLU sets them to a small negative value. This small negative value is called the leaky coefficient which is a hyperparameter that can be tuned. It is defined as follows:

\begin{equation}
    f(x) = \max(0, x) + \alpha \min(0, x)
\end{equation}

where $\alpha$ is the leaky coefficient.
Another popular activation function is the hyperbolic tangent (tanh) function, which maps the input to a value between -1 and 1. This function is similar to the sigmoid function but is symmetric around the origin, which can be useful in certain situations. Tanh could be used instead of sigmoid function to avoid the vanishing gradient problem. It is defined as follows:

\begin{equation}
    f(x) = \frac{e^x - e^{-x}}{e^x + e^{-x}}
\end{equation}

The exponential linear unit (ELU) function, invented by Clevert \cite{clevert2015fast} in 2015, which is similar to ReLU but it is defined as negative when the input is less than zero. This function can be useful to help the network to learn faster and overcome the problem of dying neurons.
\begin{equation}
    f(x) = \begin{cases}
    x & \text{if } x > 0 \\
    \alpha(e^x - 1) & \text{if } x \leq 0 \\
    \end{cases}
\end{equation}
where $\alpha$ is the negative slope coefficient, which is a hyperparameter that can be tuned.

Softmax function is another activation function that is commonly used in the output layer of a neural network for multi-class classification problems \cite{bridle1990probabilistic,bridle1989training}. The softmax function maps the input to a probability distribution over multiple classes. It is defined as follows:

\begin{equation}
    f(x) = \frac{e^x}{\sum_{i=1}^n e^x}
\end{equation}

Softplus function is another activation function that is commonly used in the output layer of a neural network for multi-class classification problems \cite{dugas2000incorporating}. The softplus function maps the input to a probability distribution over multiple classes. It is defined as follows:

\begin{equation}
	f(x) = \log(1 + e^x)
\end{equation}

Swish function is commonly used in the output layer of a neural network for multi-class classification problems \cite{ramachandran2017searching} which maps the input to a probability distribution over multiple classes. It helps to overcome the problem of vanishing gradient making it suitable for deep neural networks. It is defined as follows:

\begin{equation}
    f(x) = x \cdot \text{sigmoid}(\beta x)
\end{equation}

\begin{equation}
    f(x) = x \cdot \frac{1}{1 + e^{-\beta x}}
\end{equation}

where $\beta$ is the slope coefficient, which is a hyperparameter that can a constant or a learnable parameter.

Randomized ReLU (RRelu) is introduced by Xu et al. in 2015\cite{xu2015empirical} which is a variation of the traditional ReLU. With ReLU, the negative part of the input is dropped or set to zero, whereas with RReLU, this negative part is assigned a non-zero slope. This means that a randomly determined slope is assigned to the negative part, and can update during the training process. The advantage of using this is that it improves the performance of a convolutional neural network and can lead to better results on image classification tasks. It is defined as follows:

\begin{equation}
    f(x) = \begin{cases}
    x & \text{if } x > 0 \\
    \alpha x & \text{if } x \leq 0 \\
    \end{cases}
\end{equation}

In 2019 Misra et al. \cite{misra2019mish} introduced a new activation function called Mish, which is self-regularized non-monotonic activation function. It combines the advantages of ReLU and tanh functions to create a more robust non-linearity. It can be mathematically defined as:
\begin{equation}
    f(x) = x \cdot \text{tanh}(\text{softplus}(x)) = x \cdot \text{tanh}(ln(1 + e^x))
\end{equation}

In conclusion, activation functions play a crucial role in deep neural networks by introducing non-linearity and allowing the network to learn complex representations. Different activation functions have their own advantages and disadvantages and are suitable for different types of problems. ReLU and its variants are widely used in modern neural networks due to their effectiveness and computational efficiency.

Some of popular activation functions has been listed in the table below:
In Table~\ref{tab1}, we summarized some related works in the navigation field using deep learning.

\begin{table*}[h]
\caption{Activation Functions.\label{tab1}}

\begin{tabular*}{\linewidth}{l l l}
\toprule
\textbf{Activation} & \textbf{Equation} & \textbf{Gradient}\\ \midrule
Sigmoid \cite{mcculloch1943logical} & $ \frac{1}{1 + e^{-x}}$ & $ \frac{e^{-x}}{(1+e^{-x})^2}$  \\ \midrule
ReLU \cite{nair2010rectified}& $\text{max}(0,x)$ & $\begin{cases} 1 & \text{if } x > 0 \\ 0 & \text{if } x \leq 0 \end{cases}$  \\\midrule
Leaky ReLU \cite{maas2013rectifier} & $\text{max}(ax,x)$ & $\begin{cases} 1 & \text{if } x > 0 \\ a & \text{if } x \leq 0 \end{cases}$  \\\midrule
Softmax \cite{bridle1989training}& $\frac{e^{x_i}}{\sum_{j=1}^{n}e^{x_j}}$ & $\frac{e^{x_i}(1-\frac{e^{x_i}}{\sum_{j=1}^{n}e^{x_j}})}{\sum_{j=1}^{n}e^{x_j}}$ \\ \midrule
Tanh & $\frac{e^{x}-e^{-x}}{e^{x}+e^{-x}}$ & $\frac{4e^{2x}}{(e^{x}+e^{-x})^2}$  \\ \midrule
Softplus \cite{dugas2000incorporating} & $\ln(1+e^{x})$ & $\frac{e^{x}}{1+e^{x}}$ \\ \midrule
Swish \cite{ramachandran2017searching} & $x \cdot$sigmoid$(\beta x)$ & $x \cdot \text{sigmoid}(\beta x) \cdot (1-\text{sigmoid}(\beta x)) + \text{sigmoid}(\beta x) \cdot \beta$  \\\midrule
Mish \cite{misra2019mish} & $x \cdot \text{tanh}(\ln(1+e^{x}))$ & $x \cdot \text{tanh}(\ln(1+e^{x})) \cdot (1-\text{tanh}(\ln(1+e^{x}))) + \text{tanh}(\ln(1+e^{x})) \cdot \ln(1+e^{x})'$  \\\midrule
ELU \cite{clevert2015fast} & $\text{max}(0,x) + \min(0, \alpha(e^{x}-1))$ & $1$ \\\midrule
PReLU \cite{he2015delving} & $\begin{cases} ax & \text{if } x < 0 \\ x & \text{if } x \geq 0 \end{cases}$ & $\begin{cases} a & \text{if } x < 0 \\ 1 & \text{if } x \geq 0 \end{cases}$ \\\midrule
GELU \cite{hendrycks2016baseline} & \begin{tabular}{l l } $ \frac{1}{2}x(1+\text{tanh}(\sqrt{\frac{2}{\pi}}(x+\alpha x^3)))$ \\$ \alpha = 0.044715 $\end{tabular}  & $\frac{1}{2}(1+\text{tanh}(\sqrt{\frac{2}{\pi}}(x+\alpha x^3))) + \frac{1}{2}x\sqrt{\frac{2}{\pi}}(1-\frac{\text{tanh}^2(\sqrt{\frac{2}{\pi}}(x+\alpha x^3))}{2})$ \\\midrule
SELU \cite{klambauer2017self} & $\begin{cases} \lambda x & \text{if } x \geq 0 \\ \lambda \alpha(e^{x}-1) & \text{if } x < 0 \end{cases}$ & $\begin{cases} \lambda & \text{if } x \geq 0 \\ \lambda \alpha(e^{x}) & \text{if } x < 0 \end{cases}$  \\\midrule
Softsign & $\frac{x}{1+|x|}$ & $\frac{1}{(1+|x|)^2}$   \\\midrule
Hard shrink & $\begin{cases} x & \text{if } x > \lambda \\ x & \text{if } x < -\lambda \\ 0 & \text{otherwise} \end{cases}$ & $\begin{cases} 1 & \text{if } x > \lambda \\ 1 & \text{if } x < -\lambda \\ 0 & \text{otherwise} \end{cases}$    \\\midrule
Soft shrink & $\begin{cases} x - \lambda & \text{if } x > \lambda \\ x + \lambda & \text{if } x < -\lambda \\ 0 & \text{otherwise} \end{cases}$ & $\begin{cases} 1 & \text{if } x > \lambda \\ 1 & \text{if } x < -\lambda \\ 0 & \text{otherwise} \end{cases}$    \\\midrule
Tanh shrink & $x-\text{tanh}(x)$ & $1-\text{tanh}^2(x)$    \\\midrule
LiSHT \cite{roy2019lisht} & $x \cdot \text{tanh}(x)$ & $x \cdot \text{tanh}(x) \cdot (1-\text{tanh}(x)) + \tanh(x)$    \\\midrule
Snake \cite{ziyin2020neural} & \begin{tabular}{l l } $x + \frac{1}{a}\sin^2(ax)$ \\ or \\ $x -\frac{1}{2a}\cos(2ax)+\frac{1}{2a}$ \end{tabular} & \begin{tabular}{l l }$\frac{1}{a}\cos^2(ax)$ \\or \\ $\frac{1}{2a}\cos(2ax)-\frac{1}{2a}$\end{tabular} \vspace{0.1cm}\\ 
\bottomrule
\end{tabular*}
\end{table*}

\begin{figure}
\centering
	\includegraphics[width=0.48\textwidth]{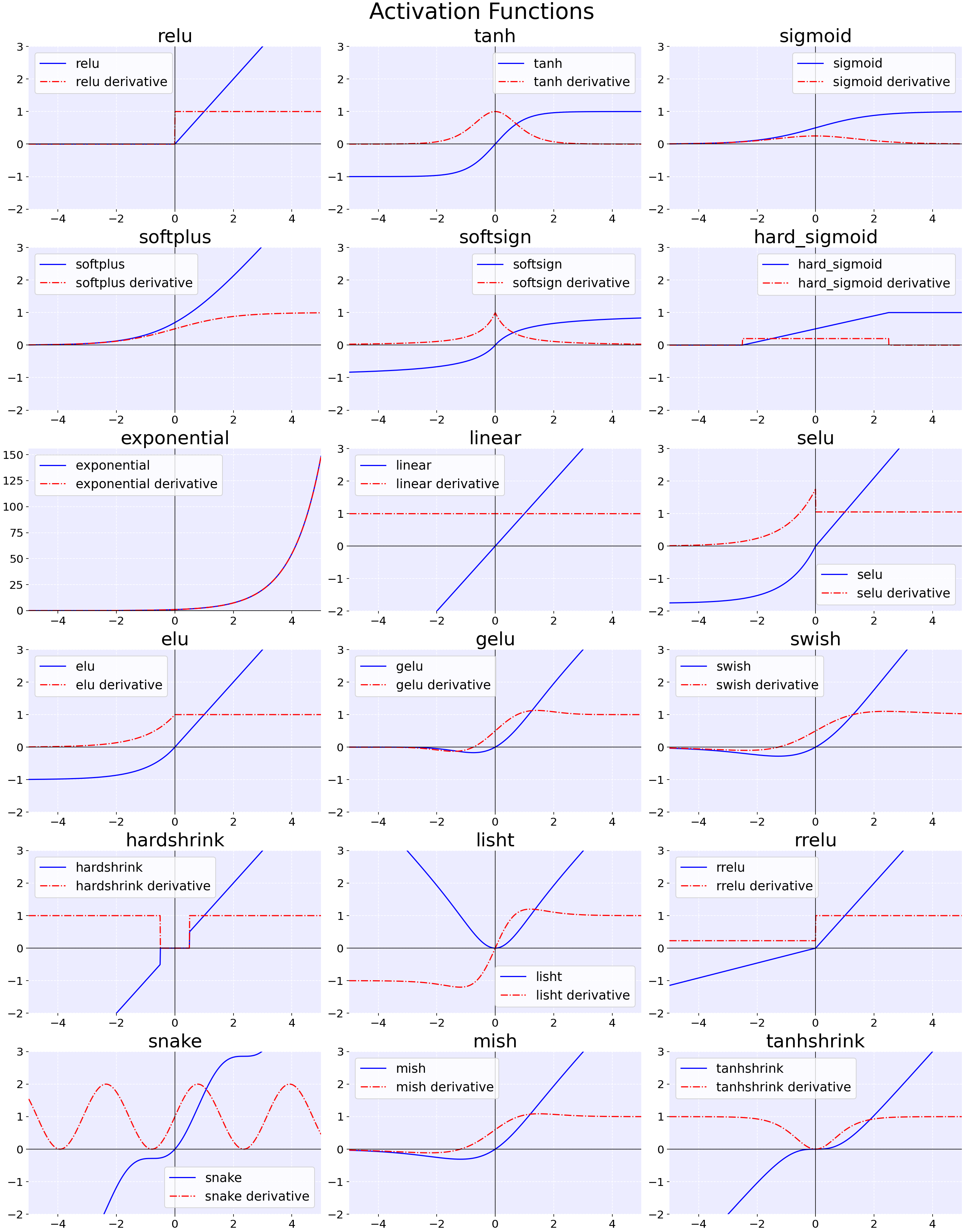}
	\caption{Activation Functions}\label{activation_functions}
\end{figure}

\begin{figure}
\centering
	\includegraphics[width=0.48\textwidth]{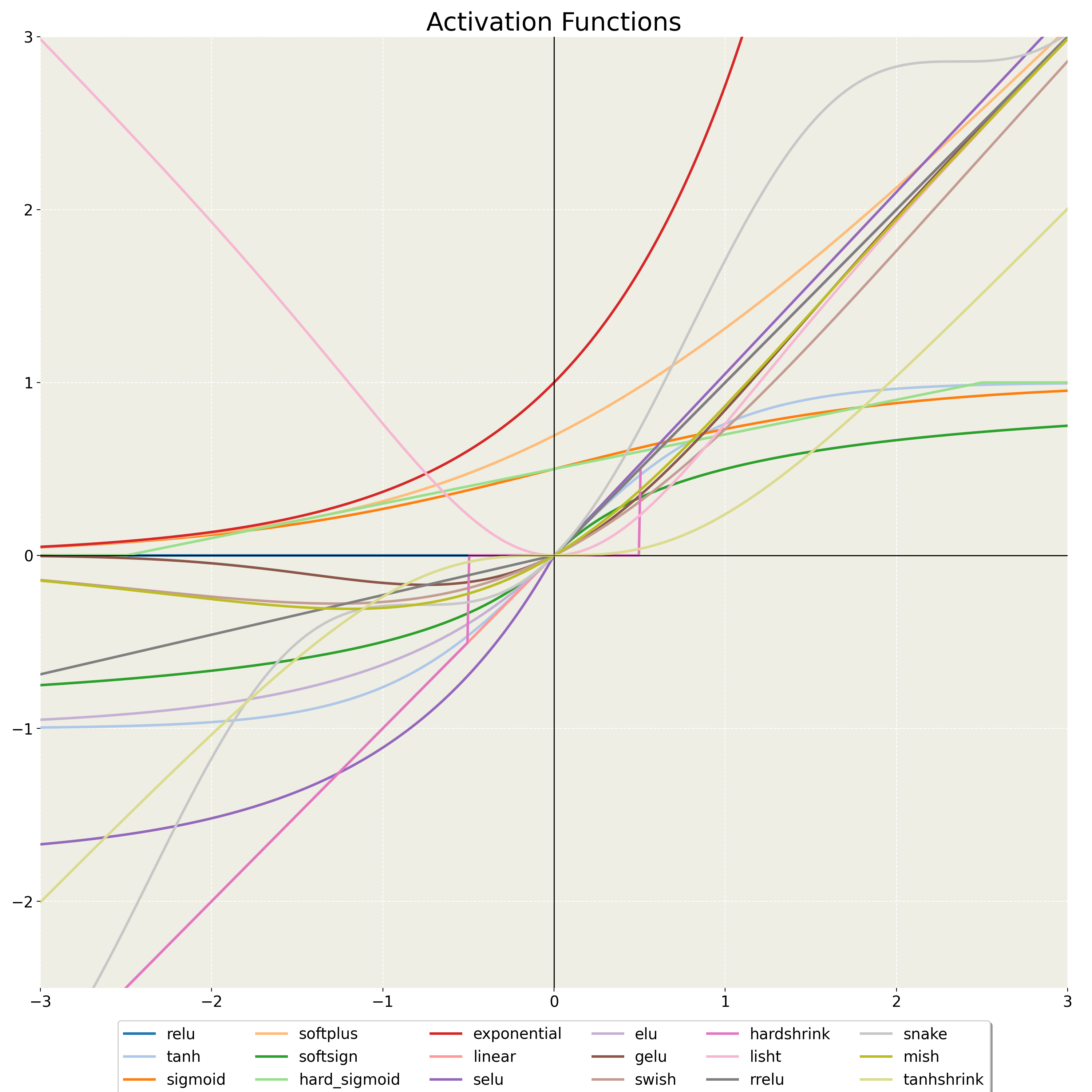}
	\caption{Compare Activation Functions}\label{activation_function_compare}
\end{figure}

In figures~\ref{activation_functions} and \ref{activation_function_compare}, the most common activation functions have been shown.


In conclusion, activation functions play a crucial role in deep learning as they add non-linearity to the model and determine the output of each neuron in a network. The choice of activation function depends on the type of problem being solved and its desired output so there is no ultimate answer for the question "What is the best activation function?". However, it is important to consider the characteristics and limitations of each activation function when designing a deep learning model.

\section{Autonomous Navigation \label{sec:autonomous_navigation}}

Navigation refers to the process of determining the position and orientation of an object and determining a suitable path to reach a destination. It is a crucial aspect of many fields including transportation, robotics, and space exploration. In the last few decades, navigation systems have evolved significantly, and with the advancement of technology, they have become even more sophisticated and reliable.

Navigation systems can be broadly categorized into two main types: \textbf{indoor} and \textbf{outdoor} navigation. \textbf{Indoor navigation} systems are used in indoor environments, such as buildings, malls, and airports. Usually, this type of navigation suffers from the weak signals and other external references. Outdoor navigation systems, on the other hand, are used to navigate in outdoor environments, such as roads, highways, and parks. Despite the availability of GPS signals, outdoor navigation systems are more challenging due to the presence of obstacles and other factors that can affect the accuracy of the system. 

Autonomy refers to the ability of a system or agent to act independently without external control or influence. In the context of engineering and robotics, autonomy often refers to the ability to operate without direct human intervention, making its own decisions and take actions based on its internal programming and external inputs from sensors and other sources. This is the basis of autonomous systems, such as autonomous vehicles, drones, and robots, which are designed to make decisions and perform tasks and  on their own, without the need for human guidance.

Autonomy could be divided into two main approach: (1) Heuristic Approach and (2) Optimal Approach. \textbf{Heuristic approach} is based on rules to make decision which relies on a set of predefined heuristics or "rules of thumb". This approach does not require a lot of computational power, but it may not always lead to an optimal decision. On the other hand, the \textbf{Optimal Approach} involves finding the best decision based on a specific objective functions and constraints, often using mathematical models and algorithms. This approach usually requires more computational power and can be more complex, on the other hand it is often more accurate and efficient. Both approaches have their own advantages and disadvantages, and the choice of approach depends on the specific application and context.

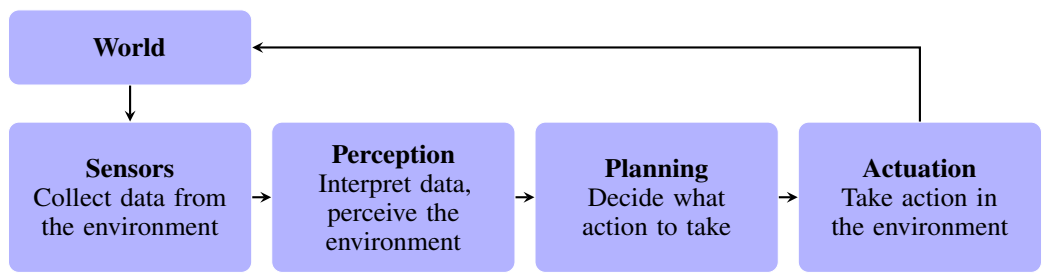
\begin{figure*}[h]
\centering
\begin{center}
	\begin{tikzpicture}[node distance=2cm]
		\node (World) [startstop] {\textbf{World}};
		\node (Sensors) [startstop1, below of=World,anchor=south, align=center, yshift=-1cm] {\textbf{Sensors} \\ Collect data from the environment };
		\node (Perception) [startstop1, right of=Sensors, xshift=1.5cm] {\textbf{Perception} \\ Interpret data, perceive the environment};
		\node (Planning) [startstop1, right of=Perception, xshift=1.5cm] {\textbf{Planning} \\ Decide what action to take};
		\node (Actuation) [startstop1, right of=Planning, xshift=1.5cm] {\textbf{Actuation} \\ Take action in the environment};

		\draw [arrow] (World) -- (Sensors);
		\draw [arrow] (Sensors) -- (Perception);
		\draw [arrow] (Perception) -- (Planning);
		\draw [arrow] (Planning) -- (Actuation);
		\draw [arrow] (Actuation) |- (World);
	\end{tikzpicture}
	\caption{The Autonomous Navigation Cycle} \label{autonomous_navigation_cycle}
\end{center}
\end{figure*}

Autonomous systems needs to interact with physical world by collecting data via sensors. The measurements are used to perceive the environment. These collected data are often noisy and/or incomplete, which makes it difficult for the system to make accurate decisions. To address this, sensor fusion can be used to combine data from multiple sources to improve the overall performance and reduce the uncertainty. For example, a GPS sensor can be used to determine the position of the system, while a camera can be used to detect obstacles and map the the environment. The data from these sensors can be combined to improve the accuracy of the pose estimation algorithm. With these information the agent or system could create a plan to reach its destination by considering the system constraints. Finally, the system could execute the plan by controlling the actuators. The process of collecting data, perceiving the environment, planning, and actuating is known as the \textbf{autonomous navigation cycle} which is shown in figure~\ref{autonomous_navigation_cycle}.

Autonomous navigation systems have become increasingly vital in recent years, especially with the rise of automation and the need for efficient and safe transportation. These systems can be found in a wide range of applications and can be further categorized based on the environment they operate in: \textbf{Land}, \textbf{Marine}, \textbf{Aerial}, \textbf{Space}, \textbf{Underwater}. Each type of system presents unique challenges and requirements, and specific constraints and characteristics of the environment must be take into account when developing these systems.

Regardless of the environment, the main components of autonomous navigation systems include:

\begin{itemize}
	\item \textbf{Obstacle detection} - Detect and avoid obstacles in the environment.
	\item \textbf{Path planning and generation} - Plan a path to reach the destination and generate a trajectory to follow the path.
	\item \textbf{Motion control} - Control the motion of the vehicle to follow the trajectory.
	\item \textbf{Localization} - Determine its position and orientation in the environment.
	\item \textbf{Mapping} - Create a map of the environment.
	\item \textbf{Environment sensing and perception} - Collect and process information about the environment.
	\item \textbf{Decision making and control} - Make decisions and take actions based on the information collected from the environment.
	\item \textbf{Trajectory optimization} - Adjust the trajectory to improve the efficiency and accuracy of the navigation.
	\item \textbf{Sensor fusion} - Combine data from multiple sensors to improve the overall performance of the navigation system.
\end{itemize}

These components work together to ensure that the navigation system can safely and efficiently move from one point to another. In conclusion, navigation is a crucial aspect of many fields, and with the advent of technology, it has become even more sophisticated and reliable. Autonomous navigation systems play a critical role in various applications and have the potential to greatly improve efficiency and safety.

\section{Applications of Deep Learning in Autonomous Navigation \label{sec:applications}}

One of the key applications of deep learning in recent years has been autonomous navigation. Using deep learning to develop advanced and sophisticated navigation systems that can operate without human intervention has revolutionized this field. Deep learning provide a powerful way to learn complex mappings between inputs and outputs, making it possible to perform a wide range of tasks in autonomous navigation, such as pose estimation, perception, decision-making, and control.

One of the main areas where deep learning has been applied is perception systems which are responsible for processing data from a range of sensors, such as cameras, LiDARs, RADARs, and Infrared to obtain information about the surrounding environment. 
\begin{itemize}
	\item \textbf{Camera:} One of the most common sensors used in autonomous navigation. They are used to obtain images of the surrounding environment, which can then be processed by deep learning algorithms to extract relevant information, such as road markings, traffic signs, and obstacles. But they are limited by the visibility conditions, such as low light, fog, and rain.
	\item \textbf{LiDAR:} Used to obtain 3D point clouds of the surrounding environment, which provides 360-degree mapping of the environment but they are limited by low reflective targets.
	\item \textbf{RADAR:} Provide mapping of the environment at medium to long distances which can be used in worse weather conditions. But they are limited by low resolution.
	\item \textbf{Infrared:} Can be used in low light conditions to obtain images of the environment.
\end{itemize}

Deep learning algorithms, particularly convolutional neural networks, have been shown to be highly effective in processing images and extracting relevant features \cite{hu2019novel, matsumoto1990several, arena2003image}, such as road markings \cite{tian2020road,lima2022road}, obstacles \cite{prabhakar2017obstacle,hu2018embedding,lamberti2023bio}, and traffic signs \cite{shustanov2017cnn, dewi2022deep}.Another important area where deep learning has been applied in autonomous navigation is in decision-making which is the ability to make decisions based on the information collected from the environment.

Deep learning has also been applied in the development of control systems for autonomous vehicles which are responsible for controlling the movements, such as steering, acceleration, and braking, to ensure that it follows a safe and efficient trajectory. 

In the following sections, we will discuss the applications of deep learning in autonomous navigation in more detail.

\subsection{Autonomous Drive}

\begin{figure*}
\centering
\begin{center}
	\includegraphics[width=\textwidth]{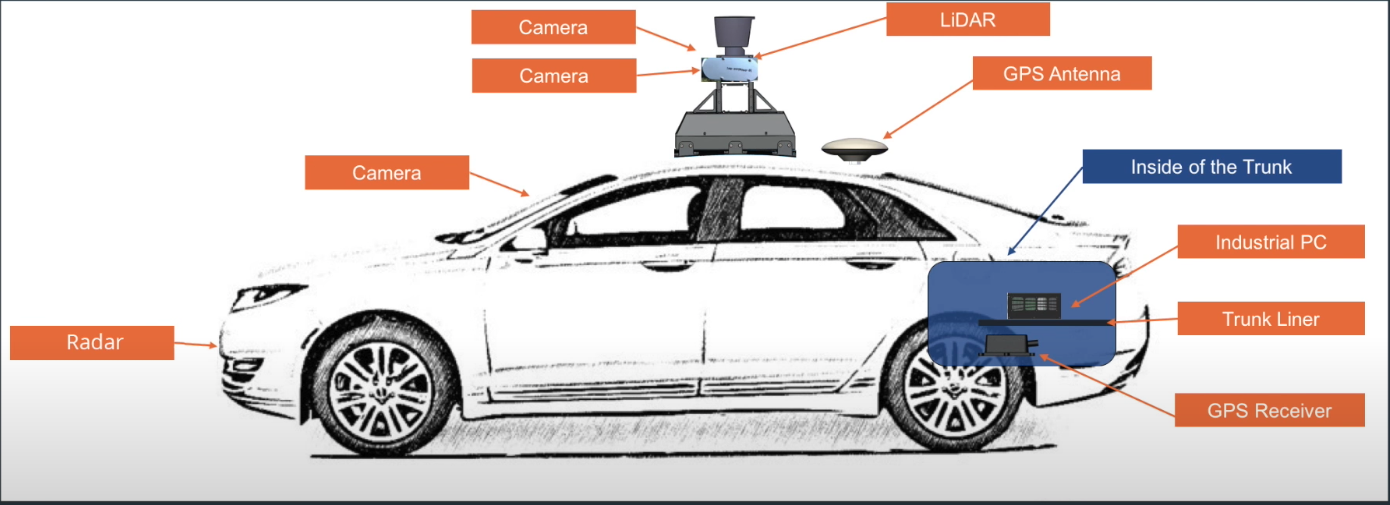}
	\caption{Sensors in Autonomous Vehicles \cite{paravarzar2020motion}} \label{autonomous_vehicle}
\end{center}
\end{figure*}

Self-driving cars are becoming an increasingly popular topic in the field of autonomous navigation. The technology behind these vehicles is being developed by companies such as Tesla, Google, and BMW, among others. Autonomous Driving uses a combination of computer vision, machine learning, and other technologies to enable vehicles to operate without human intervention, Figure \ref{autonomous_vehicle}. The goal of self-driving cars is to make driving safer, more efficient, and more convenient by eliminating human error. The key component of a self-driving car is Artificial intelligence (AI) system that enables it to make decisions and operate autonomously. Deep learning is at the forefront of this technology and is responsible for much of the recent progress in self-driving cars.

In order to make autonomous driving a reality, it is essential to have reliable and robust algorithms that can process large amounts of data from cameras, RADARs, and LiDARs in real-time. Deep learning algorithms, with their ability to learn from data, have proven to be effective in solving these complex challenges.
Deep learning algorithms have been applied to various aspects of autonomous driving, including object detection, lane detection, and trajectory prediction. Object detection is a critical component of autonomous driving, as it involves identifying and classifying objects such as pedestrians, vehicles, and traffic signs. Lane detection is another crucial aspect, as it involves accurately identifying the location of the vehicle within the lane. Trajectory prediction is also important, as it involves predicting the future behavior of other vehicles on the road.

\begin{figure*}
\centering
	\includegraphics[width=0.8\textwidth]{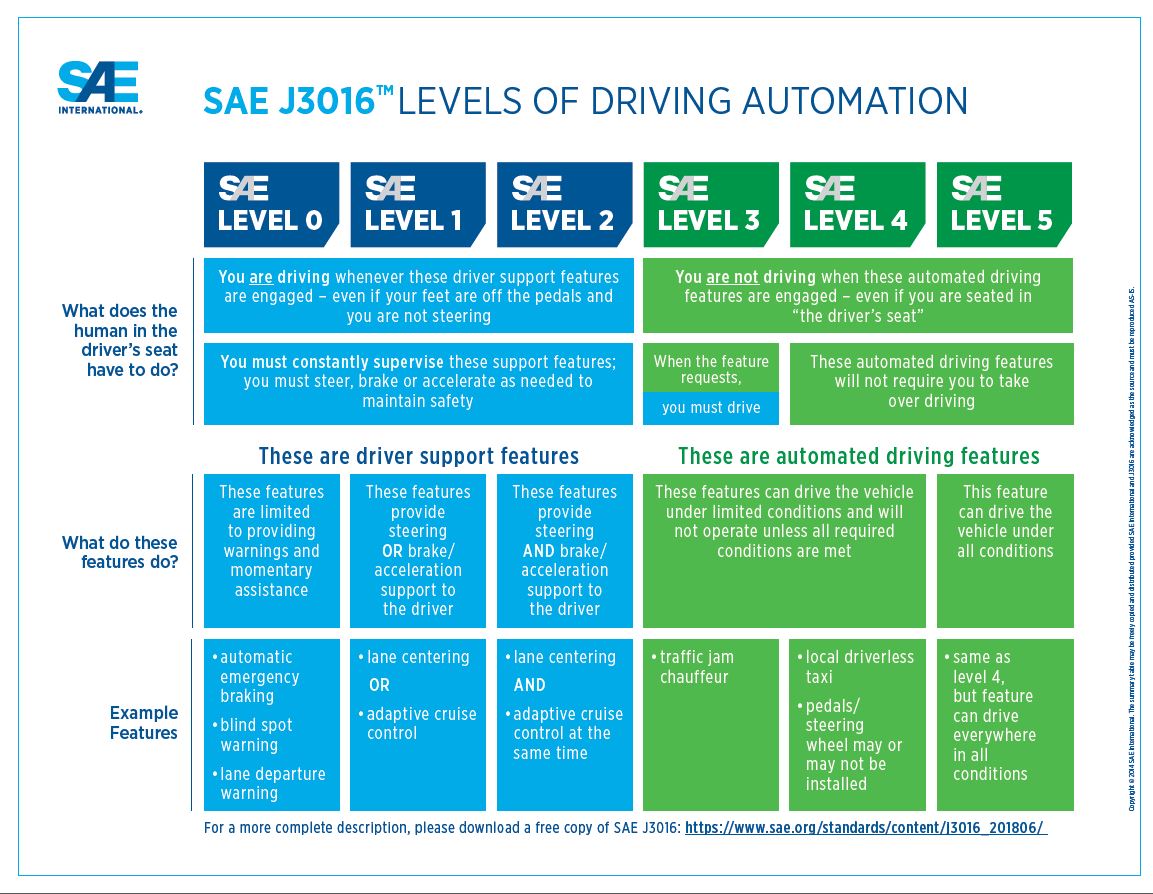}
	\caption{Levels of Automation, described by the Society of Automobile Engineers (SAE) \cite{shuttleworth2019sae}}\label{levels_of_automation}
\end{figure*}

Self-driving cars can reduce the carbon emission levels, risk of drastic accidents, and increase road safety and smoothen traffic flow \cite{fayyad2020deep}. Based on J3016 standard last update in 2019 \cite{shuttleworth2019sae}, there are six levels of driving automation from “no automation” to “full automation”. The first level is “no automation” where the driver is responsible for all driving tasks. The second level is “driver assistance” where the driver is responsible for all driving tasks but the system can assist the driver. The third level is “partial automation” where the driver is responsible for all driving tasks but the system can take over in certain situations. The fourth level is “conditional automation” where the driver is responsible for all driving tasks but the system can take over in certain situations. The fifth level is “high automation” where the driver is responsible for all driving tasks but the system can take over in certain situations. The sixth level is “full automation” where the driver is not responsible for any driving tasks and the system is responsible for all driving tasks. Figure~\ref{levels_of_automation} shows the six levels of driving automation.

Self-driving cars core components are sensors, perception, localization and mapping, path planning, and control. Sensors are the eyes of the car which are responsible for sensing the environment and providing information about the surrounding environment. Perception is the brain of the car which is responsible for processing the data from the sensors and extracting relevant information and used for tasks such as obstacle detection, traffic sign detection, and lane detection. Localization and mapping are responsible for determining the location of the vehicle within the environment, building a map of the environment and tracking the vehicle's trajectory within the map. Planning is the heart of the car that is responsible for determining the best trajectory for the vehicle to follow while avoiding all the obstacles in the environment. Control is the hands of the car to control the movements of the vehicle, such as steering, acceleration, and braking, to ensure that it follows the planned trajectory.

\subsection{Autonomous Flight}
\begin{figure}
\centering
	\includegraphics[width=0.5\textwidth]{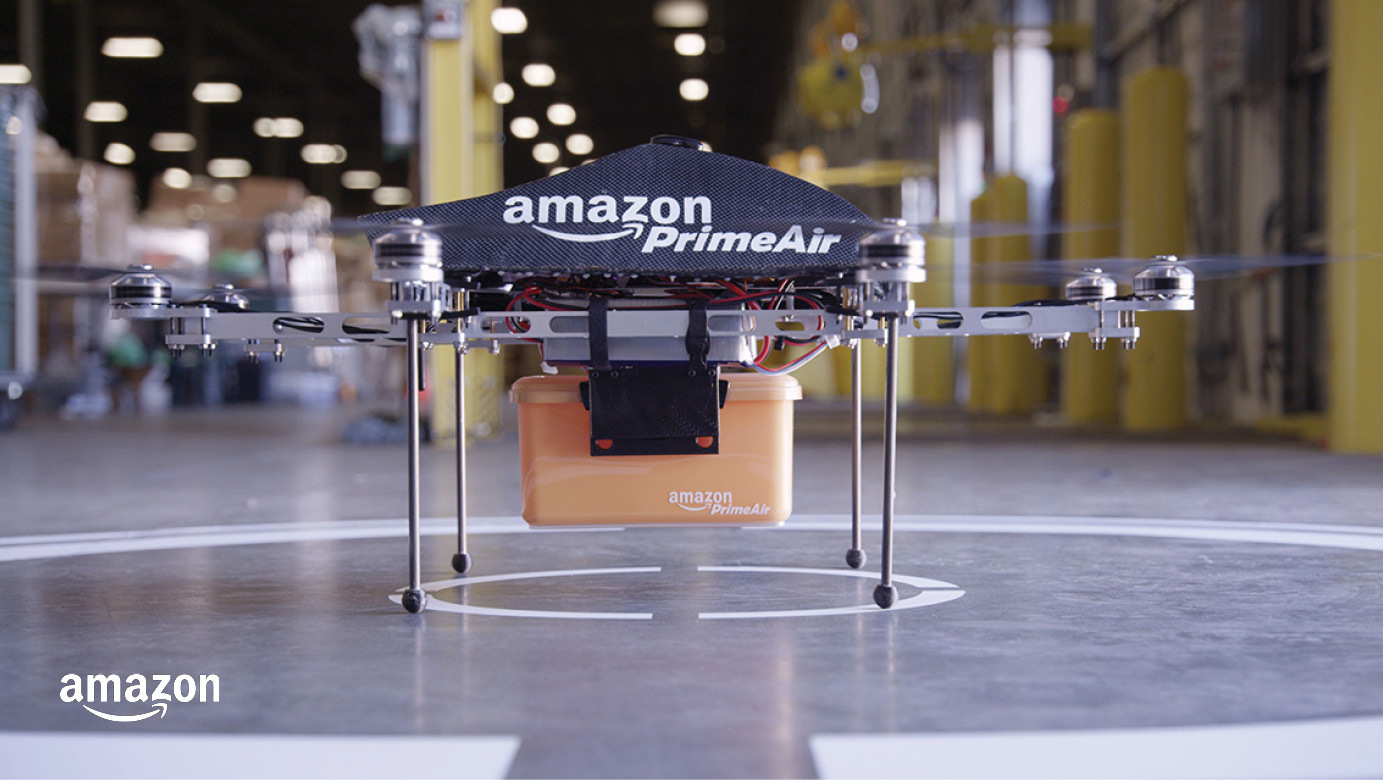}
	\caption{Autonomous Delivery Drone \cite{MK4}}\label{MK4}
\end{figure}

The use of autonomous drones has rapidly grown in recent years and is expected to continue its upward trend \cite{clarke2014understanding, merkert2020revolution,radoglou2020compilation, alzahrani2020uav}. The ability of these vehicles to operate without human intervention opens up a wide range of applications and possibilities. From delivering packages (Figure~\ref{MK4}) to performing complex aerial photography and remote sensing, autonomous flying vehicles are poised to revolutionize various industries and make our lives easier and more convenient \cite{almahamid2022autonomous}.

One of the primary goals of autonomous flying vehicles is to improve safety by reducing the risk of human error. This is especially important in hazardous scenarios such as search and rescue missions, where a misstep by a human pilot could be catastrophic. With autonomous flying vehicles, the risk of human error is greatly reduced, making these missions much safer \cite{mohsan2022towards}.

The backbone of autonomous flying vehicles is their Artificial Intelligence system that enables them to make decisions and operate autonomously. The AI system is responsible for processing vast amounts of data obtained from sensors, such as cameras and LiDARs, to gain an understanding of the surrounding environment. Using this information, the AI system then makes decisions about the flying actions of the vehicle which includes obstacle avoidance, autonomous movement, and even autonomous take-off and landing. Deep learning is playing a major role in the development of autonomous drones and has been instrumental in advancing the technology. Deep learning algorithms have been used to develop perception and decision-making systems that enable autonomous flying vehicles to perform complex tasks with high accuracy. For example, deep learning algorithms can be trained to recognize and classify objects in real-time, enabling autonomous flying vehicles to identify obstacles and avoid them \cite{rezwan2022artificial}.

The SAE's Six levels of autonomy, ranging from "no automation" to "full automation," were originally developed for autonomous ground vehicles but are applicable to any vehicle capable of autonomy, including drones. The first level is “no automation” where the pilot is responsible for all flying tasks. Our analysis maps the functionality of drone navigation against these levels, starting at Level 1 with assisted features like GPS guidance and landing zone evaluation, which are already available in commercially available drones. Level 2, known as "pilot assistance," includes specific and use case-dependent navigational operations, such as "follow me" and "track target" commands. Level 3, or "partial automation," allows for autonomous navigation in certain identified environments. At Level 4, or "conditional automation," the drone can navigate autonomously without human interaction in most use cases. Finally, Level 5, or "full automation," is the theoretical ideal of autonomy in all possible use cases, environments, and conditions which indicating that the system is responsible for all flying tasks \cite{lee2021flying}. Table ~\ref{levels_of_automation_flying} shows the six levels of flying automation and the tasks that are performed at each level.

\begin{table}[h]
\caption{Levels of Automation for Flying Vehicles}\label{levels_of_automation_flying}
\begin{tabular*}{\linewidth}{ p{3.5cm}llllll}
\toprule
\textbf{Level} & \textbf{ 0} & \textbf{ 1} & \textbf{ 2} & \textbf{ 3} & \textbf{ 4} & \textbf{ 5} \\ 
\midrule
\textbf{Obstacle Detection} & & \checkmark & \checkmark & \checkmark & \checkmark & \checkmark \\ \hline
\textbf{Adaptive Cruise Control} & & \checkmark & \checkmark & \checkmark & \checkmark & \checkmark \\\hline
\textbf{Spatial Awareness} & & \checkmark & \checkmark & \checkmark & \checkmark & \checkmark \\\hline
\textbf{Object Distinction} & & & \checkmark & \checkmark & \checkmark & \checkmark \\\hline
\textbf{Obstacle Avoidance} & & & \checkmark & \checkmark & \checkmark & \checkmark \\\hline
\textbf{Autonomous Movement} & & & \checkmark & \checkmark & \checkmark & \checkmark \\\hline
\textbf{Environmental Awareness} & & & & \checkmark & \checkmark & \checkmark \\\hline
\textbf{Non-planar Movement} & & & & \checkmark & \checkmark & \checkmark \\\hline
\textbf{Autonomous \newline Take off \& Landing} & & & &  & \checkmark & \checkmark \\\hline
\textbf{Path Generation} & & & &  & \checkmark & \checkmark \\\hline
\textbf{Autonomous Flight} & & & &  & & \checkmark \\
\bottomrule
\end{tabular*}
\end{table}

\subsection{Autonomous Space Flight}
In the past decades AI has made significant contributions to the space industry, especially in the field of Guidance, Navigation, and Control. Notably, deep learning algorithms have facilitated the development of sophisticated pose estimation \cite{proencca2020deep, park2022robust}, path planning \cite{santos2022machine}, and collision avoidance \cite{uriot2022spacecraft} systems for autonomous spacecraft, such as satellites and planetary rovers. These AI-powered navigation systems have enabled spacecraft to operate autonomously, leveraging real-time data from onboard sensors and cameras to make informed decisions without human intervention. The utilization of AI in spacecraft navigation has enhanced the efficiency and precision of spacecraft operations, thereby enabling exploration of previously inaccessible domains in space. As space exploration continues to evolve, AI is expected to play an increasingly critical role in spacecraft navigation and other space-related applications.

Furthermore, AI has addressed several key challenges in space missions, such as low earth orbit attitude determination \cite{sun2023satellite}, simultaneous navigation and characterization \cite{stacey2022robust}, pose tracking \cite{park2022adaptive}, planetary landing \cite{gaudet2020deep, furfaro2018deep}, crater detection \cite{downes2021neural, downes2020deep}, deep space missions \cite{tsukamoto2022neural,donitz2022interstellar}, gravity field modeling \cite{neamati2022learning}, motion planning \cite{nakka2021spacecraft}

AI-based autonomous systems can process and interpret data much faster than traditional methods, enabling spacecraft to react quickly to dynamic situations, such as the detection of a hazard or a change in the terrain \cite{moghe2020line}. Additionally, autonomous spacecraft navigation systems are less susceptible to human error and can operate in hazardous environments, which is particularly useful in missions to explore unknown territories in space \cite{song2022deep}. Table ~\ref{ai_spacecraft} summarizes the recent research on the main applications of AI in autonomous spacecraft.

\begin{table*}
\caption{Applications of AI in Autonomous Spacecraft}\label{ai_spacecraft}
\begin{tabular*}{\linewidth}{l l l p{8.3cm}}
\toprule
\textbf{Ref} & \textbf{Application} & \textbf{Method} & \textbf{Description} \\
\midrule 
\cite{scorsoglio2023relative} & Relative Motion Guidance & Reinforcement Learning & An actor-critical reinforcement learning algorithm based on lightweight feedback guidance was proposed for proximity operations in the cislunar environment. \\ \hline

\cite{guthrie2022image} & Relative Motion Guidance & Deep learning & Convolutional Siamese network used to estimate the attitude and rotational state of an uncooperative debris satellite with unknown geometry \\\hline
\cite{proencca2020deep} & Pose Estimation & Deep Learning & A CNN pose estimation of known uncooperative spacecrafts approach build upon SPEED dataset and presented  present a simulator called URSO built on Unreal Engine 4. \\\hline
\cite{abd2022reliable} & Fault Detection & Deep Learning & A one-dimensional convolutional neural network (1D-CNN) coupled with an LSTM network architecture is used for detecting and identifying anomalies in spacecraft reaction wheels. \\\hline
\cite{spantideas2021deep} & Magnetic Dipole modeling & Deep Learning & MDMnet, a deep-learning neural network, models the magnetic signature of spacecraft units using synthetic magnetic flux density data generated by virtual dipole sources and achieves high predictive accuracy. \\\hline
\cite{aldahoul2022rgb} & Debris Detection & Deep Learning & A multi-modal learning approach is employed to detect space objects, including spacecraft and debris which integrates a vision transformer based on RGB and CNN based network to classify 11 object categories. \\\hline
\cite{harris2022generation} & Procedure Generation & Reinforcement Learning & Examined the feasibility and safety of deep reinforcement learning for spacecraft decision-making, and the hybridization of DRL-driven policy generation algorithms with correct-by-construction control techniques. \\\hline
\cite{del2022deep} & Navigation & Deep Learning & The Cascade Mask R-CNN was used for detecting craters from monocular images for autonomous spacecraft navigation, and for learning transfer functions from real lunar images acquired by the lunar reconnaissance orbiter. \\\hline
\cite{siew2022space} & Sensor Tracking & Reinforcement Learning & A deep reinforcement learning agent to task a space-based narrow-FOV sensor in low Earth orbit for space situational awareness \\\hline
\cite{siew2022optimal} & Sensor Tracking & Reinforcement Learning & Detect, characterize, and track resident space objects in increasingly complex environments using ground-based optical telescopes. \\\hline
\cite{oestreich2021autonomous} & Control & Reinforcement Learning & Optimized proximal policies for 6 DoF closed-loop control for cooperating clients in spacecraft docking missions \\ 
\bottomrule
\end{tabular*}
\end{table*}

Despite the considerable progress made in the field of AI-enabled spacecraft navigation, numerous challenges remain, which require further research and development. A major challenge is the limited computational resources available in space, which means that algorithms must be optimized for both accuracy and efficiency. Moreover, since spacecraft operate in unpredictable and demanding conditions, the reliability and robustness of AI-enabled systems must be thoroughly tested and verified to ensure mission success. Therefore, the development of reliable and efficient AI-enabled spacecraft navigation systems requires a multifaceted approach incorporating cutting-edge research in artificial intelligence, rigorous engineering design, testing and validation.

In summary, the implementation of AI technology in autonomous spacecraft has represented a paradigm shift in the space industry by providing the ability to operate autonomously, leveraging sensory data to make real-time decisions. As the field of space exploration continues to progress, it is evident that AI technology will become increasingly indispensable for space-related applications, thus paving the way for humanity to explore new and unprecedented frontiers in space. Indeed, the continued development of AI-enabled spacecraft navigation systems promises to revolutionize the space industry and usher in a new era of space exploration, with boundless potential for discovery and innovation.

Table~\ref{recent_ai_spacecraft} summarizes recent research in AI-enabled space missions. The table includes 

\begin{table*}[h]
	\caption{Recent Research in AI-enabled Space Missions}\label{recent_ai_spacecraft}
	\begin{tabular*}{\linewidth}{l l l l l}
	\toprule
	\textbf{Ref} & \textbf{Method} & \textbf{Backbone} & \textbf{Dataset} & \textbf{Task} \\
	\midrule
	\cite{mahendrakar2023spaceyolo} & Supervised & YOLO & Synthetic & Fault Detection \\
	\cite{park2019towards} & Supervised & YOLO + MobileNet & SPEED & Pose Estimation \\
	\cite{chen2019satellite} & Supervised & Faster R-CNN + HRNet & SPEED & Pose Estimation \\
	\cite{scorsoglio2020safe} & Reinforcement Learning & Deep Reinforcement Learning & \xmark & Landing \\
	\cite{harris2022generation} & Reinforcement Learning & Deep Reinforcement Learning & \xmark & Decision Making \\ 
	\cite{elkins2020autonomous} & Reinforcement Learning & Deep Reinforcement Learning & \xmark & Control \\
	\cite{singh2022stochastic} & Supervised &  Gaussian Process Regression & Synthetic & Control \\
	\cite{hao2021intelligent} & Supervised & ResNet-50  & OrViS & Pose Estimation \\
	\cite{yun2020multi}& Supervised & Dense & Synthetic & Motion Planning \\
	\cite{zhou20213d} & Supervised & Track3D & Synthetic & Tracking \\
	\cite{huan2020pose} & Supervised & CNN & Synthetic & Pose Estimation \\
	\cite{oestreich2021autonomous} & Reinforcement Learning & Deep Reinforcement Learning & \xmark & Control \\
	\cite{roberts2021deep} & Reinforcement Learning & Deep Reinforcement Learning & \xmark & Space Situational Awareness \\
	\cite{roberts2021geosynchronous} & Supervised & CNN & Synthetic & Maneuver Classification \\ 
	\cite{siew2022optimal} & Reinforcement Learning & Deep Reinforcement Learning & \xmark & Space Situational Awareness \\
	\cite{abd2022reliable} & Supervised & CNN-LSTM & Synthetic & Fault Detection \\
	\cite{liu2022multi} & Supervised & DeepLabv3+ & SPEED + URSO &  Image Segmentation \\
	\cite{siew2022space} & Reinforcement Learning & Deep Reinforcement Learning & \xmark & Space Situational Awareness \\
	\cite{rondao2022chinet} & Supervised & RNN-CNN & Synthetic & Pose Estimation \\
	\cite{aldahoul2022rgb} & Supervised & CNN & SPARK & Space Situational Awareness \\
	\cite{liu2023spacecraft} & Supervised & TCN & SMAP/MSL & Anomaly Detection \\
	\cite{scorsoglio2023relative} & Reinforcement Learning & Deep Reinforcement Learning & \xmark & Relative Motion Guidance \\
	\bottomrule
\end{tabular*}
\end{table*}

\section{Deep Learning Components in Autonomous Navigation \label{sec:deep_learning_components}}

\subsection{Signal Processing}
A key component of autonomous navigation is signal processing, which generates and processes sensory data so that relevant information about the environment can be extracted. Signal processing is used to improve the quality of sensory data, which in turn improves the accuracy of the decision-making process of autonomous systems. Filtering, segmentation, feature extraction, classification, and registration are some of the signal processing techniques used in autonomous navigation systems. The use of deep learning techniques for signal processing has improved the performance of autonomous navigation systems in a variety of areas, including image segmentation and data denoising.

Furthermore, deep learning techniques have been utilized for multi sensor data fusion, which involves combining data from multiple sensors to reduce the uncertainty and obtain a more comprehensive representation of the environment. The fusion of data from different sensors, such as RADAR, camera and LiDAR, has been shown to improve the accuracy of object detection and classification. The use of deep neural networks in sensor fusion has enabled the system to learn more complex representations of the environment and make more accurate decisions.

Signal denoising is a fundamental task in signal processing and has wide applications in various domains such as inertial navigation, attitude estimation, image processing, and biomedical signal analysis. Traditional methods for denoising such as wavelet transforms and filtering techniques have limitations in effectively removing noise while preserving the signal's fidelity. Recently, deep learning techniques have been applied to denoise signals, and shown promising results due to their ability to learn complex features and patterns directly from raw data. 

Tian et. al. \cite{tian2020deep} conducted a comparative investigation of deep learning techniques for the purpose of image denoising. The study aimed to evaluate the effectiveness of various deep models in addressing noise-related challenges in image processing. The authors classified deep CNNs for four categories of noisy tasks, including additive white noisy images, blind denoising, real noisy images, and hybrid noisy images, and analyzed the motivations and principles of the different types of these methods. The authors also compared the state-of-the-art methods on public denoising datasets using quantitative and qualitative analysis. However, the authors noted some potential challenges and directions of future research, including improving hardware devices to better suppress noise and effectively recovering the latent clean image from the superposed noisy image, which are urgent challenges that researchers and scholars need to address.

For instance, \cite{jiang2018mems} used LSTM network to filter MEMS-based gyroscope outputs to improve the accuracy of standalone INS. The results indicate that the denoising scheme was effective for improving MEMS INS accuracy, and the proposed LSTM method was more preferable in this application. This study also compares LSTM with Auto Regressive and Moving Average (ARMA) models with fixed parameters, and the LSTM method performed better in this application. Similarly \cite{brossard2020denoising} demonstrated that CNNs can denoise IMU measurements to achieve remarkable accuracy in attitude estimation, by filtering out noise caused by sensor drift, environmental disturbances, and dynamic effects .

Four hybrid models consist of LSTM-LSTM, GRU-GRU, and LSTM-GRU LSTM-LSTM, GRU-GRU, and LSTM-GRU, are proposed in \cite{han2021hybrid}, and static and dynamic experiments were conducted to validate the effectiveness of the proposed networks. The results show that LSTM-GRU network has the best noise reduction effect, while GRU-LSTM network is overfitting the large amount of data. The study demonstrates the potential of using deep learning algorithms in MEMS gyroscope noise reduction and improving the accuracy of MEMS-IMU systems.

Chen et al.\cite{chen2022towards} utilized CNN to reduce noises from IMU sensors in autonomous vehicles. The proposed system does not need external measuring sources such as GPS or vehicle sensors, unlike EKF approaches, and the method can be applied for both high-grade and low-grade IMUs. Three CNNs were developed, and the results showed up to 32.5\% error improvement in straight-path motion and up to 38.69\% error improvement in oval motion compared with the ground truth. The authors' primary objective was to develop algorithms with higher performance and lower complexity that would allow implementation on ultra-low power artificial intelligence microcontrollers such as the Analog Devices MAX78000. Similarly, RNNs \cite{engelsman2022data} have been used to remove noise from accelerometer measurements, 

Further, in \cite{yuan2023simple}, authors suggested a novel approach for IMU denoising called IMUDB using self-supervised learning and future-aware inference techniques . The method was evaluated using end-to-end navigation on two benchmark datasets, EuRoC and TUM-VI, and demonstrated promising results. The proposed method includes three self-supervised tasks and an uncertainty-aware inference method. The article highlights the efficiency and superiority of the proposed method over competing methods, as demonstrated through experiments on different datasets. Also, they identifies the limitations of the proposed method, such as the sequence size and hyper-parameters, and suggests future works to address them. 

In \cite{hu2021novel} an end-to-end learning framework developed for denoising LiDAR signals. The method utilizes the encoding and decoding characteristics of the autoencoder to construct convolutional neural networks to extract deep features of LiDAR return signals. The method was compared with wavelet threshold and VMD methods and found to have significantly better denoising effects. It has strong adaptive ability and has the potential to eliminate complex noise in lidar signals while retaining the complete details of the signal. The article concludes that the method is feasible and practical and can be further improved by adding more data for training the network. Additionally, DLC-SLAM \cite{liu2023dlc} presented to address the issue of low accuracy and limited robustness of current LiDAR-SLAM systems in complex environments with many noises and outliers caused by reflective materials. The proposed system includes a lightweight network for large-scale point cloud denoising and an efficient loop closure network for place recognition in global optimization. The authors conducted extensive experiments and benchmark studies, which showed that the proposed system outperformed current LiDAR-SLAM systems in terms of localization accuracy without significantly increasing computational cost. The article concludes by providing open-source code to the community.

Deep learning approaches have the advantage of learning complex noise patterns, which can vary depending on the input signal's type and nature. As a result, deep learning-based denoising techniques have the potential to optimize denoising performance for various applications, thus enhancing the accuracy and reliability of autonomous navigation systems. In addition, signal processing, including tasks such as filtering, segmentation, feature extraction, classification, registration, and sensor data fusion, plays a crucial role in the development of autonomous navigation systems. Deep learning has demonstrated significant improvements in the accuracy and speed of signal processing tasks. Furthermore, deep learning has been applied to denoise signals with promising results, improving the accuracy and reliability of sensor data for various applications, including inertial navigation, attitude estimation, and image processing. The ability of deep learning algorithms to learn complex noise patterns makes them an attractive option for enhancing the accuracy and reliability of autonomous navigation systems.

Numerous research has been conducted on the utilization of deep learning techniques for denoising various types of data, such as images \cite{pang2021recorrupted, cui2019pet}, electrocardiogram (ECG) signals \cite{arsene2019deep, nurmaini2020deep}, wind speed data \cite{peng2020novel}, sound \cite{zhang2023birdsoundsdenoising, xu2020listening}, and magnetic resonance imaging (MRI) \cite{tian2022sdndti}. However, there is a lack of research investigating the benefits of employing these techniques to enhance autonomous navigation.

The use of deep learning for denoising data in autonomous navigation is a promising area of research. By effectively removing noise from data collected by sensors such as IMU and radar, the accuracy and reliability of autonomous navigation systems can be greatly improved. For instance, in inertial-based navigation systems, the denoising of inertial data can enhance the accuracy of the navigation algorithm and reduce the bias and drift in the output. Similarly, the denoising of radar data can improve the performance of radar-based obstacle detection and tracking.Further research in the area of data denoising and preprocessing is needed to fully realize the benefits of these techniques for autonomous navigation.

\subsection{Perception} 

Perception is a critical aspect of robotics and engineering, which involves the ability to interpret sensory data to understand and make sense of the surrounding environment. In the context of autonomous navigation systems, perception is a fundamental element enables the system to detect and analyze the physical world around it, make decisions based on that information, and take actions accordingly. A more detailed discussion can be found in \cite{thrun2002probabilistic}.

Perception is a complex process which involves multiple stages. In the first stage the system takes in data from various sources, including camera, LiDAR, RADAR, and inertial sensors. The second stage involves data preprocessing, where the raw sensory data is filtered and processed to extract meaningful information and features. The final stage is interpretation, where the system uses the extracted information and features to make sense and perceive the environment and take appropriate action.

In recent years, deep learning algorithms have been extensively used for perception tasks in various systems. These algorithms have proven to be highly effective for tasks such as object detection, object recognition, and scene understanding, which are crucial for navigation and decision making.

As the system trained on vast amounts of data, they have the ability to learn and improve over time. This will enables them to recognize and classify object and obstacles with high accuracy. As the system continues to operate, it can refine and improve its perception abilities through continuous learning. Some of the datasets listed in Table~\ref{datasets}

\begin{table*}[h]
\caption{Some of the most popular publicly available datasets for autonomous navigation}
\label{datasets}
\begin{tabular*}{\linewidth}{p{3.5cm} l l l p{2.5cm}}
\toprule
\textbf{Dataset} & \textbf{Year} & \textbf{Sensor} & \textbf{Scene} & \textbf{Application} \\
\midrule
KITTI \cite{geiger2012we} & 2012 & Camera + LiDAR + IMU & Urban & Ground Vehicle \\
EuRoC MAV \cite{burri2016euroc} & 2014 & Cameras + IMU & Indoor & Drone \\
NCLT \cite{carlevaris2016university} & 2015 & Camera + LiDAR + IMU + GPS + Wheel & Urban & Ground Vehicle \\
SUN RGB-D \cite{song2015sun} & 2015 & Camera & Indoor & Scene Understanding \\
Cityscapes \cite{cordts2016cityscapes} & 2016 & Camera + GPS & Urban & Ground Vehicle \\
LISA Traffic Light \cite{jensen2016vision} & 2016 & Camera & Urban & Traffic light detection \\
Stanford Drone \cite{robicquet2016learning} & 2016 & Camera & Outdoor & UAV + Trajectory Prediction \\
UAV123 dataset \cite{mueller2016benchmark} & 2016 & Camera  & Indoor & Drone \\
Zurich Urban MAV \cite{majdik2017zurich} & 2017 & Camera + GPS + IMU & Outdoor & Drone \\
Oxford RobotCar \cite{maddern20171} & 2017 & Camera + LiDAR + RADAR + GPS & Urban & Ground Vehicle \\
DAVIS Dataset \cite{perazzi2016benchmark} & 2017 & Camera & -- & Object Tracking \\
BDD100K \cite{yu2020bdd100k} & 2017 & Camera + GPS + IMU & Urban & Ground Vehicle \\
ApolloScape \cite{huang2018apolloscape} & 2018 & Camera + LiDAR + GPS + IMU & Urban & Ground Vehicle \\
UPenn Fast Flight \cite{sun2018robust} & 2018 & Camera + GPS + IMU & Outdoor & Drone \\
ScanNetV2 \cite{dai2017scannet} & 2018 & Camera + Mesh & Indoor & Scene Understanding \\
KAIST \cite{choi2018kaist} & 2018 & Camera + LiDAR + GPS + IMU & Urban & Ground Vehicle \\
UZH-FPV \cite{Delmerico19icra} & 2019 & Camera + IMU & Indoor + Outdoor & Drone \\
Argoverse \cite{chang2019argoverse} & 2019 & Camera + LiDAR + GPS & Urban & Ground Vehicle \\
H3D \cite{patil2019h3d} & 2019 & Camera + LiDAR + GPS + IMU & Urban & Ground Vehicle \\
Lyft \cite{kesten2019lyft} & 2019 & Camera + LiDAR & Urban & Ground Vehicle \\
A2D2 \cite{geyer2020a2d2} & 2019 & Camera + LiDAR & Urban & Ground Vehicle \\
A*3D \cite{pham20203d} &  2019 & Camera + LiDAR & Urban & Ground Vehicle \\
Carla Simulator \cite{deschaud2021kitti} & 2019 & Camera + LiDAR & Simulation & Ground Vehicle \\
nuScenes \cite{caesar2020nuscenes} & 2019 & Camera + LiDAR + RADAR + GPS + IMU & Urban & Ground Vehicle \\
Waymo Open \cite{sun2020scalability} & 2020 & Camera + LiDAR & Urban & Ground Vehicle \\
Blackbird \cite{antonini2020blackbird} & 2020 & Camera + IMU & Indoor + Outdoor & Drone \\
VisDrone dataset \cite{zhu2021detection} & 2021 & Camera & Outdoor& Drone \\
WHUVID \cite{chen2022whuvid} & 2022 & Camera + IMU & Urban & Ground Vehicle \\
HDIN \cite{chang2022hdin} & 2022 & Camera & Indoor & Drone \\
\bottomrule
\end{tabular*}
\end{table*}

Despite the many benefits of deep learning-based perception systems, there are still some challenges that need to be addressed. One significant challenge is the potential for bias in the training data, which can result in inaccuracies in perception. Additionally, these systems can struggle in complex and dynamic environments, where there is a high degree of uncertainty and unpredictability. Also, these systems may require a high computation power.

In this section, we discuss the use of deep learning algorithms for perception in autonomous navigation systems. 

\subsubsection{Object Detection}

Object detection is the process of identifying an object's presence and class in an image or video frame and determining its location. This technique is widely used in computer vision applications such as robotics, surveillance, and autonomous navigation, and involves creating bounding boxes around objects and labeling them with class labels \cite{zou2023object}. Table~\ref{object_detection} provides an overview of the most popular object detection algorithms.

\begin{table*}[h]
\caption{Object Detection Algorithms}\label{object_detection}
\begin{tabular*}{\linewidth}{l p{14.5cm}}
	\toprule
\textbf{Algorithm} & \textbf{Description} \\
 \midrule
VJ Det. \cite{viola2004robust} & Viola-Jones detector, a real-time face detection algorithm \\
HOG Det. \cite{dalal2005histograms} & Histogram of Oriented Gradients, used to detect objects in images \\
DPM \cite{felzenszwalb2008discriminatively} & Deformable Parts Model, used for detecting objects with a combination of local parts \\
RCNN \cite{girshick2014rich} & Region-based Convolutional Neural Networks, uses selective search for region proposals and a CNN to classify objects \\
SPPNet \cite{he2015spatial} & Spatial Pyramid Pooling Network, a CNN architecture that can handle images of arbitrary size and aspect ratio \\
Fast RCNN \cite{girshick2015fast} & Faster Region-based Convolutional Neural Networks, an improvement of RCNN that shares computation among the region proposals \\
Faster RCNN \cite{ren2015faster} & A faster version of Fast RCNN that uses a Region Proposal Network to generate region proposals \\
YOLOv1 \cite{redmon2016you} & You Only Look Once version 1, a single-shot object detection model that predicts the bounding boxes and class probabilities directly from the image \\
YOLO9000 \cite{redmon2017yolo9000} & You Only Look Once version 2, an improved version of YOLOv1 that uses anchor boxes to improve the accuracy of object detection \\
YOLOv3 \cite{redmon2018yolov3} & You Only Look Once version 3, an even more accurate version of YOLO that uses feature maps to predict object classes and bounding boxes \\
YOLOv4 \cite{bochkovskiy2020yolov4} & You Only Look Once version 4, the latest version of YOLO that uses several new techniques, including spatial pyramid pooling, path aggregation networks, and cross-stage partial connections \\
SSD \cite{liu2016ssd} & Single Shot MultiBox Detector, a single-shot object detection model that uses a series of convolutional layers to predict object classes and bounding boxes \\
FPN \cite{lin2017feature} & Feature Pyramid Network, a CNN architecture that can handle images of different scales and resolutions by constructing a feature pyramid \\
RetinaNet \cite{lin2017focal} & A single-shot object detection model that uses a feature pyramid network and a focal loss function to handle the class imbalance problem \\
CornerNet \cite{law2018cornernet} & A keypoint-based object detection model that uses corners as keypoints to detect objects and their orientations \\
CenterNet \cite{zhou2019objects} & A keypoint-based object detection model that predicts the center point of objects and their size and orientation \\
DETR \cite{carion2020end} & Detection Transformer, a transformer-based object detection model that uses an attention mechanism to directly predict the set of objects in an image \\
Cascade R-CNN \cite{cai2018cascade} & A variant of Faster R-CNN that uses a cascade of detectors to improve accuracy.\\
EfficientDet \cite{tan2020efficientdet} & A family of object detection models that use EfficientNet as a backbone and incorporate various design changes to improve speed and accuracy.\\
Sparse R-CNN \cite{sun2021sparse} & A variant of Faster R-CNN that uses sparse convolutional layers to reduce computation and memory requirements.\\
RepPoints \cite{yang2019reppoints} & A method that represents object instances as a set of points and performs detection by matching these points to features in the image.\\
FCOS \cite{tian2019fcos} & A fully convolutional object detection method that predicts the bounding box, class, and objectness score for each pixel in the feature map.\\
\bottomrule
\end{tabular*}
\end{table*}

\subsubsection{Obstacle Detection}

Obstacle detection is a critical component and core technical problem in autonomous navigation systems, as it involves identifying and classifying static and dynamics objects such as vehicles, pedestrians, and other obstacles that may be present in the vehicle's path. In the past decades, various sensors have been utilized and developed for this purpose, including cameras, ultrasonic sensors, LiDARs, and RADARs. However, the most common sensor used for obstacle detection is the camera, as it is the most cost-effective and widely available sensor which are also capable of capturing a wide field of view, making it possible to detect objects that are far away from the vehicle.

\subsubsection{Segmentation}

\begin{figure}[h]
\centering
\includegraphics[width=0.99\linewidth]{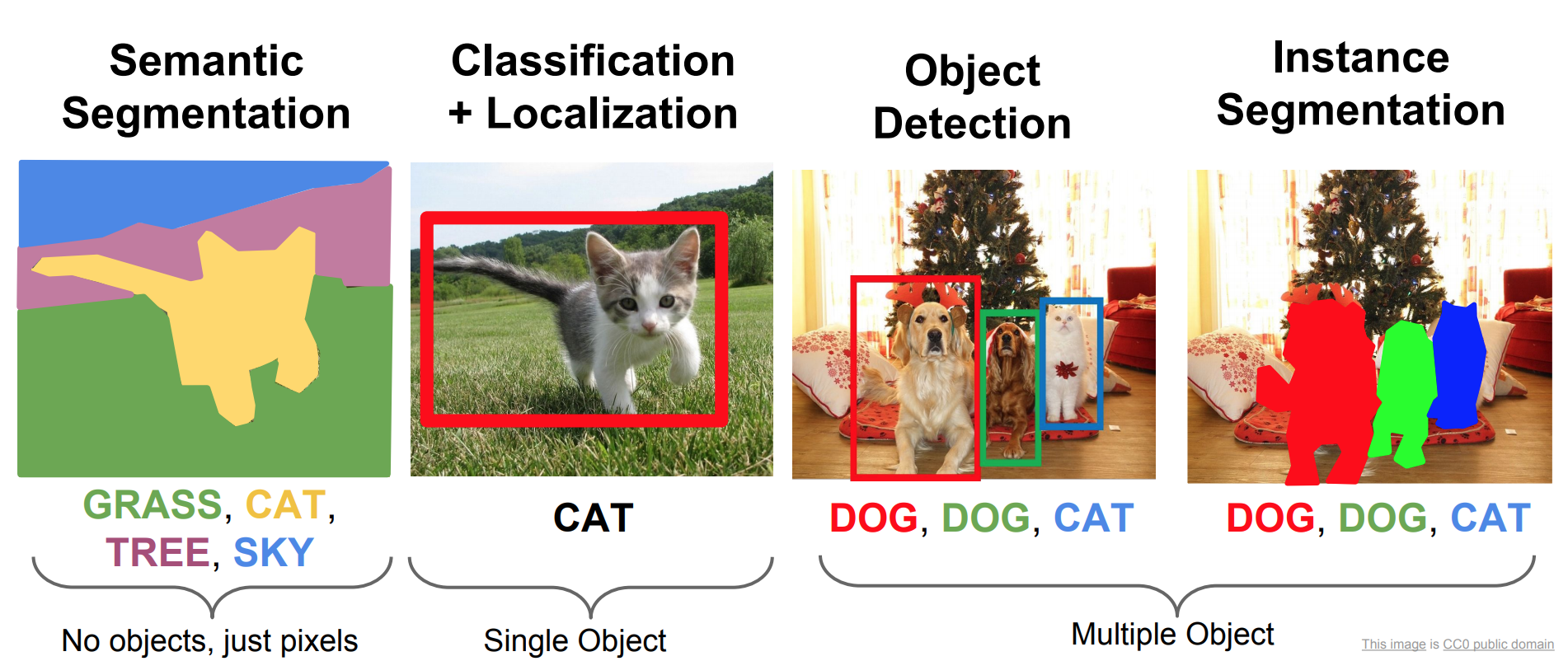}
\caption{Image segmentation, semantic segmentation, and object detection. \cite{li2017lecture}}\label{fig:seg_sem_seg_obj_det}
\end{figure}

Segmentation, involves identifying and labeling each pixel in an image or video as belonging to a particular object or class. This technique is more precise than object detection as it provides a more detailed understanding and partitioning the digital image or video frames. It can be used for tasks such as satellite image analysis, medical image analysis, face recognition and detection, object tracking and scene understanding. Segmentation algorithms use various techniques, such as thresholding, clustering, and deep learning, to segment the objects. On the other hand, semantic segmentation is a type of image segmentation that assigns each pixel in an image a semantic label or category. The goal is to identify and classify the objects and regions of interest within an image, such as cars, buildings, people, or background. The output of semantic segmentation is a pixel-level map that associates each pixel with a specific object or region class \cite{lin2014microsoft}. In summary, while image segmentation aims to group pixels into visually similar parts, semantic segmentation goes further by assigning a semantic meaning to each of these parts. Figure~\ref{fig:seg_sem_seg_obj_det} shows the difference between image segmentation, semantic segmentation, and object detection.

In Table~\ref{image_segmentation}, we summarize the most popular object segmentation algorithms.

\begin{table*}[h]
\caption{Image Segmentation Algorithms}\label{image_segmentation}
\begin{tabular*}{\linewidth}{l p{14.5cm}}
\toprule
\textbf{Algorithm} & \textbf{Description} \\
\midrule
FCN \cite{long2015fully} & Fully Convolutional Network, a CNN architecture that can output pixel-level predictions for semantic segmentation \\
UNet \cite{ronneberger2015u} & A CNN architecture that combines convolutional and deconvolutional layers to produce high-resolution segmentation masks \\
SegNet \cite{badrinarayanan2017segnet} & A CNN architecture that uses an encoder-decoder structure with skip connections to segment images \\
DeepLab \cite{chen2017deeplab} & A CNN architecture that uses atrous convolution and a global pooling layer to capture context for semantic segmentation \\
PSPNet \cite{zhao2017pyramid} & Pyramid Scene Parsing Network, a CNN architecture that uses spatial pyramid pooling and multi-scale feature aggregation for semantic segmentation \\
ICNet \cite{zhao2018icnet} & A CNN architecture that performs real-time semantic segmentation by using a multi-resolution cascade and a shared feature extractor \\
ENet \cite{paszke2016enet} & An efficient CNN architecture for real-time semantic segmentation on mobile devices \\
DeepLabv3+ \cite{chen2017rethinking} & An improved version of DeepLab that uses a feature pyramid and an encoder-decoder structure for semantic segmentation \\
HRNet \cite{wang2020deep} & A CNN architecture that uses a high-resolution representation for semantic segmentation, achieving state-of-the-art performance on various datasets \\
Mask R-CNN \cite{he2017mask} & An extension of Faster R-CNN that adds a segmentation branch to predict the pixel-level mask for each object instance\\
LinkNet \cite{chaurasia2017linknet} & A lightweight encoder-decoder architecture for fast and accurate segmentation, which uses residual connections between the encoder and decoder to improve information flow\\
RefineNet \cite{lin2017refinenet} & A multi-path refinement network that employs residual connections and multi-scale feature fusion to produce accurate and detailed segmentations\\
BiSeNet \cite{yu2018bisenet} & A two-pathway network for real-time semantic segmentation, which combines spatial path (short and fast processing path) and context path (long and slow processing path) for accurate segmentation\\
\bottomrule
\end{tabular*}
\end{table*}

There are several limitations associated with object detection and segmentation techniques in autonomous navigation. One of the major constraints is the need for extensive datasets and computational resources to train the models effectively. Obtaining sufficient training data in real-world environments can be challenging, and the computational resources required to process image or video frames in real-time can be limiting, particularly for embedded systems. Despite these limitations, object detection and segmentation techniques have demonstrated significant potential in enabling autonomous navigation in a variety of applications.

As the technology continues to evolve, it is anticipated that these techniques will become even more accurate and efficient, resulting in safer and more reliable autonomous navigation. Object detection and segmentation enable the system to identify and locate objects, such as obstacles, objects, and pedestrians in real-time. Deep learning algorithms, specifically convolutional neural networks, have proven highly effective in object detection tasks, with one of their main advantages being their ability to learn rich and complex representations of the objects being detected. This allows them to identify objects with more accuracy and robustness, even in challenging scenarios such as adverse weather conditions or low-light environments. 

Furthermore, deep learning algorithms can be trained on large and diverse datasets, enabling them to learn to detect a wide range of objects. These approaches can handle real-time data from various sensors, including cameras, LiDARs, and radars. The information from these sensors can be processed and analyzed by deep learning algorithms in real-time, allowing the autonomous vehicle to respond quickly and accurately to any obstacles that may be present in its path. \par

Object detection techniques, such as Faster R-CNN and YOLO, have been shown to achieve high accuracy and speed in object detection, making them ideal for real-time applications. These techniques use deep learning models to analyze the digital image or video frames and identify the objects of interest. For example, in \cite{mahendrakar2022performance}, Faster R-CNN and YOLOv5 has been used for relative navigation task in On-Orbit Servicing and Active Space Debris Removal Technology. They showed that in a formation flight simulation, while Faster R-CNN is more accurate than YOLOv5 but YOLO is 10 times faster. Visual SLAM has been used for semantic labeling and object detection by combining DeepCNN with Real-Time Graph-Based SLAM for an open-source ground robot \cite{sadeghi2022deep}.

In \cite{shin2022environment} a multi model based object detection with environment-context awareness introduced to solve the environment class imbalance problem for an autonomous mobile robots. They used YOLOv5m \cite{jocher2020ultralytics} with Objects365 \cite{shao2019objects365} dataset for environment-specific object detection , ResNet \cite{he2016deep} trained by Places365 \cite{zhou2017places} for scene classification. Authors in \cite{zhou2022sgm3d} use KITTI \cite{geiger2012we} and Lyft \cite{houston2021one} dataset to train a doubled stereo-guided monocular 3D (SGM3D) object detection framework based on monocular images for  autonomous vehicles navigation.

Furthermore, deep learning has the potential to leverage robots' indoor navigation performance. In a recent study by Afif et al. \cite{afif2022evaluation}, lightweight EfcientDet \cite{tan2020efficientdet} was evaluated as an assistive navigation system for the visually impaired people. The study demonstrated that deep learning-based object detection and classification can effectively aid indoor navigation for individuals with visual impairments. By utilizing the lightweight and efficient object detection algorithm EfficientDet, mobile robots with limited resources can accurately detect obstacles and environmental features in real-time in indoor environments. This technology can greatly enhance the safety and efficiency of indoor navigation, particularly for visually impaired individuals.

In recent years, there have been many advances in the field of deep learning-based obstacle detection. For example, researchers in \cite{talele2019detection} used TensorFlow \cite{abadi2016tensorflow} and OpenCV \cite{bradski2000opencv} for real-time classify and detect obstacles in each image pixel. Also, it provides a high resolution binary obstacle image.

A local and global defogging algorithm based on deep learning models for object detection in harsh weather scenarios proposed by Jiao and Wang \cite{jiao2021road}. The approach fuses the rich representation capability of visual sensors and the weather-resilient performance of inertial sensors to achieve robust positioning in all-weather conditions. Markov random field model is used by fusing gradient, curvature prior, and depth variance potential for obstacle segmentation in \cite{haris2020obstacle}. A CNN model is trained to predict the steering wheel angle and navigate the vehicle safely in a complex environment. LiteSeg \cite{luu2020traditional} introduced by Luu et al. which is a lightweight architecture for adaptive road detection method that combines lane lines and obstacle boundaries. This model simultaneously detect lanes and avoid obstacles. They used NVIDIA Jetson TX2 and Unity software to test their model in a simulated environment. Deep Simple Online and Realtime Tracking (deep SORT) was combined with YOLOv3 to detect and track the dynamic obstacles in \cite{qiu2020vision}. This model is used to detect and track buffaloes and human in paddy field environment.

Authors in \cite{cervera2022u19} have developed a very deep CNN, so called U19-Net which improved encoder-decoder architecture for vehicle and pedestrian detection. The output of this model is a pixel-level mask which detects the presence of vehicles and pedestrians in each pixel. 
Ci et al. presented a new unexpected obstacle detection method for Advanced Driving Assistance Systems (ADAS) based on computer vision \cite{ci2022novel} which consists of two independent detection methods: a semantic segmentation method and an open-set recognition algorithm, that are combined in a Bayesian framework. This model is tested on the Lost and Found dataset \cite{pinggera2016lost} and shows improved detection rate and relatively low false-positive rate, particularly when detecting long-distance and small-size obstacles. The authors acknowledge the limitations of the method, including the false-positive rate of the open-set recognition algorithm, and suggest further work to reduce this rate and to improve the semantic segmentation network. 
Researchers in \cite{wang2022farmland} used Deformable Detection Transformer to detect farmland obstacles in aerial images. In another study \cite{he2022improved}, the ME Mask R-CNN is proposed to conduct fine-grained detection of 11 kinds of obstacles so that the detection accuracy of train obstacle identification can be improved. 
Li et al. \cite{li2022stereovoxelnet} presented a real-time computational efficient object detection framework. They collected a real world stereo dataset which consists of indoor and outdoor scenes.The authors used a stereo camera to generate a 3D point cloud and then used a voxelization method to convert the point cloud into a voxel grid. The voxel grid is then fed into a 3D convolutional neural network to detect objects. The authors compared their method with other 3D object detection methods and showed that their method is more computationally efficient than other methods. 
Further, Cortés et al. \cite{cortes2022dali} presented an unsupervised domain adaptation training strategy for the semantic labeling of 3D point clouds. It bridges the domain gap caused by different LiDAR beams in 3D object detection tasks using LiDAR Distillation. This is done by training a model on data from one domain and then applying it to data from another domain, allowing the model to recognize objects in 3D point clouds even when they are misjudged as background points. 
The improved YOLOv3 algorithm (YOLOv3-4L) is used in \cite{wang2022obstacle} to detect obstacles by simplifying darknet-53 and forming a four-scale detection structure. The output is the type and position of obstacles coincident with dangerous areas. 
In \cite{franke2022towards}, authors presented an on-board and UAV-based vision systems for long-range obstacle detection in railways. 
A CNN-based Obstacle Detection and Collision Prevention developed in \cite{devi2022vision} which has a Raspberry Pi 4 Model B as its control and processing unit. 
A support system for visually impaired person is developed in \cite{hayashida2022obstacle} to detect obstacles and notify the person. This model used YOLOv3-tiny with seven CNNs to extract the features from the image and identify the obstacles. 
Byun et al. \cite{byun2022design} proposed an obstacle detection method for autonomous driving in agricultural environments based on point clouds from pulsed LiDAR technology. Deep learning model is used to detect property information of obstructions. 
YOLO model is used for lane detection and obstacle identification in \cite{huu2022proposing} to Control Self-Driving Cars. They used real-time videos and the TuSimple dataset to test the proposed algorithm. Another study used YOLOv5 for real-time obstacle detection in maritime environments for autonomous berthing in \cite{chen2022real}.
Authors improved YOLOv5s by Convolutional Block Attention Module (CBAM) also, they developed squeeze-and-excitation block called  YOLOv5-SE. 
Additionally in \cite{ma2022fast} YOLO model is used to detect and track obstacles for trains.

In the context of self-driving cars, object detection plays a critical role in detecting and tracking other vehicles, pedestrians, and traffic signs for ensuring safe navigation. One crucial aspect of the object detection system is its ability to accurately determine the presence of objects and classify them into their respective categories. Achieving high accuracy in object detection is crucial for the overall safety and success of autonomous driving systems. To address this problem Gasperini et al. \cite{gasperini2021certainnet} proposed CertainNet to estimate the uncertainty of the outputs (presence of object, and its class, location and size). They used KITTI dataset to train the model.
In another study \cite{hu2022investigating}, the researchers investigated how LiDARs placement affects on the performance of object detection models. They showed that sensor placement is significant in 3D point cloud-based object detection, contributing to 5\% to 10\% performance discrepancy.

Another method for object detection in low visibility conditions is to fuse mmWave radar with visual sensor. The authors \cite{chang2020spatial} proposed a method for object detection in low visibility conditions by using a combination of mmWave radar  and camera. This method used in to develop Spatial Attention Fusion (SAF) for merging the feature maps of radar and camera. Their model has build upon FCOS \cite{tian2019fcos} framework and trained on nuScenes dataset.

Autonomous Underwater Vehicles (AUVs) are one the key component for ocean exploration and surveillance. However, the underwater environment due to the presence of turbid water makes it difficult for AUVs to detect and recognize objects. In \cite{pranav2022deeprecog} the authors proposed DeepRecog framework for improving the performance of AUV vision systems for underwater exploration. The framework employs a three-pronged approach for image deblurring using CNNs, adaptive, and transformative filters, in addition to an ensemble object detection and recognition module that surpasses existing algorithms in terms of precision, such as YOLOv3, FasterRCNN + VGG16, and FasterRCNN. This approach offers real-time detection and recognition. They provide a practical solution to tackle the challenges faced by AUV vision systems in underwater settings. The combination of image deblurring and object detection within a single framework is a noteworthy achievement. The experimental results demonstrate the superior performance of the proposed method and its potential for real-world applications.

To enhance the safety of train operations it is crucial to detect the objects in rail region autonomously. Authors in \cite{zhang2023automatic}, used LiDAR and camera to detect the obstacle within the rail region by utilizing CNNs for semantic segmentation and pixel-wise object detection. An encoder-decoder architecture based on CNNs was used to segment images for the purpose of identifying rail regions. In addition, the residual network structure and depth-wise separable convolution were used for obstacle detection. By using LiDAR, the model was able to cope with unpredictable weather conditions more robustly. The results shows a precision of 99.994\%.

The advancements in object detection and segmentation techniques have shown great potential in the field of autonomous navigation. However, there are several limitations that need to be addressed in order to ensure safe and reliable autonomous navigation. One of the primary challenges is the requirement for large and diverse datasets as well as significant computational resources for training and processing image or video frames in real-time. Obtaining sufficient training data in real-world environments is difficult and computational constraints can be limiting, particularly for embedded systems.

Despite these challenges, object detection and segmentation techniques have proven to be highly effective in enabling autonomous navigation in various applications. As the technology continues to evolve, it is expected that these techniques will become more accurate and efficient, further improving the safety and reliability of autonomous navigation.

Deep learning algorithms, particularly convolutional neural networks, have demonstrated their effectiveness in object detection tasks by learning rich and complex representations of the objects being detected. This allows them to identify objects with greater accuracy and robustness, even in challenging scenarios such as adverse weather conditions or low-light environments. These deep learning algorithms can be trained on large and diverse datasets, enabling them to detect a wide range of objects. Moreover, these approaches can handle real-time data from various sensors, such as cameras, LiDARs, and radars, which can be processed and analyzed by deep learning algorithms in real-time, allowing the autonomous vehicle to respond quickly and accurately to any obstacles that may be present in its path.

Object detection techniques, such as Faster R-CNN and YOLO, have been shown to achieve high accuracy and speed in real-time object detection, making them ideal for various applications. For instance, in \cite{mahendrakar2022performance}, Faster R-CNN and YOLOv5 were utilized for relative navigation tasks in on-orbit servicing and active space debris removal technology. In another study, \cite{sadeghi2022deep} combined Visual SLAM with deep CNN for semantic labeling and object detection in an open-source ground robot.

Several recent studies have demonstrated the potential of deep learning-based object detection and classification for indoor navigation. For example, Afif et al. \cite{afif2022evaluation} evaluated the performance of lightweight EfficientDet as an assistive navigation system for individuals with visual impairments. The study showed that deep learning-based object detection and classification can effectively aid indoor navigation for visually impaired individuals. Furthermore, researchers in \cite{talele2019detection} utilized TensorFlow and OpenCV for real-time obstacle detection in each image pixel, providing a high resolution binary obstacle image.

In summary, deep learning-based object detection and segmentation techniques have shown great potential in enabling safe and reliable autonomous navigation in various applications. Despite the challenges associated with obtaining sufficient training data and computational resources, deep learning algorithms have proven highly effective in identifying and locating objects in real-time. Future advancements in this field are expected to further improve the accuracy and efficiency of object detection and segmentation techniques, enhancing the safety and reliability of autonomous navigation.

\subsection{Localization and Mapping}

Localization and mapping are fundamental technologies in the field of autonomous systems, such as mobile robots and autonomous vehicles, enabling them to operate effectively and safely in dynamic and complex environments. The main objective of these systems is to determine the position and orientation of the vehicle relative to the surrounding environment and to create a map of the environment as a result.

However, a significant challenge in localization and mapping is the presence of uncertainties in sensor measurements and motion models. To address this challenge, various probabilistic methods such as Kalman filters \cite{ullah2019localization}, particle filters \cite{raja2021pfin}, and graph-based SLAM \cite{grisetti2010tutorial} have been developed. These methods allow the system to estimate the robot's position and orientation and update the map of the environment as the robot moves.

In recent years, advancements in sensor technology, algorithms, and machine learning have led to significant progress in localization and mapping. Deep learning-based methods, in particular, have shown promising results in visual localization and mapping tasks by leveraging large-scale datasets and powerful neural network architectures. Moreover, multi-sensor fusion and cooperative localization have contributed to more robust and accurate localization and mapping in complex environments.

Overall, localization and mapping play a vital role in the development of autonomous systems, and ongoing research and development are expected to lead to further improvements in the accuracy, robustness, and efficiency of these systems.

In the following section, we will delve deeper into the application of deep learning for localization and mapping techniques.

\subsubsection{Simultaneous Localization and Mapping (SLAM)}

Simultaneous Localization and Mapping is a method for building a map of an unknown environment while simultaneously determining the position of the system within the environment. This is a challenging problem as the system needs to build a map while simultaneously figuring out where it is within that map. The goal of SLAM is to provide the system with a consistent and accurate map of the environment while allowing it to navigate and make decisions based on its location within the map. SLAM is a fundamental problem in robotics and is considered a crucial component in the development of autonomous systems.

SLAM can be affected by various types of errors that can impact the accuracy and reliability of the system. These errors can be categorized into two main types: measurement errors and modeling errors. Measurement errors occur due to inaccuracies in sensor measurements, such as noise or calibration errors, which can result in inaccurate estimates of the robot's position or the surrounding environment. Modeling errors occur due to the limitations of the mathematical models used to represent the system and the environment. These errors can result in inconsistencies between the estimated and actual measurements, which can further impact the accuracy of the SLAM system. Additionally, SLAM systems can also be affected by environmental changes and dynamic objects, which can cause further errors and require the system to update the map in real-time.

SLAM has been widely studied and has various applications in fields such as autonomous vehicles \cite{qin2020avp}, unmanned aerial vehicles \cite{bavle2020vps, celik2013monocular}, planetary rovers \cite{geromichalos2020slam}, spacecrafts \cite{baldini2018autonomous}, and robotics \cite{tian2022kimera}. It is a fundamental problem in robotics and is considered a crucial component in the development of autonomous systems.

The majority of current SLAM techniques are built around the idea of using visual geometry to model the camera projections, movements, and surroundings. These techniques are characterized by their reliance on explicit modeling and as such, they are often referred to as model-based visual SLAM methods. Model-based SLAM techniques are designed to perform well in environments where the camera projections, motions, and environments can be accurately represented using mathematical models. These techniques are generally robust and reliable, but they can also be sensitive to errors in the models, particularly in complex and changing environments. These techniques can be divided into two main categories: direct methods and feature-based methods. Direct methods directly estimate the camera pose and map from the camera observations. Feature-based methods extract features from the camera observations and use them to estimate the camera pose and map.

In the field of visual SLAM, traditional algorithms have encountered challenges such as scale ambiguity, scale drift, and pure rotation, which have impeded significant progress since 2017 \cite{zhang2022deep}. Deep learning has emerged as a promising approach to tackle these challenges. Nonetheless, the adoption of deep learning techniques also introduces new obstacles, such as the requirement for large datasets and substantial computational resources \cite{beghdadi2022comprehensive}. Despite these limitations, recent advancements in deep learning have given rise to SLAM applications that replicate previously proposed approaches. In recent years, deep learning has emerged as a promising approach to enhance the accuracy and robustness of SLAM algorithms in various applications, including visual SLAM, LiDAR SLAM, and RGB-D SLAM. Deep learning-based visual SLAM methods use deep neural networks to estimate depth information from monocular \cite{amiri2019semi} or stereo camera inputs \cite{li2019depth}. On the other hand, deep learning-based LiDAR SLAM methods use deep neural networks to classify and segment the environment into different objects \cite{xiao2019dynamic, wang2021lidar,yu2020machining,langer2020domain}, making it easier to build a map. Deep learning has also been utilized to address challenges such as handling dynamic objects, improving real-time performance, and dealing with large-scale environments.

One example of deep learning applied to SLAM is the DynaVINS approach presented by Song et al. \cite{song2022dynavins}. This approach enhances the robustness of visual-inertial SLAM in dynamic environments by enabling the system to continue building and updating the map even when the environment undergoes changes. The DynaVINS approach utilizes a deep learning-based approach to predict the motion and location of dynamic objects in the scene, enabling the system to accurately estimate its position and the environment's map.

A method for constructing object-oriented semantic maps developed in \cite{sun2022multi}. It combines the semantic information extracted from instance segmentation with RGB-D version of ORB-SLAM2 \cite{mur2017orb}. The method not only reduces the absolute positional errors (APE) and improves the positioning performance of the system, but also segments the point cloud model of objects in the environment with high accuracy. The proposed method that can help a robot better understand and interact with its environment, with potential applications in the field of human-computer interaction. The paper emphasizes that in order for a robot to have sophisticated interactions with its surroundings, it needs access to 3D semantic information about the environment. The paper also outlines the method of target tracking based on instance segmentation and provides experimental results on a TUM dataset \cite{sturm2012benchmark}.

A visual SLAM-based robotic mapping method, which utilizes a stereo camera system on a rover, has been proposed in \cite{hong2021visual} to address the planetary construction problem. The effectiveness of the proposed method for local 3D terrain mapping has been evaluated with point-clouds from terrestrial LiDAR. However, the paper notes that the camera system on the rover is susceptible to varying illumination conditions, and global localization remains a concern for correcting the rover's trajectory estimate. To create a dense 3D point-cloud map, a self-supervised CNN model is trained using the stereo camera system to estimate a disparity map. The paper acknowledges the method's technical limitations and proposes adaptive robotic mapping as a means to enhance the stereo SLAM's capabilities for the construction of highly detailed and accurate 3D point-clouds.

A robust monocular visual SLAM with direct Truncated Signed Distance Function (TSDF) mapping based on a Sparse Voxelized Recurrent Network, called SVR-Net presented in \cite{lang2023svr}. The proposed end-to-end pipeline utilizes a sparse voxelized structure to reduce the memory occupation of voxel features and gated recurrent units to search for optimal matches on correlation maps, resulting in enhanced robustness. Additionally, Gauss-Newton updates are embedded in iterations to impose geometrical constraints, ensuring accurate pose estimation. The method achieves state-of-the-art localization performance on the TUM-RGBD benchmark and provides direct TSDF mapping, a suitable dense map representation for downstream tasks such as navigation and planning. The approach makes contributions towards the development of robust monocular visual SLAM systems and direct TSDF mapping. \par

\begin{figure*}
	\centering
		\includegraphics[width=\textwidth]{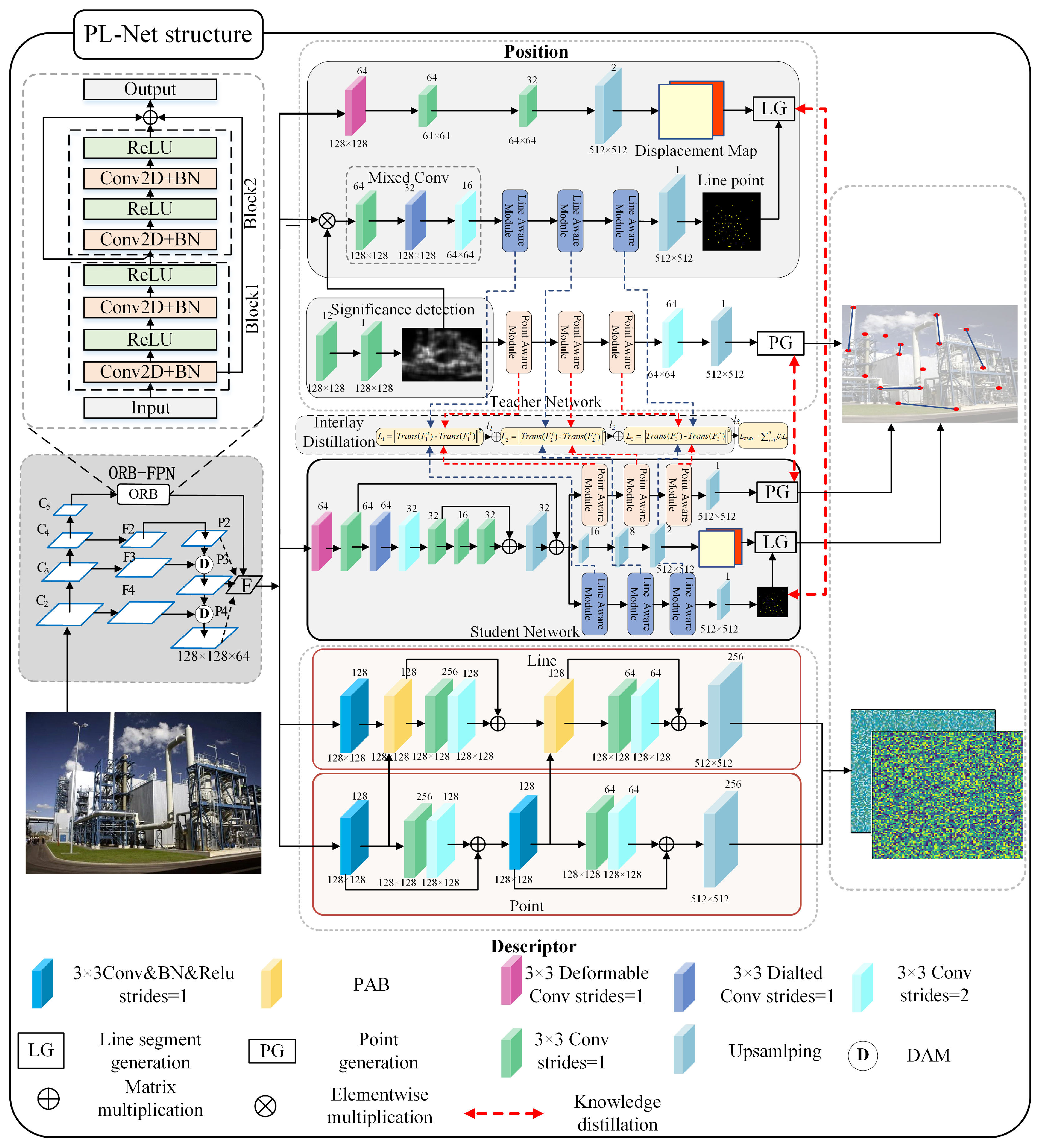}
		\caption{PL-Net, A point-line feature extraction network \cite{lian2023point}}
		\label{fig:plnet}
\end{figure*}

Another visual-SLAM system based on a point-line-aware heterogeneous graph attention network presented in \cite{lian2023point}. The paper proposes a point-line synchronous geometric feature extraction network (PL-Net Figure~\ref{fig:plnet}) to solve the problem of weak point-line extraction ability in complex industrial scenarios. To enhance the accuracy and efficiency of the point-line matching process, the paper presents a heterogeneous attention graph neural network (HAGNN), which uses an edge-aggregated graph attention network (EAGAT) to iterate the vertices of the heterogeneous graph constructed from points and lines. The paper also proposes a greedy inexact proximal point method for optimal transport (GIPOT) to calculate the optimal feature assignment matrix to find the global optimal solution for the point-line matching problem. The experiments on the KITTI dataset and a self-made dataset demonstrate that the proposed method is more effective, accurate, and adaptable than the state-of-the-art methods in visual SLAM. The paper also shows that the multi-feature fusion not only improves the accuracy of the algorithm but also avoids the degradation problem that may occur in the pose solution algorithm when using a single feature.

For fault detection in the scan-matching algorithm of SLAM in dynamic environments a CNN-based approach developed in \cite{jeong2023cnn}. In such environments, the presence of dynamic objects causes changes in the environment detected by the LiDAR sensor, leading to faults in the scan-matching process. The proposed method acquires raw scan data from a 2D LiDAR and performs Iterative Closest Points scan matching. The matched scans are converted into images and fed to a CNN model for training to detect faults in scan matching. The proposed method has been evaluated in various dynamic environments, and the results show high accuracy in detecting faults. The article highlights the significance of solutions for SLAM problems in dynamic environments with only a LiDAR, and deep learning applications for 2D SLAM. The proposed method's main contributions are online processing of scan data, a method to form training images for effective CNN-model training, and high accuracy fault detection in various dynamic real environments. The proposed method's process includes raw scan data acquisition, scan matching, matched scan image generation, and fault detection of scan matching. The CNN model's architecture consists of five 2D convolutional layers with max-pooling layers to extract features and softmax activation to classify scan matching as normal or faulty.

While deep learning-based SLAM methods have shown promising results, they also come with some limitations and challenges. One major limitation is the requirement for a large amount of high-quality labeled training data, which can be time-consuming and expensive to obtain. Additionally, the performance of deep learning-based methods may degrade when operating in highly dynamic and unstructured environments, where the trained models may not generalize well. Another challenge is the potential for overfitting to specific environmental conditions, which can lead to poor generalization when deployed in different environments. This is particularly relevant for visual SLAM methods, where lighting conditions, camera hardware, and environmental texture can have a significant impact on the performance of the deep learning-based models. Also, the computational complexity of deep learning-based SLAM methods could be a barrier, which may require specialized hardware or cloud computing resources to operate in real-time. This can be a significant bottleneck for applications such as autonomous vehicles, where real-time operation is critical. Finally, deep learning-based SLAM methods may also be susceptible to adversarial attacks, where an attacker can manipulate the input data to cause the system to make incorrect predictions. This is a particular concern for safety-critical applications, such as autonomous vehicles.

Overall, while deep learning-based SLAM methods have shown great promise, their limitations and challenges need to be carefully considered in order to ensure their practical applicability in real-world scenarios.

\subsubsection{Attitude Estimation}

Accurate attitude estimation is an essential requirement for effective navigation in any environment. It refers to the process of determining the orientation of a robot relative to a fixed reference frame \cite{asgharpoor2022design}. It plays a critical role in the development of autonomous systems. Different types of sensors can be used to perform attitude estimation, including visual, inertial, and GNSS-based sensors \cite{asgharpoor4387238generalizable}. 

Inertial sensors measure the robot's linear and angular velocity, which are then integrated over time to accurately estimate its position and orientation. However, the integration process can be susceptible to a range of error sources such as sensor noise, drift, and biases. To ensure reliable and precise attitude estimation, sensor fusion techniques can be employed to combine the measurements from various sensors to minimize the impact of individual sensor errors \cite{golroudbari2023end}. There are several techniques for attitude estimation, each with its own strengths and weaknesses.

\textbf{Complementary Filter}: A simple and computationally efficient method that combines the high-frequency response of the gyroscope with the low-frequency response of the accelerometer. This approach works well for estimating the attitude of a robot that is subject to low-frequency disturbances, but it may suffer from drift over time due to bias errors in the gyroscope and accelerometer.

\textbf{Extended Kalman Filter (EKF)}: An advanced version of the Kalman filter that uses a probabilistic model to estimate the true state of the system from noisy sensor measurements. It is capable of handling nonlinear systems, making it a popular choice for attitude estimation. The EKF can provide accurate estimates of attitude even in the presence of sensor noise and measurement errors. However, the EKF is computationally expensive and requires a lot of memory to store the covariance matrix.

\textbf{Unscented Kalman Filter (UKF)}: A more advanced version of the EKF that uses a nonlinear transformation of the state variables to generate a set of sigma points. These sigma points are then propagated through the nonlinear system model to estimate the state of the system. The UKF is more accurate than the EKF and requires less computation than other advanced filtering methods such as the particle filter.

\textbf{Particle Filter}: A non-parametric filtering method that uses a set of weighted particles to represent the probability distribution of the robot's state. It is capable of handling nonlinear systems and can provide accurate estimates of attitude even in the presence of nonlinearities and sensor noise. However, the particle filter is computationally expensive and requires a large number of particles to represent the state of the system accurately.

Reliable and accurate attitude estimation allows a robot to determine its position and orientation relative to a fixed reference frame, which is necessary for navigation and obstacle avoidance tasks. The development of advanced attitude estimation techniques has made it possible to build autonomous systems capable of operating in complex and dynamic environments, making attitude estimation a vital component of the robotics and autonomous systems field.

In recent decades, numerous MSDF techniques and Deep Learning models have been developed to enhance the accuracy and reliability of attitude estimation techniques. Attitude can be estimated using a minimum of a 6-Degree-of-Freedom (6DoF) Sensor Fusion Algorithm (SFA) in MSDF methods. In 6DoF SFAs, a three-axis accelerometer is fused with a three-axis gyroscope to estimate the attitude. However, 6DoF SFAs are not suitable for attitude and heading estimation/determination as the accelerometer cannot measure the yaw (heading) angle, and the gyroscope can only measure the yaw angle's rate \cite{asgharpoor4387238generalizable}. An alternative method is to fuse magnetometer readings with a 6DoF SFA to estimate the full orientation (attitude and heading). However, a magnetometer's primary disadvantage is the magnetic disturbances, which adversely affect its performance, mainly when used for indoor navigation. Several techniques have been developed to reduce the effect of magnetic disturbances on the filter performance, such as the Factorized Quaternion Algorithm (FQA) \cite{lee2012factorized}.

Most SFAs are developed and parameterized based on the system's dynamic model, which requires a precise choice of model parameters \cite{fauske2007estimation}. However, no algorithm can handle all types of motions, and it is challenging to design a generalizable algorithm for all possible scenarios. The traditional approach is to develop a model for a specific scenario and then adapt it to different situations. However, this approach is time-consuming and requires extensive domain knowledge. In recent years, deep learning models have shown great potential in solving sequential data problems, including attitude estimation \cite{hoang2022yaw,asgharpoor2022design}, error modeling \cite{zhao2022attitude}. These models can learn the hidden patterns and relationships within the data and handle complex and nonlinear relationships between sensor measurements and attitude. Several studies have shown the potential of deep learning models for attitude estimation, including recurrent neural networks (RNNs) and convolutional neural networks \cite{wang2020recent,xiao2018opportunities,zulqarnain2020comparative,nevavuori2020crop,bouktif2019single}.

The field of robotic perception has seen significant research attention in recent years, with many studies focusing on the development of odometry or the fusion of heterogeneous sensors for attitude estimation, including inertial-GPS or inertial-visual fusion, and learning-based frameworks coupled with conventional filtering methods. However, there remains a notable gap in research efforts devoted to end-to-end learning-based inertial attitude estimation. Although relying solely on inertial data can result in significant drift or biases, the benefits of developing such models in scenarios where visual data is not available or GPS is denied cannot be understated. Therefore, there is a pressing need to design a learning-based model that can achieve end-to-end inertial attitude estimation, thereby providing highly accurate and reliable estimates of a system's orientation based solely on inertial sensor data. Such a model holds great potential for a range of applications, including robotics, navigation, and aerospace engineering.

There are currently only four models that utilize end-to-end learning-based methods for inertial attitude estimation. RIANN \cite{weber2021riann} was the first to introduce a GRU-based approach, but their article lacked sufficient quantitative information. Another study \cite{narkhede2021incremental} presented an LSTM-based model but was limited to only one sampling rate, not mentioned in their publication and was not tested on publicly available inertial datasets. The third study \cite{brotchie2022leveraging} employed a self-attention mechanism, but their model was only trained and tested on the OxIOD dataset \cite{chen2018oxiod}, which limited it to one type of motion and a specific sampling rate. In the article by Asgharpoor et al. \cite{asgharpoor4387238generalizable}, three end-to-end learning frameworks were presented for the estimation of orientation. These frameworks were designed to generalize across different environments and sensor sampling rates. However, it should be noted that the use of windows of IMU data in these frameworks can result in a delay.

\subsubsection{Odometry}

Odometry is the process to determine the relative change in the position and orientation between tow or more coordinate frames \cite{aqel2016review}. There are various odometry techniques, including visual \cite{nister2004visual}, inertial \cite{solin2018inertial}, encoder \cite{brossard2019learning}, optical flow \cite{muller2017flowdometry}, GPS/GNSS \cite{ohno2004differential}, and lots of others.

The most common odometry techniques are as follows \cite{chen2020deep}:
\begin{itemize}
	\item \textbf{Inertial Odometry}: 
	\item \textbf{Visual Odometry}: Utilizing computer vision to estimate the position and orientation of a camera relative to a fixed reference frame. Deep learning can be employed to extract high-level features from visual data which could be used as an alternative solution to the Odometry problem. 
	\item \textbf{Visual-Inertial Odometry}: Fusing visual and inertial data to estimate pose and trajectory. It involves the integration of measurements from the IMU with visual data from a camera. This enables the reconstruction of the 3D scene structure and the camera's motion through it. The visual data collected from the camera is egocentric, meaning it is from the camera's point of view, while the IMU data is in the global reference frame.
\end{itemize}

Further, visual odometry techniques can be broadly divided into two distinct categories:
\begin{itemize}
	\item \textbf{Supervised}: The most popular VO techniques and used when the ground truth data is available which maps the extracted features from image to motion transformation \cite{han2019deepvio}.
	\item \textbf{Unsupervised}: A learning scheme where the model is trained on unlabeled data, without the need for explicit ground truth information. In the context of visual data, unsupervised learning can be used to extract meaningful features or representations from an unlabeled dataset and when ground truth labels are not available or too expensive to obtain.
\end{itemize}

\begin{figure*}
\centering
	\includegraphics[width=\textwidth]{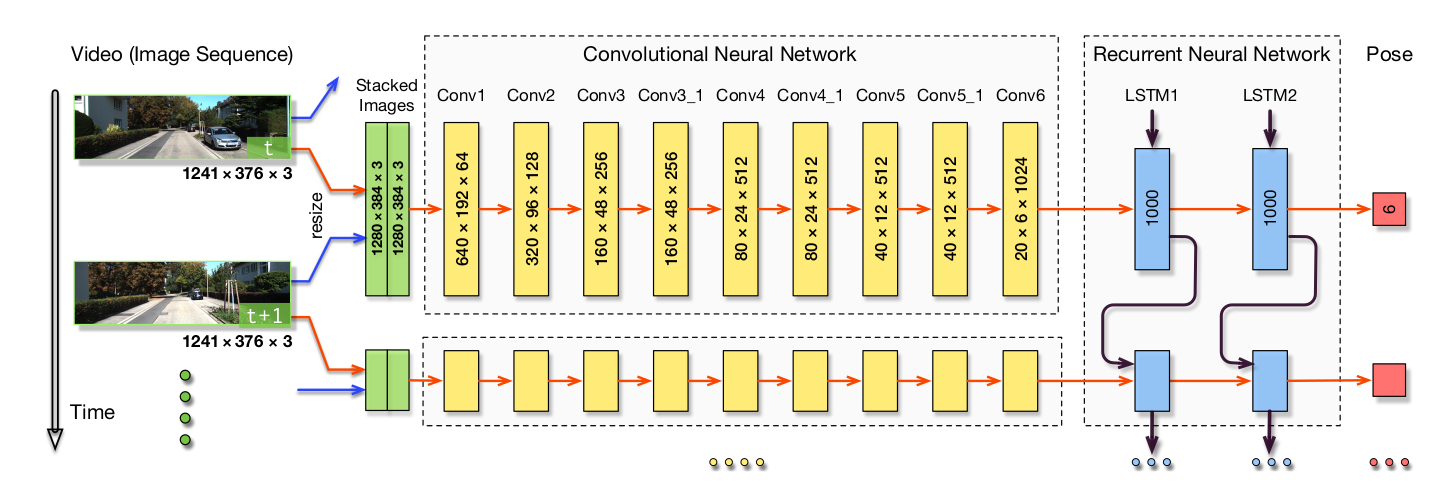}
	\caption{DeepVO, A typical structure of an End-to-End Visual Odometry \cite{wang2017deepvo}}
	\label{fig:odometry}
\end{figure*}

One of the earliest works on the supervised VO, conducted by Konda et al. \cite{konda2015learning}. They utilized CNN to predict velocity and changes in direction. DeepVO \cite{wang2017deepvo} combined CNN and RNN to develop an End-to-End deep learning framework for pose estimation. Authors used deep neural network to extract the features from images via CNN and learn the temporal dependencies through RNNs. They used KITTI dataset \cite{geiger2013vision} to train and test the model. To calculate the error between the prediction and true position and orientation they used the Mean Square Error (MSE) of all positions $\mathbf{p}$ and orientations $\mathbf{\phi}$ in the sequence. The MSE of the position is defined as:

\begin{equation}
	\theta^{\star} = \frac{1}{N}\sum_{i=1}^{N} \sum_{t=1}^{T} \left\| \mathbf{p}^{\star}_t - \mathbf{p}_t \right\|_2^2 + \left\| \mathbf{\phi}^{\star}_t - \mathbf{\phi}_t \right\|_2^2
\end{equation}
where $\mathbf{p}^{\star}_t$ and $\mathbf{\phi}^{\star}_t$ are the predicted position and orientation at time $t$, $\mathbf{p}_t$ and $\mathbf{\phi}_t$ are the true position and orientation at timestep $t$, and $N$ is the number of samples. Based on this architecture, various models have been developed, since. Similarly, MagicVO \cite{jiao2019magicvo} combined CNN and Bi-LSTM to 6 DoF pose estimation. They replaced LSTM layers with Bi-LSTM layers followed by a 256 unit dense layer. The results shows an improvement in terms of translational and rotational error. However their comparison results consist of only five sequences.

In contrast with DeepVO, UnDeepVO \cite{li2018undeepvo}, employed unsupervised techniques to develop an end-to-end VO framework . They used VGG-based CNN \cite{simonyan2014very} as the pose estimator and it trained using stereo images instead of consecutive monocular images

\begin{figure}
	\centering
		\includegraphics[width=0.5\textwidth]{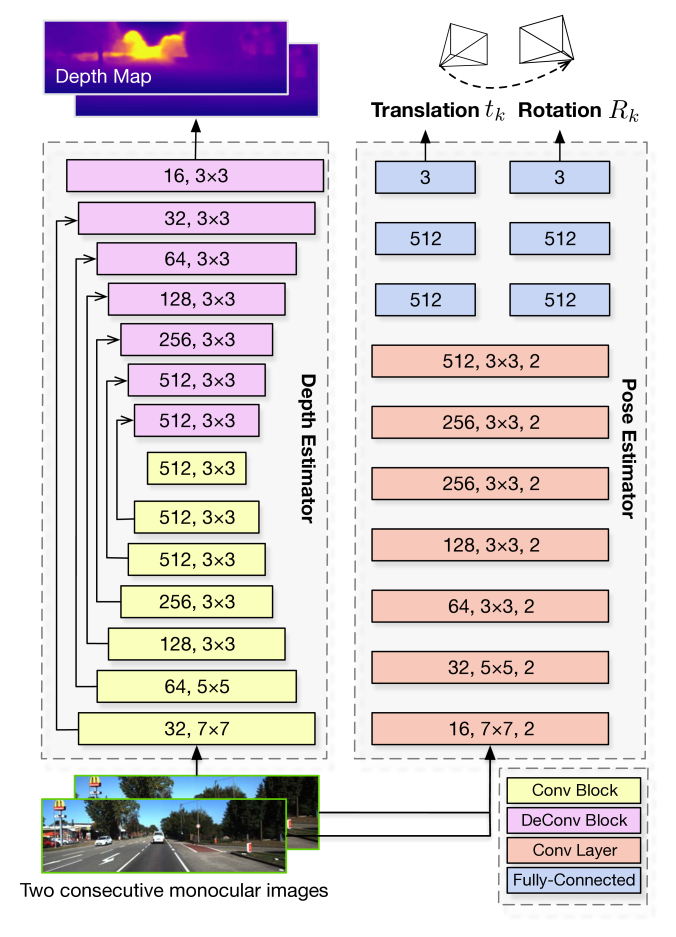}
		\caption{UnDeepVO Architecture for Unsupervised End-to-End Visual Odometry \cite{li2018undeepvo}}
		\label{fig:undeepvo}
\end{figure}

The monocular visual odometry algorithm UnDeepVO, proposed by Li et al. \cite{li2018undeepvo}, offers an alternative approach to DeepVO. Unlike DeepVO, UnDeepVO is an end-to-end framework for visual odometry that employs unsupervised learning to estimate the camera's motion. It uses a VGG-based CNN \cite{simonyan2014very} as the pose estimator, which is trained on stereo images, rather than consecutive monocular images (Figure~\ref{fig:undeepvo}). The authors use unsupervised learning to train the CNN, incorporating both spatial and temporal dense information in the loss function to improve accuracy and robustness.

By incorporating unsupervised learning and stereo images, UnDeepVO can estimate both pose and depth using a deep neural network. Compared to other monocular methods such as ORB-SLAM without loop closure, this model has been shown to be more accurate and robust. Overall, UnDeepVO provides a promising approach to visual odometry that does not rely on labeled data and can achieve comparable or superior performance to other monocular methods. Its use of unsupervised learning and stereo images provides a foundation for future research in this area.

GANVO \cite{almalioglu2019ganvo} used Generative Adversarial Network for unsupervised learning to estimate pose and generate depth map for a
monocular video sequences. Ther result shows that the model outperform conventional and other unsupervised methods. An end-to-rnd, Sequence-to-sequence Probabilistic Visual Odometry (ESP-VO) deep learning framework is presented by Wang et al. \cite{wang2018end} which used monocular camera and Recurrent Convolutional Neural Network (RCNN) for pose estimation directly from raw images.


VINet \cite{clark2017vinet} use monocular camera and IMU for pose estimation. It was one of the first End-to-End Visual-Inertial Odometry (VIO) deep learning models which formulates the odometry a sequential learning problem. VINet contains two main parts, (1) CNN and (2) LSTM. CNN takes two sequential images to produce the feature vector describing the motion. In contrast, LSTM network process the IMU measurements to extract the temporal motion characteristics hidden in sensor raw data. Furthermore, a core LSTM networks combined and fused the features obtained from CNN and LSTM to predict the pose information. Employing both visual and inertial data increased its robustness. VINet trained and tested on KITTI dataset.
Similarly, HVIOnet proposed for Unmanned Aerial System \cite{aslan2022hvionet}, which is hybrid CNN-BiLSTM network and trained on EuRoC dataset. Compare to VINet, it has multiple fully connected layer after CNN and LSTM networks. However they only tested their network on EuRoC dataset and ROS environment. Yang et al. \cite{yang2022efficient} presented an adaptive visual inertial odometry system that reduces computational redundancy by selectively deactivating the visual modality. The proposed method employs a policy network that learns to deactivate the visual feature extractor based on the current motion state and IMU readings, achieving up to 78.8\% computational complexity reduction for KITTI dataset evaluation. The authors argue that visual data processing is considerably more resource-intensive than that of the inertial measurement unit, and may not always enhance pose estimation accuracy. The findings suggest that the proposed approach outperforms simple sub-optimal strategies and offers an interpretable, scenario-dependent adaptive behavior. Notably, the authors note that the learned strategy is model-agnostic and can be easily implemented in other deep VIO systems.

Another study \cite{george2023visual} was conducted to assess the feasibility of using VIO as a redundant positioning system for high-altitude drone operation and the collected data through this research has been published in \cite{fgi}. The study shows that the stereo-VIO provided the results, with a baseline of 30 cm providing better results than the mono-VIO for all flying altitudes of 40-100 m. Also they found that external conditions affected the estimation accuracy, and fusing IMU data improved positioning performance and the robustness of the system to external conditions. The authors suggest future research could focus on developing well-engineered systems capable of real-time estimation, integrating additional sensors to reduce drifts, evaluating whether robust positioning with stereo-VIO outperforms commonly used mono-VIO algorithms, and studying the effects of system calibration models.

Tu et al. have presented a novel EMA-VIO framework \cite{tu2022ema} that utilizes external memory and attention-based fusion mechanism to estimate the position and translation of mobile agents using visual-inertial data. The proposed framework has shown superior performance in different scenarios, including challenging ones such as overcast days and water-filled ground, and has outperformed traditional and learning-based VIO baselines. The paper introduces a multi-state geometric constraint in the pose loss function that accelerates training convergence and provides more accurate pose results. The experimental results on both the KITTI dataset and the authors' own robot dataset validate the effectiveness of the proposed framework, especially in challenging environments such as overcast days and wet ground that are difficult for traditional VIOs to extract and match visual features. The findings demonstrate that EMA-VIO surpasses both traditional VIO models and other learning-based VIO models.

SelfVIO \cite{almalioglu2022selfvio} is a deep learning-based monocular Visual-Inertial Odometry (VIO) system that utilizes self-supervised learning to estimate both the camera pose (position and orientation) and depth of objects in the environment. Monocular VIO systems only use a single camera, as opposed to stereo or multi-camera systems, and use both visual information and inertial data from an IMU to estimate the camera's pose and depth. The system is trained on unlabeled data and learns to predict the motion of the camera and depth of objects based on the observations themselves.

Authors in \cite{cortes2018deep} used CNN-FFNN to estimate the speed by using IMU raw measurements over a window of 2 seconds.
IONet \cite{chen2018ionet}, one of the first Inertial End-to-End deep learning network for odometry. It consist of two stacked Bi-LSTM layers, each layer has 96 units. The first layer takes the accelerometer and gyroscope readings in a windows of 200 frames and stride of 10 as input, and the second layer takes the output of the first layer as input. The output is the polar vector as the relative position and orientation of the sensor.

\begin{equation}
	(\mathbf{a}, \mathbf{\omega})_{200,6} \rightarrow (\mathbf{\delta p}, \mathbf{\delta \phi})_1
\end{equation}

Authors in \cite{chen2019deep} used deep learning to estimate displacement in polar coordinates and the corresponding uncertainty over a window of 200 frames by using IMU raw data. The model input shape is $(200, 6)$, where 6 is the number of features (accelerometer and gyroscope readings) and 200 is the window size. The input fed a fully connected layer with 128 units, followed by a 128 and 256 units Bi-LSTM layers. The output of the last Bi-LSTM fed into a 128 units fully connected layer, which is followed by double 2 units fully connected layer to predict the displacement in polar coordinates amd the uncertainty.

AbolDeepIO \cite{esfahani2019aboldeepio} utilized a novel triple channel LSTM-based network for autonomous vehicle inertial odometry. The proposed model extract the relationship between accelerometer and gyroscope readings followed by their sampling rate to estimate any changes in position and orientation of the sensor.

Lima et al. \cite{silva2019end} proposed a deep inertial odometry method that utilizes a combination of convolutional neural networks (CNN) and bidirectional long short-term memory (BiLSTM) layers. The model takes two distinct inputs of the shape $(200, 3)$, which represents the window size of 200 and the number of axis. The CNN extracts features from the inertial measurement unit (IMU) sensor readings, which are then passed to two BiLSTM layers. The output of the final BiLSTM layer is fed into a fully connected layer with seven units, representing the orientation in quaternion form and position in metric form. Additionally, a multi-task learning framework is integrated to balance the pose distance metrics. The equation multi-task learning is as follows:

\begin{equation}
\mathbf{L}_{MTL} = \Sigma_{i=1}^{N} \text{exp} (-\log \sigma^2_i)L_i + \log \sigma^2_i
\label{eq:multi-task}
\end{equation}

where $\sigma_i$ and $L_i$ represent the variance and loss function of task $i$, respectively. The multi-task learning framework is implemented by training the model to predict both the device's translation and rotation. The loss function is a weighted sum of the translation and rotation error, and the weights of the two tasks are dynamically adjusted during training to balance the contributions of the translation and rotation errors. The additional output layer predicts the rotation quaternion between two sequential moments. The loss function for the multi-task learning is a combination of the mean absolute error of the predicted translation and rotation, weighted by two coefficients that control the relative importance of the two tasks. The coefficients are updated during training using a technique called dynamic weighting, which balances the contributions of the two tasks based on their performance.

The multi-task learning approach allows the model to learn to balance the importance of rotation and translation, leading to improved accuracy in estimating the device's position and orientation. The proposed method outperforms recent inertial odometry methods in both qualitative and quantitative evaluations.

One of the primary constraints of Lima et al.'s method is its generalizability. This model demonstrates optimal performance solely on the dataset on which it was trained. Furthermore, due to the model's inability to handle variations in gyroscope sampling rates, it is incapable of accurately estimating the position and orientation of the sensor when exposed to diverse sampling rates \cite{asgharpoor2022design}.

Anothe deep inertial odonetry nodel, called Tightly-coupled Extended Kalman Filter framework (TILO) developed by Liu et al. \cite{liu2020tlio}. This model used IMU measurements to estimate relative position and orientation by using a deep neural networked coupled by and EKF. Two Bidirectional-LSTM is used to estimate the foot trajectory in gait analysis by using the accelerometer and gyroscope readings with window size of 256 \cite{guimaraes2021deep}. In \cite{soyer2021efficient}, authors presented a real-time pedestrian position estimation based on smartphone internal IMU measurements. They used window size of 150, the first 100 frames are selected from the past measurements and the rest of 50 frames are selected from the future measurements. This method help them to reduce the latency of the system in comparison to the other methods such as IONet. The model is similar to Lima et al. network, consist of two distinct input (accelerometer and gyroscope), each input followed by two CNN layer and a max pooling layer with concatenated and fed into two-Bidirectional-LSTM layer.

Uno et al. \cite{uno2022deep} introduced Deep Inertial Underwater Odometry (DIUO) system for underwater 6-DOF odometry based on TLIO network. Convolutional block attention modules is combined with Res2Net in \cite{chen2022deep}, to estimate the velocity in $x$ and $y$ axis. They test their model on the RONIN and OxIOD dataset and the results show that the proposed model outperforms the other models in most of scenarios. A LSTM-based model for visual-inertial odometry has been proposed for UAVs in GNSS denied environments by Deraz et al. \cite{deraz2022deep}. Wang et al. introduced A2DIO model which used a hybrid CNN-LSTM architecture with a temporal attention block for pedestrian odometry based on a window of 2-second IMU measurements \cite{wang2022a2dio}.

One of the major drawbacks of using GNSS and GPS for navigation is their weakness against signal obstruction, such as tall buildings or natural obstacles like mountains. The address the lack of reliable indoor positioning alternatives to GNSS/INS, authors presented L5IN+ \cite{shoushtari2022l5in+}. The proposed methods consist of two approaches, namely deep inertial odometry and Kalman Filtering, to predict velocity vector elements and relative positions with noisy labeled data from 5G networks. The authors also develop an analytical platform for data collection and a simulation website for researchers to generate ground truth trajectories and simulate cellular measurements with assigned quality and exact error values. The major contribution of this paper is the development of a novel combination of DNN and KF for the relative and absolute positioning. The DNN model is designed to be robust to noisy labelled data, for example, data from a 5G network. The authors also introduce a semantic error generation approach to their simulation model.

The errors and time-varying biases present in the measurements of MEMS-based IMUs cause large drift in pose estimates, and only relying on the integration of the inertial measurements for state estimation is infeasible. In \cite{cioffi2022learned}, authors proposed a learning-based odometry algorithm for autonomous drone racing tasks that uses an IMU as the only sensor modality.  The proposed algorithm aims to overcome this problem by combining an EKF, which is driven by the IMU measurements, with a learning-based module, which has access to the control commands, in the form of commanded collective thrust. The learning-based module is a TCN that takes as input a buffer of commanded collective thrust and gyroscope measurements and outputs an estimate of the distance traveled by the quadrotor, which is used to update the filter. The proposed algorithm is compared to state-of-the-art filter-based and optimization-based visual-inertial odometry algorithms, as well as the state-of-the-art learned-inertial odometry algorithm TLIO, and achieves superior performance. The main contribution of this work is the development of an inertial odometry algorithm that combines a model-based filter with a learning-based module to improve the accuracy of state estimation for agile quadrotor flight. Another study \cite{zhang2022dido}
conducted on improving quadrotor state estimation. The proposed framework, Dido, employs GRU-based networks to learn both IMU and dynamic properties for better performance in quadrotor state estimation. The paper demonstrates that the system provides precise observations with only raw IMU and tachometer data, and combines quadrotor dynamic constraints and network observations within a two-stage EKF to jointly estimate kinematic, dynamic states and extrinsic parameters. 

Legged robots offer advantages over wheeled robots in traversing uneven terrain, but controlling and navigating them is challenging due to the complex dynamics of legged locomotion. Navigation and odometry are crucial, involving determining the robot's position and movement through sensor data analysis. Legged robots commonly use multiple sensors such as cameras, IMUs, and range finders to improve accuracy and reliability. Ongoing research focuses on the development of advanced algorithms and sensors to enhance legged robot navigation and odometry. Legged robots have the potential to play a vital role in exploring remote and hazardous environments, disaster response, and search and rescue operations.

To address the challenges in legged robots navigation, Buchanan et al. \cite{buchanan2022learning} proposed a new method to improve proprioceptive state estimation. The approach is based on using a learned displacement measurement from IMU readings, which reduces the drift in state estimation caused by unreliable leg odometry on challenging terrains such as slipping and compressible surfaces. The displacement measurement from IMU readings is combined with traditional leg odometry, and results from real robot experiments show a reduction in relative pose error by 37\% in challenging scenarios and a 22\% reduction in error when used with vision systems in visually degraded environments. The article's contributions include a kinematics-inertial estimator based on learned IMU displacement measurements, integration with both a filter-based estimator and a factor graph, and extensive testing on a quadruped robot, including field experiments navigating a mine. The study concludes that the addition of the learned measurement significantly improved state estimation compared to a traditional kinematic-inertial estimator without the learned measurement.

The heading/yaw angle could calculated by magnetometer however, this method suffers from the magnetic interference, drift, bias. The authors of \cite{wang2022magnetic} proposed a system to estimate the magnetometer bias online to provide a global consistent heading estimation in magnetic disturbances. They used a graph-optimization-based system which fused learning-based inertial odometry and magnetometer readings. The proposed system showed a 50\% improvement in positioning accuracy in an indoor scenario. They simplified the model to perform a real-time estimation and compensation procedure and achieved superior positioning performance when compared to calibrated magnetometer data and raw magnetometer data. The proposed method is suitable for consumer-grade devices as it does not require calibration before use. The study combined network and traditional empirical models to form a robust pedestrian positioning solution. The testing stage fuses the information from IMU, the pre-trained network, and magnetometer to provide a trajectory without heading drift and is not affected by local magnetic field disturbance. The system uses the long-term mean value of the geomagnetic field to obtain a consistent global heading direction and stores magnetic field observations for a sufficiently long time or keyframes to estimate magnetometer biases.

The research conducted by Saha and colleagues \cite{saha2022tinyodom} proposes a new framework for neural inertial navigation that addresses the challenge of deploying machine learning models on ultra-resource-constrained (URC) devices in GPS-denied environments. The framework, named TinyOdom, utilizes a hardware-in-the-loop (HIL) automated machine learning (AutoML) framework based on Bayesian Optimization (BO) to create lightweight inertial odometry models that are suitable for deployment on URC devices. The framework incorporates hardware and quantization-aware Bayesian neural architecture search (NAS) and a TCN backbone to develop lightweight models. Moreover, the proposed framework incorporates a magnetometer, physics, and velocity-centric sequence learning formulation to enhance the localization performance of the lightweight models, along with a barometric g-h filter that tracks altitude with high accuracy and is resistant to disturbances and dynamics. The authors evaluated their framework on four different URC hardware platforms for four different applications, and the results show that it is an effective and efficient solution for neural inertial navigation on URC devices. They also suggest that their work has the potential to enhance the cost and energy footprint of embedded odometry while being adaptable to any reduced footprint hardware.

\subsection{Planning and Decision Making}
\subsubsection{Path Planning}

\subsubsection{Obstacle Avoidance}
This sub-section could describe the use of deep learning in the development of algorithms for autonomous navigation systems to plan safe and efficient paths while avoiding obstacles in the environment. It could discuss the use of deep reinforcement learning, deep neural networks, and other deep learning algorithms to model the relationships between the vehicle's actions and the environment.

Authurs in \cite{visca2022meta} developed an adaptive deep meta-learning energy-aware path planner that uses a 1D convolutional neural network to analyze terrains sequentially, as experienced by an autonomous mobile robot (AMR), to provide energy estimates in real-time. The paper provides evidence of the method's improved robustness to provide more informed energy estimations and energy-efficient paths when navigating challenging uneven terrains. The method integrates the adaptive energy estimator into a state-lattice A* path planner to consider actual robot mobility constraints. The experimental results show the model's performance in simulation over several typologies of natural terrains and unstructured geometries, and it is compared with alternative state-of-the-art deep learning solutions. The method's main novelty is the development of a deep meta-learning architecture that efficiently adapts its energy estimates to new terrains based on a small number of local measurements. The paper is well-structured, and the proposed method is adequately explained with diagrams and figures. The results are promising and show the method's potential in improving energy efficiency in autonomous mobile robots.


\section{Conclusion \label{sec:conclusion}}

However one of the main challenging part of utilizing deep learning for autonomous navigation is training, not only the computational cost, but also the availbality of datasets. While for various applications you should sentesize their own dataset because of unavilibity of dataset, for other applications you must spend much time for data preprocessing due to missing data, false lables, or incomplete annotations. Despite these challenges, recent advancements in deep learning applications and methods for autonomous navigation have shown promising results and have the potential to revolutionize various industries.

One of the key areas where deep learning has made significant strides is in object detection and recognition. Convolutional Neural Networks (CNNs) have been employed to effectively detect and classify objects in real-time. The introduction of architectures such as Faster R-CNN, SSD, and YOLO has significantly improved the accuracy and efficiency of object detection algorithms. These advancements have paved the way for autonomous vehicles to better perceive their surroundings and make informed decisions based on the detected objects.

Furthermore, deep learning techniques have been successfully applied to semantic segmentation tasks in autonomous navigation. Semantic segmentation aims to assign specific labels to each pixel in an image, enabling more precise understanding of the scene. Convolutional Encoder-Decoder networks, such as U-Net and DeepLab, have shown remarkable performance in segmenting different objects and regions in complex environments. This level of detailed scene understanding is crucial for autonomous systems to navigate safely and efficiently.

Another area of advancement in deep learning for autonomous navigation is Simultaneous Localization and Mapping (SLAM). SLAM algorithms combine sensor data with deep learning methods to simultaneously build a map of the environment and estimate the vehicle's position within that map. Traditional SLAM methods relied on handcrafted feature extraction and matching techniques, but deep learning approaches, such as Deep SLAM and ORB-SLAM, have demonstrated superior performance by directly learning the features and associations from raw sensor data. These techniques have proven to be robust in challenging scenarios with dynamic environments, making them suitable for real-world autonomous navigation applications.

Moreover, reinforcement learning has emerged as a powerful paradigm for training autonomous agents. Deep Reinforcement Learning (DRL) algorithms, such as Deep Q-Networks (DQN) and Proximal Policy Optimization (PPO), have been successfully applied to autonomous navigation tasks. By learning through trial and error, these algorithms can navigate complex environments and learn optimal policies for decision-making. DRL has shown great potential for training autonomous vehicles to handle diverse situations and adapt to changing environments.

Despite the remarkable advancements in deep learning for autonomous navigation, there are still several challenges that need to be addressed. One such challenge is the robustness of deep learning models to adverse conditions, such as poor lighting, occlusions, or sensor noise. Ensuring the reliability and safety of autonomous systems in real-world scenarios is of utmost importance, and further research and development are needed to improve the resilience of deep learning models.

Additionally, the interpretability and explainability of deep learning models remain areas of concern. Autonomous systems should be able to provide clear and understandable justifications for their decisions, especially in critical situations. Techniques such as attention mechanisms and model visualization can provide insights into the internal workings of deep learning models, enabling humans to trust and understand their behavior.

In conclusion, recent advancements in deep learning applications and methods have shown tremendous potential for autonomous navigation. Object detection, semantic segmentation, SLAM, and reinforcement learning techniques have significantly improved the perception, mapping, decision-making, and control capabilities of autonomous systems. However, further research and development are necessary to overcome challenges related to training, robustness, interpretability, and safety. With continued progress in deep learning, we can expect autonomous navigation systems to become more reliable, efficient, and capable, leading to widespread adoption in industries such as transportation, robotics, and surveillance.

\section*{Acknowledgements}
The study presented in this paper is based on A. Asgharpoor Golroudbari's M.Sc. Thesis ("Design and Simulation of Attitude and Heading Estimation Algorithm", Department of Aerospace, Faculty of New Sciences \& Technologies, University of Tehran).
We would like to express our sincere gratitude to Prof. Parvin Pasalar, Dr. Farsad Nourizade, and Dr. Maryam Karbasi Motlagh at the Students' Scientific Research Center at Tehran University of Medical Sciences. Their invaluable scientific advice and support in the form of access to computational resources were instrumental in the success of our research. We are deeply appreciative of their contributions and the time they dedicated to helping us.
\section*{Conflict of Interest}

The authors declare no competing interests.
\bibliographystyle{unsrt}  
\bibliography{references}

\begin{thebibliography}{100}

\bibitem{britting2010inertial}
Kenneth~R Britting.
\newblock {\em Inertial navigation systems analysis}.
\newblock Artech House Print on Demand, 2010.

\bibitem{ni2022improved}
Jianjun Ni, Kang Shen, Yinan Chen, Weidong Cao, and Simon~X Yang.
\newblock An improved deep network-based scene classification method for
  self-driving cars.
\newblock {\em IEEE Transactions on Instrumentation and Measurement}, 71:1--14,
  2022.

\bibitem{lee2021flying}
Thomas Lee, Susan Mckeever, and Jane Courtney.
\newblock Flying free: A research overview of deep learning in drone navigation
  autonomy.
\newblock {\em Drones}, 5(2):52, 2021.

\bibitem{niroui2019deep}
Farzad Niroui, Kaicheng Zhang, Zendai Kashino, and Goldie Nejat.
\newblock Deep reinforcement learning robot for search and rescue applications:
  Exploration in unknown cluttered environments.
\newblock {\em IEEE Robotics and Automation Letters}, 4(2):610--617, 2019.

\bibitem{muscettola2002idea}
Nicola Muscettola, Gregory~A Dorais, Chuck Fry, Richard Levinson, Christian
  Plaunt, and Daniel Clancy.
\newblock Idea: Planning at the core of autonomous reactive agents.
\newblock In {\em Sixth International Conference on AI Planning and
  Scheduling}, pages~,, 2002.

\bibitem{haith2013model}
Adrian~M Haith and John~W Krakauer.
\newblock Model-based and model-free mechanisms of human motor learning.
\newblock In {\em Progress in motor control: Neural, computational and dynamic
  approaches}, pages 1--21. Springer, 2013.

\bibitem{wolek2017model}
Artur Wolek and Craig~A Woolsey.
\newblock Model-based path planning.
\newblock {\em Sensing and Control for Autonomous Vehicles: Applications to
  Land, Water and Air Vehicles}, pages 183--206, 2017.

\bibitem{kontoudis2019kinodynamic}
George~P Kontoudis and Kyriakos~G Vamvoudakis.
\newblock Kinodynamic motion planning with continuous-time q-learning: An
  online, model-free, and safe navigation framework.
\newblock {\em IEEE transactions on neural networks and learning systems},
  30(12):3803--3817, 2019.

\bibitem{zhu2021deep}
Kai Zhu and Tao Zhang.
\newblock Deep reinforcement learning based mobile robot navigation: A review.
\newblock {\em Tsinghua Science and Technology}, 26(5):674--691, 2021.

\bibitem{abbeel2008apprenticeship}
Pieter Abbeel, Dmitri Dolgov, Andrew~Y Ng, and Sebastian Thrun.
\newblock Apprenticeship learning for motion planning with application to
  parking lot navigation.
\newblock In {\em 2008 IEEE/RSJ International Conference on Intelligent Robots
  and Systems}, pages 1083--1090. IEEE, 2008.

\bibitem{el2021systematic}
Fatma El-Zahraa El-Taher, Ayman Taha, Jane Courtney, and Susan Mckeever.
\newblock A systematic review of urban navigation systems for visually impaired
  people.
\newblock {\em Sensors}, 21(9):3103, 2021.

\bibitem{wang2022applications}
Tianhai Wang, Bin Chen, Zhenqian Zhang, Han Li, and Man Zhang.
\newblock Applications of machine vision in agricultural robot navigation: A
  review.
\newblock {\em Computers and Electronics in Agriculture}, 198:107085, 2022.

\bibitem{o2018deep}
Niall O’Mahony, Sean Campbell, Lenka Krpalkova, Daniel Riordan, Joseph Walsh,
  Aidan Murphy, and Conor Ryan.
\newblock Deep learning for visual navigation of unmanned ground vehicles: A
  review.
\newblock In {\em 2018 29th Irish Signals and Systems Conference (ISSC)}, pages
  1--6. IEEE, 2018.

\bibitem{almahamid2022autonomous}
Fadi AlMahamid and Katarina Grolinger.
\newblock Autonomous unmanned aerial vehicle navigation using reinforcement
  learning: A systematic review.
\newblock {\em Engineering Applications of Artificial Intelligence},
  115:105321, 2022.

\bibitem{badrloo2022image}
Samira Badrloo, Masood Varshosaz, Saied Pirasteh, and Jonathan Li.
\newblock Image-based obstacle detection methods for the safe navigation of
  unmanned vehicles: A review.
\newblock {\em Remote Sensing}, 14(15):3824, 2022.

\bibitem{song2022deep}
Jianing Song, Duarte Rondao, and Nabil Aouf.
\newblock Deep learning-based spacecraft relative navigation methods: A survey.
\newblock {\em Acta Astronautica}, 191:22--40, 2022.

\bibitem{turan2022autonomous}
Erdem Turan, Stefano Speretta, and Eberhard Gill.
\newblock Autonomous navigation for deep space small satellites: Scientific and
  technological advances.
\newblock {\em Acta Astronautica}, 2022.

\bibitem{chang2023hierarchical}
Lu~Chang, Liang Shan, Weilong Zhang, and Yuewei Dai.
\newblock Hierarchical multi-robot navigation and formation in unknown
  environments via deep reinforcement learning and distributed optimization.
\newblock {\em Robotics and Computer-Integrated Manufacturing}, 83:102570,
  2023.

\bibitem{hou2023hinnet}
Xinyu Hou and Jeroen~HM Bergmann.
\newblock Hinnet: Inertial navigation with head-mounted sensors using a neural
  network.
\newblock {\em Engineering Applications of Artificial Intelligence},
  123:106066, 2023.

\bibitem{zhuang2023multi}
Yuan Zhuang, Xiao Sun, You Li, Jianzhu Huai, Luchi Hua, Xiansheng Yang,
  Xiaoxiang Cao, Peng Zhang, Yue Cao, Longning Qi, et~al.
\newblock Multi-sensor integrated navigation/positioning systems using data
  fusion: From analytics-based to learning-based approaches.
\newblock {\em Information Fusion}, 95:62--90, 2023.

\bibitem{yu2023study}
Jiya Yu, Jiye Zhang, Aijing Shu, Yujie Chen, Jianneng Chen, Yongjie Yang, Wei
  Tang, and Yanchao Zhang.
\newblock Study of convolutional neural network-based semantic segmentation
  methods on edge intelligence devices for field agricultural robot navigation
  line extraction.
\newblock {\em Computers and Electronics in Agriculture}, 209:107811, 2023.

\bibitem{huang2023goal}
Wenhui Huang, Yanxin Zhou, Xiangkun He, and Chen Lv.
\newblock Goal-guided transformer-enabled reinforcement learning for efficient
  autonomous navigation.
\newblock {\em arXiv preprint arXiv:2301.00362}, 2023.

\bibitem{kuriakose2023deepnavi}
Bineeth Kuriakose, Raju Shrestha, and Frode~Eika Sandnes.
\newblock Deepnavi: A deep learning based smartphone navigation assistant for
  people with visual impairments.
\newblock {\em Expert Systems with Applications}, 212:118720, 2023.

\bibitem{sleaman2023monocular}
Walead~Kaled Sleaman, Alaa~Ali Hameed, and Akhtar Jamil.
\newblock Monocular vision with deep neural networks for autonomous mobile
  robots navigation.
\newblock {\em Optik}, 272:170162, 2023.

\bibitem{ludwig2023urwalking}
Bernd Ludwig, Gregor Donabauer, Dominik Ramsauer, and Karema~al Subari.
\newblock Urwalking: Indoor navigation for research and daily use.
\newblock {\em KI-K{\"u}nstliche Intelligenz}, pages 1--8, 2023.

\bibitem{yuan2023simple}
Kaiwen Yuan and Z~Jane Wang.
\newblock A simple self-supervised imu denoising method for inertial aided
  navigation.
\newblock {\em IEEE Robotics and Automation Letters}, 2023.

\bibitem{dang2023multi}
Thai-Viet Dang and Ngoc-Tam Bui.
\newblock Multi-scale fully convolutional network-based semantic segmentation
  for mobile robot navigation.
\newblock {\em Electronics}, 12(3):533, 2023.

\bibitem{martinez2023drone}
Alejandro Martinez~Acosta.
\newblock Drone navigation using octrees and object recognition for intelligent
  inspections.
\newblock In {\em AIAA SCITECH 2023 Forum}, page 1075, 2023.

\bibitem{liu2022deep}
Yaohua Liu, Qingsong Luo, and Yimin Zhou.
\newblock Deep learning-enabled fusion to bridge gps outages for ins/gps
  integrated navigation.
\newblock {\em IEEE Sensors Journal}, 22(9):8974--8985, 2022.

\bibitem{ayyalasomayajula2020deep}
Roshan Ayyalasomayajula, Aditya Arun, Chenfeng Wu, Sanatan Sharma,
  Abhishek~Rajkumar Sethi, Deepak Vasisht, and Dinesh Bharadia.
\newblock Deep learning based wireless localization for indoor navigation.
\newblock In {\em Proceedings of the 26th Annual International Conference on
  Mobile Computing and Networking}, pages 1--14, 2020.

\bibitem{liu2021efficient}
Zhijian Liu, Alexander Amini, Sibo Zhu, Sertac Karaman, Song Han, and Daniela~L
  Rus.
\newblock Efficient and robust lidar-based end-to-end navigation.
\newblock In {\em 2021 IEEE International Conference on Robotics and Automation
  (ICRA)}, pages 13247--13254. IEEE, 2021.

\bibitem{golroudbari2023end}
Arman~Asgharpoor Golroudbari and Mohammad~Hossein Sabour.
\newblock End-to-end deep learning framework for real-time inertial attitude
  estimation using 6dof imu.
\newblock {\em arXiv preprint arXiv:2302.06037}, 2023.

\bibitem{carrio2017review}
Adrian Carrio, Carlos Sampedro, Alejandro Rodriguez-Ramos, and Pascual Campoy.
\newblock A review of deep learning methods and applications for unmanned
  aerial vehicles.
\newblock {\em Journal of Sensors}, 2017, 2017.

\bibitem{wong2017adaptive}
Cuebong Wong, Erfu Yang, Xiu-Tian Yan, and Dongbing Gu.
\newblock Adaptive and intelligent navigation of autonomous planetary
  rovers—a survey.
\newblock In {\em 2017 NASA/ESA Conference on Adaptive Hardware and Systems
  (AHS)}, pages 237--244. IEEE, 2017.

\bibitem{alom2019state}
Md~Zahangir Alom, Tarek~M Taha, Chris Yakopcic, Stefan Westberg, Paheding
  Sidike, Mst~Shamima Nasrin, Mahmudul Hasan, Brian~C Van~Essen, Abdul~AS
  Awwal, and Vijayan~K Asari.
\newblock A state-of-the-art survey on deep learning theory and architectures.
\newblock {\em electronics}, 8(3):292, 2019.

\bibitem{ejaz2019autonomous}
Muhammad~Mudassir Ejaz, Tong~Boon Tang, and Cheng-Kai Lu.
\newblock Autonomous visual navigation using deep reinforcement learning: An
  overview.
\newblock In {\em 2019 IEEE Student Conference on Research and Development
  (SCOReD)}, pages 294--299. IEEE, 2019.

\bibitem{emmert2020introductory}
Frank Emmert-Streib, Zhen Yang, Han Feng, Shailesh Tripathi, and Matthias
  Dehmer.
\newblock An introductory review of deep learning for prediction models with
  big data.
\newblock {\em Frontiers in Artificial Intelligence}, 3:4, 2020.

\bibitem{grigorescu2020survey}
Sorin Grigorescu, Bogdan Trasnea, Tiberiu Cocias, and Gigel Macesanu.
\newblock A survey of deep learning techniques for autonomous driving.
\newblock {\em Journal of Field Robotics}, 37(3):362--386, 2020.

\bibitem{li2020inertial}
You Li, Ruizhi Chen, Xiaoji Niu, Yuan Zhuang, Zhouzheng Gao, Xin Hu, and Naser
  El-Sheimy.
\newblock Inertial sensing meets artificial intelligence: Opportunity or
  challenge?
\newblock {\em arXiv preprint arXiv:2007.06727}, 2020.

\bibitem{ni2020survey}
Jianjun Ni, Yinan Chen, Yan Chen, Jinxiu Zhu, Deena Ali, and Weidong Cao.
\newblock A survey on theories and applications for self-driving cars based on
  deep learning methods.
\newblock {\em Applied Sciences}, 10(8):2749, 2020.

\bibitem{chen2020survey}
Changhao Chen, Bing Wang, Chris~Xiaoxuan Lu, Niki Trigoni, and Andrew Markham.
\newblock A survey on deep learning for localization and mapping: Towards the
  age of spatial machine intelligence, 2020.

\bibitem{li2021survey}
Zewen Li, Fan Liu, Wenjie Yang, Shouheng Peng, and Jun Zhou.
\newblock A survey of convolutional neural networks: analysis, applications,
  and prospects.
\newblock {\em IEEE transactions on neural networks and learning systems},
  2021.

\bibitem{roy2021survey}
Priya Roy and Chandreyee Chowdhury.
\newblock A survey of machine learning techniques for indoor localization and
  navigation systems.
\newblock {\em Journal of Intelligent \& Robotic Systems}, 101(3):63, 2021.

\bibitem{silvestrini2022deep}
Stefano Silvestrini and Mich{\`e}le Lavagna.
\newblock Deep learning and artificial neural networks for spacecraft dynamics,
  navigation and control.
\newblock {\em Drones}, 6(10):270, 2022.

\bibitem{qin2022survey}
Jiangying Qin, Ming Li, Deren Li, Jiageng Zhong, and Ke~Yang.
\newblock A survey on visual navigation and positioning for autonomous uuvs.
\newblock {\em Remote Sensing}, 14(15):3794, 2022.

\bibitem{wen2022deep}
Li-Hua Wen and Kang-Hyun Jo.
\newblock Deep learning-based perception systems for autonomous driving: A
  comprehensive survey.
\newblock {\em Neurocomputing}, 2022.

\bibitem{tang2022perception}
Yang Tang, Chaoqiang Zhao, Jianrui Wang, Chongzhen Zhang, Qiyu Sun, Wei~Xing
  Zheng, Wenli Du, Feng Qian, and J{\"u}rgen Kurths.
\newblock Perception and navigation in autonomous systems in the era of
  learning: A survey.
\newblock {\em IEEE Transactions on Neural Networks and Learning Systems},
  2022.

\bibitem{goodfellow2016deep}
Ian Goodfellow, Yoshua Bengio, and Aaron Courville.
\newblock {\em Deep learning}.
\newblock MIT press, 2016.

\bibitem{mcculloch1943logical}
Warren~S McCulloch and Walter Pitts.
\newblock A logical calculus of the ideas immanent in nervous activity.
\newblock {\em The bulletin of mathematical biophysics}, 5:115--133, 1943.

\bibitem{rosenblatt1958perceptron}
Frank Rosenblatt.
\newblock The perceptron: a probabilistic model for information storage and
  organization in the brain.
\newblock {\em Psychological review}, 65(6):386, 1958.

\bibitem{rumelhart1986learning}
David~E Rumelhart, Geoffrey~E Hinton, and Ronald~J Williams.
\newblock Learning representations by back-propagating errors.
\newblock {\em nature}, 323(6088):533--536, 1986.

\bibitem{lecun1998gradient}
Yann LeCun, L{\'e}on Bottou, Yoshua Bengio, and Patrick Haffner.
\newblock Gradient-based learning applied to document recognition.
\newblock {\em Proceedings of the IEEE}, 86(11):2278--2324, 1998.

\bibitem{hinton2012deep}
Geoffrey Hinton, Li~Deng, Dong Yu, George~E Dahl, Abdel-rahman Mohamed, Navdeep
  Jaitly, Andrew Senior, Vincent Vanhoucke, Patrick Nguyen, Tara~N Sainath,
  et~al.
\newblock Deep neural networks for acoustic modeling in speech recognition: The
  shared views of four research groups.
\newblock {\em IEEE Signal processing magazine}, 29(6):82--97, 2012.

\bibitem{krizhevsky2017imagenet}
Alex Krizhevsky, Ilya Sutskever, and Geoffrey~E Hinton.
\newblock Imagenet classification with deep convolutional neural networks.
\newblock {\em Communications of the ACM}, 60(6):84--90, 2017.

\bibitem{otter2020survey}
Daniel~W Otter, Julian~R Medina, and Jugal~K Kalita.
\newblock A survey of the usages of deep learning for natural language
  processing.
\newblock {\em IEEE transactions on neural networks and learning systems},
  32(2):604--624, 2020.

\bibitem{deng2014ensemble}
Li~Deng and John Platt.
\newblock Ensemble deep learning for speech recognition.
\newblock In {\em Proc. interspeech}, pages~,, 2014.

\bibitem{rao2018deep}
Qing Rao and Jelena Frtunikj.
\newblock Deep learning for self-driving cars: Chances and challenges.
\newblock In {\em Proceedings of the 1st international workshop on software
  engineering for AI in autonomous systems}, pages 35--38, 2018.

\bibitem{esteva2019guide}
Andre Esteva, Alexandre Robicquet, Bharath Ramsundar, Volodymyr Kuleshov, Mark
  DePristo, Katherine Chou, Claire Cui, Greg Corrado, Sebastian Thrun, and Jeff
  Dean.
\newblock A guide to deep learning in healthcare.
\newblock {\em Nature medicine}, 25(1):24--29, 2019.

\bibitem{angermueller2016deep}
Christof Angermueller, Tanel P{\"a}rnamaa, Leopold Parts, and Oliver Stegle.
\newblock Deep learning for computational biology.
\newblock {\em Molecular systems biology}, 12(7):878, 2016.

\bibitem{justesen2019deep}
Niels Justesen, Philip Bontrager, Julian Togelius, and Sebastian Risi.
\newblock Deep learning for video game playing.
\newblock {\em IEEE Transactions on Games}, 12(1):1--20, 2019.

\bibitem{aggarwal2021generative}
Alankrita Aggarwal, Mamta Mittal, and Gopi Battineni.
\newblock Generative adversarial network: An overview of theory and
  applications.
\newblock {\em International Journal of Information Management Data Insights},
  1(1):100004, 2021.

\bibitem{vaswani2017attention}
Ashish Vaswani, Noam Shazeer, Niki Parmar, Jakob Uszkoreit, Llion Jones,
  Aidan~N Gomez, {\L}ukasz Kaiser, and Illia Polosukhin.
\newblock Attention is all you need.
\newblock {\em Advances in neural information processing systems}, 30, 2017.

\bibitem{openai2023gpt}
OpenAI.
\newblock Gpt-4 technical report.
\newblock {\em arXiv}, 2023.

\bibitem{ruff2021alphafold}
Kiersten~M Ruff and Rohit~V Pappu.
\newblock Alphafold and implications for intrinsically disordered proteins.
\newblock {\em Journal of Molecular Biology}, 433(20):167208, 2021.

\bibitem{silver2016mastering}
David Silver, Aja Huang, Chris~J Maddison, Arthur Guez, Laurent Sifre, George
  Van Den~Driessche, Julian Schrittwieser, Ioannis Antonoglou, Veda
  Panneershelvam, Marc Lanctot, et~al.
\newblock Mastering the game of go with deep neural networks and tree search.
\newblock {\em nature}, 529(7587):484--489, 2016.

\bibitem{schmidhuber2015deep}
J{\"u}rgen Schmidhuber.
\newblock Deep learning in neural networks: An overview.
\newblock {\em Neural networks}, 61:85--117, 2015.

\bibitem{hinton2006fast}
Geoffrey~E Hinton, Simon Osindero, and Yee-Whye Teh.
\newblock A fast learning algorithm for deep belief nets.
\newblock {\em Neural computation}, 18(7):1527--1554, 2006.

\bibitem{hochreiter1998vanishing}
Sepp Hochreiter.
\newblock The vanishing gradient problem during learning recurrent neural nets
  and problem solutions.
\newblock {\em International Journal of Uncertainty, Fuzziness and
  Knowledge-Based Systems}, 6(02):107--116, 1998.

\bibitem{cho2014learning}
Kyunghyun Cho, Bart Van~Merri{\"e}nboer, Caglar Gulcehre, Dzmitry Bahdanau,
  Fethi Bougares, Holger Schwenk, and Yoshua Bengio.
\newblock Learning phrase representations using rnn encoder-decoder for
  statistical machine translation.
\newblock {\em arXiv preprint arXiv:1406.1078}, 2014.

\bibitem{hochreiter1997long}
Sepp Hochreiter and J{\"u}rgen Schmidhuber.
\newblock Long short-term memory.
\newblock {\em Neural computation}, 9(8):1735--1780, 1997.

\bibitem{goodfellow2020generative}
Ian Goodfellow, Jean Pouget-Abadie, Mehdi Mirza, Bing Xu, David Warde-Farley,
  Sherjil Ozair, Aaron Courville, and Yoshua Bengio.
\newblock Generative adversarial networks.
\newblock {\em Communications of the ACM}, 63(11):139--144, 2020.

\bibitem{arjovsky2017wasserstein}
Martin Arjovsky, Soumith Chintala, and L{\'e}on Bottou.
\newblock Wasserstein generative adversarial networks.
\newblock In {\em International conference on machine learning}, pages
  214--223. PMLR, 2017.

\bibitem{gulrajani2017improved}
Ishaan Gulrajani, Faruk Ahmed, Martin Arjovsky, Vincent Dumoulin, and Aaron~C
  Courville.
\newblock Improved training of wasserstein gans.
\newblock {\em Advances in neural information processing systems}, 30, 2017.

\bibitem{berthelot2017began}
David Berthelot, Thomas Schumm, and Luke Metz.
\newblock Began: Boundary equilibrium generative adversarial networks.
\newblock {\em arXiv preprint arXiv:1703.10717}, 2017.

\bibitem{he2022improved}
Deqiang He, Yefeng Qiu, Jian Miao, Zhiheng Zou, Kai Li, Chonghui Ren, and
  Guoqiang Shen.
\newblock Improved mask r-cnn for obstacle detection of rail transit.
\newblock {\em Measurement}, 190:110728, 2022.

\bibitem{aftf2019indoor}
Mouna Aftf, Riadh Ayachi, Yahia Said, Edwige Pissaloux, and Mohamed Atri.
\newblock Indoor object c1assification for autonomous navigation assistance
  based on deep cnn model.
\newblock In {\em 2019 IEEE International Symposium on Measurements \&
  Networking (M\&N)}, pages 1--4. IEEE, 2019.

\bibitem{wang2020cnn}
Wei Wang, Hui Lin, and Junshu Wang.
\newblock Cnn based lane detection with instance segmentation in edge-cloud
  computing.
\newblock {\em Journal of Cloud Computing}, 9:1--10, 2020.

\bibitem{li2022traffic}
Xiaomei Li, Zhijiang Xie, Xiong Deng, Yanxue Wu, and Yangjun Pi.
\newblock Traffic sign detection based on improved faster r-cnn for autonomous
  driving.
\newblock {\em The Journal of Supercomputing}, pages 1--21, 2022.

\bibitem{kouris2018learning}
Alexandros Kouris and Christos-Savvas Bouganis.
\newblock Learning to fly by myself: A self-supervised cnn-based approach for
  autonomous navigation.
\newblock In {\em 2018 IEEE/RSJ International Conference on Intelligent Robots
  and Systems (IROS)}, pages 1--9. IEEE, 2018.

\bibitem{gong2021deepnav}
Jian Gong, Ju~Ren, and Yaoxue Zhang.
\newblock Deepnav: A scalable and plug-and-play indoor navigation system based
  on visual cnn.
\newblock {\em Peer-to-Peer Networking and Applications}, 14:3718--3736, 2021.

\bibitem{chumuang2022feature}
Narumol Chumuang, Adil Farooq, Muhammad Irfan, Sumair Aziz, and Moomal Qureshi.
\newblock Feature matching and deep learning models for attitude estimation on
  a micro-aerial vehicle.
\newblock In {\em 2022 International Conference on Cybernetics and Innovations
  (ICCI)}, pages 1--6. IEEE, 2022.

\bibitem{brossard2020denoising}
Martin Brossard, Silvere Bonnabel, and Axel Barrau.
\newblock Denoising imu gyroscopes with deep learning for open-loop attitude
  estimation.
\newblock {\em IEEE Robotics and Automation Letters}, 5(3):4796--4803, 2020.

\bibitem{topini2020lstm}
Edoardo Topini, Alberto Topini, Matteo Franchi, Alessandro Bucci, Nicola
  Secciani, Alessandro Ridolfi, and Benedetto Allotta.
\newblock Lstm-based dead reckoning navigation for autonomous underwater
  vehicles.
\newblock In {\em Global Oceans 2020: Singapore--US Gulf Coast}, pages 1--7.
  IEEE, 2020.

\bibitem{tong2019cascade}
Qianqian Tong, Xiaosa Li, Kai Lin, Caizi Li, Weixin Si, and Zhiyong Yuan.
\newblock Cascade-lstm-based visual-inertial navigation for magnetic levitation
  haptic interaction.
\newblock {\em IEEE Network}, 33(3):74--80, 2019.

\bibitem{du2021vtnet}
Heming Du, Xin Yu, and Liang Zheng.
\newblock Vtnet: Visual transformer network for object goal navigation.
\newblock {\em arXiv preprint arXiv:2105.09447}, 2021.

\bibitem{yu2020spatio}
Cunjun Yu, Xiao Ma, Jiawei Ren, Haiyu Zhao, and Shuai Yi.
\newblock Spatio-temporal graph transformer networks for pedestrian trajectory
  prediction.
\newblock In {\em Computer Vision--ECCV 2020: 16th European Conference,
  Glasgow, UK, August 23--28, 2020, Proceedings, Part XII 16}, pages 507--523.
  Springer, 2020.

\bibitem{roy2019vehicle}
Debaditya Roy, Tetsuhiro Ishizaka, C~Krishna Mohan, and Atsushi Fukuda.
\newblock Vehicle trajectory prediction at intersections using interaction
  based generative adversarial networks.
\newblock In {\em 2019 IEEE Intelligent transportation systems conference
  (ITSC)}, pages 2318--2323. IEEE, 2019.

\bibitem{dimas2019obstacle}
George Dimas, Charis Ntakolia, and Dimitris~K Iakovidis.
\newblock Obstacle detection based on generative adversarial networks and fuzzy
  sets for computer-assisted navigation.
\newblock In {\em Engineering Applications of Neural Networks: 20th
  International Conference, EANN 2019, Xersonisos, Crete, Greece, May 24-26,
  2019, Proceedings 20}, pages 533--544. Springer, 2019.

\bibitem{mohammadi2018path}
Mehdi Mohammadi, Ala Al-Fuqaha, and Jun-Seok Oh.
\newblock Path planning in support of smart mobility applications using
  generative adversarial networks.
\newblock In {\em 2018 IEEE International Conference on Internet of Things
  (iThings) and IEEE Green Computing and Communications (GreenCom) and IEEE
  Cyber, Physical and Social Computing (CPSCom) and IEEE Smart Data
  (SmartData)}, pages 878--885. IEEE, 2018.

\bibitem{fukushima1975cognitron}
Kunihiko Fukushima.
\newblock Cognitron: A self-organizing multilayered neural network.
\newblock {\em Biological cybernetics}, 20(3-4):121--136, 1975.

\bibitem{maas2013rectifier}
Andrew~L Maas, Awni~Y Hannun, Andrew~Y Ng, et~al.
\newblock Rectifier nonlinearities improve neural network acoustic models.
\newblock In {\em Proc. icml}, volume~30, page~3. Atlanta, Georgia, USA, 2013.

\bibitem{clevert2015fast}
Djork-Arn{\'e} Clevert, Thomas Unterthiner, and Sepp Hochreiter.
\newblock Fast and accurate deep network learning by exponential linear units
  (elus).
\newblock {\em arXiv preprint arXiv:1511.07289}, 2015.

\bibitem{bridle1990probabilistic}
John~S Bridle.
\newblock Probabilistic interpretation of feedforward classification network
  outputs, with relationships to statistical pattern recognition.
\newblock In {\em Neurocomputing: Algorithms, architectures and applications},
  pages 227--236. Springer, 1990.

\bibitem{bridle1989training}
John Bridle.
\newblock Training stochastic model recognition algorithms as networks can lead
  to maximum mutual information estimation of parameters.
\newblock {\em Advances in neural information processing systems}, 2, 1989.

\bibitem{dugas2000incorporating}
Charles Dugas, Yoshua Bengio, Fran{\c{c}}ois B{\'e}lisle, Claude Nadeau, and
  Ren{\'e} Garcia.
\newblock Incorporating second-order functional knowledge for better option
  pricing.
\newblock {\em Advances in neural information processing systems}, 13, 2000.

\bibitem{ramachandran2017searching}
Prajit Ramachandran, Barret Zoph, and Quoc~V Le.
\newblock Searching for activation functions.
\newblock {\em arXiv preprint arXiv:1710.05941}, 2017.

\bibitem{xu2015empirical}
Bing Xu, Naiyan Wang, Tianqi Chen, and Mu~Li.
\newblock Empirical evaluation of rectified activations in convolutional
  network.
\newblock {\em arXiv preprint arXiv:1505.00853}, 2015.

\bibitem{misra2019mish}
Diganta Misra.
\newblock Mish: A self regularized non-monotonic neural activation function.
\newblock {\em arXiv preprint arXiv:1908.08681}, 4(2):10--48550, 2019.

\bibitem{nair2010rectified}
Vinod Nair and Geoffrey~E Hinton.
\newblock Rectified linear units improve restricted boltzmann machines.
\newblock In {\em Proceedings of the 27th international conference on machine
  learning (ICML-10)}, pages 807--814, 2010.

\bibitem{he2015delving}
Kaiming He, Xiangyu Zhang, Shaoqing Ren, and Jian Sun.
\newblock Delving deep into rectifiers: Surpassing human-level performance on
  imagenet classification.
\newblock In {\em Proceedings of the IEEE international conference on computer
  vision}, pages 1026--1034, 2015.

\bibitem{hendrycks2016baseline}
Dan Hendrycks and Kevin Gimpel.
\newblock A baseline for detecting misclassified and out-of-distribution
  examples in neural networks.
\newblock {\em arXiv preprint arXiv:1610.02136}, 2016.

\bibitem{klambauer2017self}
G{\"u}nter Klambauer, Thomas Unterthiner, Andreas Mayr, and Sepp Hochreiter.
\newblock Self-normalizing neural networks.
\newblock {\em Advances in neural information processing systems}, 30, 2017.

\bibitem{roy2019lisht}
Swalpa~Kumar Roy, Suvojit Manna, Shiv~Ram Dubey, and Bidyut~Baran Chaudhuri.
\newblock Lisht: Non-parametric linearly scaled hyperbolic tangent activation
  function for neural networks.
\newblock {\em arXiv preprint arXiv:1901.05894}, 2019.

\bibitem{ziyin2020neural}
Liu Ziyin, Tilman Hartwig, and Masahito Ueda.
\newblock Neural networks fail to learn periodic functions and how to fix it.
\newblock {\em Advances in Neural Information Processing Systems},
  33:1583--1594, 2020.

\bibitem{hu2019novel}
Zhenming Hu, Weibo Fang, Tong Gou, Wenshuai Wu, Jiumei Hu, Shufang Zhou, and
  Ying Mu.
\newblock A novel method based on a mask r-cnn model for processing dpcr
  images.
\newblock {\em Analytical Methods}, 11(27):3410--3418, 2019.

\bibitem{matsumoto1990several}
T~Matsumoto, T~Yokohama, H~Suzuki, R~Furukawa, A~Oshimoto, T~Shimmi,
  Y~Matsushita, T~Seo, and LO~Chua.
\newblock Several image processing examples by cnn.
\newblock In {\em IEEE International Workshop on Cellular Neural Networks and
  their Applications}, pages 100--111. IEEE, 1990.

\bibitem{arena2003image}
Paolo Arena, Adriano Basile, Maide Bucolo, and Luigi Fortuna.
\newblock Image processing for medical diagnosis using cnn.
\newblock {\em Nuclear Instruments and Methods in Physics Research Section A:
  Accelerators, Spectrometers, Detectors and Associated Equipment},
  497(1):174--178, 2003.

\bibitem{tian2020road}
Jin Tian, Jiazheng Yuan, and Hongzhe Liu.
\newblock Road marking detection based on mask r-cnn instance segmentation
  model.
\newblock In {\em 2020 international conference on computer vision, image and
  deep learning (CVIDL)}, pages 246--249. IEEE, 2020.

\bibitem{lima2022road}
Aklima~Akter Lima, Md~Mohsin Kabir, Sujoy~Chandra Das, Md~Nahid Hasan, and
  MF~Mridha.
\newblock Road sign detection using variants of yolo and r-cnn: An analysis
  from the perspective of bangladesh.
\newblock In {\em Proceedings of the International Conference on Big Data, IoT,
  and Machine Learning: BIM 2021}, pages 555--565. Springer, 2022.

\bibitem{prabhakar2017obstacle}
Gowdham Prabhakar, Binsu Kailath, Sudha Natarajan, and Rajesh Kumar.
\newblock Obstacle detection and classification using deep learning for
  tracking in high-speed autonomous driving.
\newblock In {\em 2017 IEEE region 10 symposium (TENSYMP)}, pages 1--6. IEEE,
  2017.

\bibitem{hu2018embedding}
Chaowei Hu, Yunpeng Wang, Guizhen Yu, Zhangyu Wang, Ao~Lei, and Zhehua Hu.
\newblock Embedding cnn-based fast obstacles detection for autonomous vehicles.
\newblock Technical report, SAE Technical Paper, 2018.

\bibitem{lamberti2023bio}
Lorenzo Lamberti, Luca Bompani, Victor~Javier Kartsch, Manuele Rusci, Daniele
  Palossi, and Luca Benini.
\newblock Bio-inspired autonomous exploration policies with cnn-based object
  detection on nano-drones.
\newblock {\em arXiv preprint arXiv:2301.12175}, 2023.

\bibitem{shustanov2017cnn}
Alexander Shustanov and Pavel Yakimov.
\newblock Cnn design for real-time traffic sign recognition.
\newblock {\em Procedia engineering}, 201:718--725, 2017.

\bibitem{dewi2022deep}
Christine Dewi, Rung-Ching Chen, Xiaoyi Jiang, and Hui Yu.
\newblock Deep convolutional neural network for enhancing traffic sign
  recognition developed on yolo v4.
\newblock {\em Multimedia Tools and Applications}, 81(26):37821--37845, 2022.

\bibitem{paravarzar2020motion}
Shahrokh Paravarzar and Belqes Mohammad.
\newblock Motion prediction on self-driving cars: A review.
\newblock {\em arXiv preprint arXiv:2011.03635}, 2020.

\bibitem{fayyad2020deep}
Jamil Fayyad, Mohammad~A Jaradat, Dominique Gruyer, and Homayoun Najjaran.
\newblock Deep learning sensor fusion for autonomous vehicle perception and
  localization: A review.
\newblock {\em Sensors}, 20(15):4220, 2020.

\bibitem{shuttleworth2019sae}
Jennifer Shuttleworth.
\newblock Sae standards news: J3016 automated-driving graphic update.
\newblock {\em SAE International}, 2019.

\bibitem{clarke2014understanding}
Roger Clarke.
\newblock Understanding the drone epidemic.
\newblock {\em Computer Law \& Security Review}, 30(3):230--246, 2014.

\bibitem{merkert2020revolution}
Rico Merkert and James Bushell.
\newblock Revolution or epidemic? a systematic literature review on the
  effective control of airborne drones.
\newblock 2020.

\bibitem{radoglou2020compilation}
Panagiotis Radoglou-Grammatikis, Panagiotis Sarigiannidis, Thomas Lagkas, and
  Ioannis Moscholios.
\newblock A compilation of uav applications for precision agriculture.
\newblock {\em Computer Networks}, 172:107148, 2020.

\bibitem{alzahrani2020uav}
Bander Alzahrani, Omar~Sami Oubbati, Ahmed Barnawi, Mohammed Atiquzzaman, and
  Daniyal Alghazzawi.
\newblock Uav assistance paradigm: State-of-the-art in applications and
  challenges.
\newblock {\em Journal of Network and Computer Applications}, 166:102706, 2020.

\bibitem{MK4}
Amazon.
\newblock Mk4, amazon prime air drone, 2022.

\bibitem{mohsan2022towards}
Syed Agha~Hassnain Mohsan, Muhammad~Asghar Khan, Fazal Noor, Insaf Ullah, and
  Mohammed~H Alsharif.
\newblock Towards the unmanned aerial vehicles (uavs): A comprehensive review.
\newblock {\em Drones}, 6(6):147, 2022.

\bibitem{rezwan2022artificial}
Sifat Rezwan and Wooyeol Choi.
\newblock Artificial intelligence approaches for uav navigation: Recent
  advances and future challenges.
\newblock {\em IEEE Access}, 2022.

\bibitem{proencca2020deep}
Pedro~F Proen{\c{c}}a and Yang Gao.
\newblock Deep learning for spacecraft pose estimation from photorealistic
  rendering.
\newblock In {\em 2020 IEEE International Conference on Robotics and Automation
  (ICRA)}, pages 6007--6013. IEEE, 2020.

\bibitem{park2022robust}
Tae~Ha Park and Simone D'Amico.
\newblock Robust multi-task learning and online refinement for spacecraft pose
  estimation across domain gap.
\newblock {\em arXiv preprint arXiv:2203.04275}, 2022.

\bibitem{santos2022machine}
Rogerio~R Santos, Domingos~A Rade, and Ijar~M da~Fonseca.
\newblock A machine learning strategy for optimal path planning of space
  robotic manipulator in on-orbit servicing.
\newblock {\em Acta Astronautica}, 191:41--54, 2022.

\bibitem{uriot2022spacecraft}
Thomas Uriot, Dario Izzo, Lu{\'\i}s~F Sim{\~o}es, Rasit Abay, Nils Einecke,
  Sven Rebhan, Jose Martinez-Heras, Francesca Letizia, Jan Siminski, and Klaus
  Merz.
\newblock Spacecraft collision avoidance challenge: Design and results of a
  machine learning competition.
\newblock {\em Astrodynamics}, 6(2):121--140, 2022.

\bibitem{sun2023satellite}
Zibin Sun, Jules Simo, and Shengping Gong.
\newblock Satellite attitude identification and prediction based on neural
  network compensation.
\newblock {\em Space: Science \& Technology}, 2023.

\bibitem{stacey2022robust}
Nathan Stacey and Simone D'Amico.
\newblock Robust autonomous spacecraft navigation and environment
  characterization.
\newblock {\em Authorea Preprints}, 2022.

\bibitem{park2022adaptive}
Tae~Ha Park and Simone D'Amico.
\newblock Adaptive neural network-based unscented kalman filter for spacecraft
  pose tracking at rendezvous.
\newblock {\em arXiv preprint arXiv:2206.03796}, 2022.

\bibitem{gaudet2020deep}
Brian Gaudet, Richard Linares, and Roberto Furfaro.
\newblock Deep reinforcement learning for six degree-of-freedom planetary
  landing.
\newblock {\em Advances in Space Research}, 65(7):1723--1741, 2020.

\bibitem{furfaro2018deep}
Roberto Furfaro, Ilaria Bloise, Marcello Orlandelli, Pierluigi Di~Lizia,
  Francesco Topputo, Richard Linares, et~al.
\newblock Deep learning for autonomous lunar landing.
\newblock {\em Advances in the Astronautical Sciences}, 167:3285--3306, 2018.

\bibitem{downes2021neural}
Lena~M Downes, Ted~J Steiner, and Jonathan~P How.
\newblock Neural network approach to crater detection for lunar terrain
  relative navigation.
\newblock {\em Journal of Aerospace Information Systems}, 18(7):391--403, 2021.

\bibitem{downes2020deep}
Lena Downes, Ted~J Steiner, and Jonathan~P How.
\newblock Deep learning crater detection for lunar terrain relative navigation.
\newblock In {\em AIAA SciTech 2020 Forum}, page 1838, 2020.

\bibitem{tsukamoto2022neural}
Hiroyasu Tsukamoto, Soon-Jo Chung, Benjamin Donitz, Michel Ingham, Declan
  Mages, and Yashwanth~Kumar Nakka.
\newblock Neural-rendezvous: Learning-based robust guidance and control to
  encounter interstellar objects.
\newblock {\em arXiv preprint arXiv:2208.04883}, 2022.

\bibitem{donitz2022interstellar}
Benjamin~PS Donitz, Declan Mages, Hiroyasu Tsukamoto, Peter Dixon, Damon
  Landau, Soon-Jo Chung, Erica Bufanda, Michel Ingham, and Julie
  Castillo-Rogez.
\newblock Interstellar object accessibility and mission design.
\newblock {\em arXiv preprint arXiv:2210.14980}, 2022.

\bibitem{neamati2022learning}
Daniel Neamati, Yashwanth Kumar~K Nakka, and Soon-Jo Chung.
\newblock Learning-based methods to model small body gravity fields for
  proximity operations: Safety and robustness.
\newblock In {\em AIAA SCITECH 2022 Forum}, page 2271, 2022.

\bibitem{nakka2021spacecraft}
Yashwanth~Kumar Nakka.
\newblock {\em Spacecraft Motion Planning and Control under Probabilistic
  Uncertainty for Coordinated Inspection and Safe Learning}.
\newblock PhD thesis, California Institute of Technology, 2021.

\bibitem{moghe2020line}
Rahul Moghe and Renato Zanetti.
\newblock On-line hazard detection algorithm for precision lunar landing using
  semantic segmentation.
\newblock In {\em AIAA Scitech 2020 Forum}, page 0462, 2020.

\bibitem{scorsoglio2023relative}
Andrea Scorsoglio, Roberto Furfaro, Richard Linares, and Mauro Massari.
\newblock Relative motion guidance for near-rectilinear lunar orbits with path
  constraints via actor-critic reinforcement learning.
\newblock {\em Advances in Space Research}, 71(1):316--335, 2023.

\bibitem{guthrie2022image}
Ben Guthrie, Minkwan Kim, Hodei Urrutxua, and Jonathon Hare.
\newblock Image-based attitude determination of co-orbiting satellites using
  deep learning technologies.
\newblock {\em Aerospace Science and Technology}, 120:107232, 2022.

\bibitem{abd2022reliable}
Abd-Elsalam~R Abd-Elhay, Wael~A Murtada, and Mohamed~I Youssef.
\newblock A reliable deep learning approach for time-varying faults
  identification: Spacecraft reaction wheel case study.
\newblock {\em IEEE Access}, 10:75495--75512, 2022.

\bibitem{spantideas2021deep}
Sotirios~T Spantideas, Anastasios~E Giannopoulos, Nikolaos~C Kapsalis, and
  Christos~N Capsalis.
\newblock A deep learning method for modeling the magnetic signature of
  spacecraft equipment using multiple magnetic dipoles.
\newblock {\em IEEE Magnetics Letters}, 12:1--5, 2021.

\bibitem{aldahoul2022rgb}
Nouar AlDahoul, Hezerul~Abdul Karim, and Mhd~Adel Momo.
\newblock Rgb-d based multi-modal deep learning for spacecraft and debris
  recognition.
\newblock {\em Scientific Reports}, 12(1):3924, 2022.

\bibitem{harris2022generation}
Andrew Harris, Trace Valade, Thibaud Teil, and Hanspeter Schaub.
\newblock Generation of spacecraft operations procedures using deep
  reinforcement learning.
\newblock {\em Journal of Spacecraft and Rockets}, 59(2):611--626, 2022.

\bibitem{del2022deep}
Roberto Del~Prete, Alfonso Saveriano, and Alfredo Renga.
\newblock A deep learning-based crater detector for autonomous vision-based
  spacecraft navigation.
\newblock In {\em 2022 IEEE 9th International Workshop on Metrology for
  AeroSpace (MetroAeroSpace)}, pages 231--236. IEEE, 2022.

\bibitem{siew2022space}
Peng~Mun Siew, Daniel Jang, Thomas~G Roberts, and Richard Linares.
\newblock Space-based sensor tasking using deep reinforcement learning.
\newblock {\em The Journal of the Astronautical Sciences}, pages 1--38, 2022.

\bibitem{siew2022optimal}
Peng~Mun Siew and Richard Linares.
\newblock Optimal tasking of ground-based sensors for space situational
  awareness using deep reinforcement learning.
\newblock {\em Sensors}, 22(20):7847, 2022.

\bibitem{oestreich2021autonomous}
Charles~E Oestreich, Richard Linares, and Ravi Gondhalekar.
\newblock Autonomous six-degree-of-freedom spacecraft docking with rotating
  targets via reinforcement learning.
\newblock {\em Journal of Aerospace Information Systems}, 18(7):417--428, 2021.

\bibitem{mahendrakar2023spaceyolo}
Trupti Mahendrakar, Ryan~T White, Markus Wilde, and Madhur Tiwari.
\newblock Spaceyolo: A human-inspired model for real-time, on-board spacecraft
  feature detection.
\newblock {\em arXiv preprint arXiv:2302.00824}, 2023.

\bibitem{park2019towards}
Tae~Ha Park, Sumant Sharma, and Simone D'Amico.
\newblock Towards robust learning-based pose estimation of noncooperative
  spacecraft.
\newblock {\em arXiv preprint arXiv:1909.00392}, 2019.

\bibitem{chen2019satellite}
Bo~Chen, Jiewei Cao, Alvaro Parra, and Tat-Jun Chin.
\newblock Satellite pose estimation with deep landmark regression and nonlinear
  pose refinement.
\newblock In {\em Proceedings of the IEEE/CVF International Conference on
  Computer Vision Workshops}, pages 0--0, 2019.

\bibitem{scorsoglio2020safe}
Andrea Scorsoglio, Andrea D’Ambrosio, Luca Ghilardi, Roberto Furfaro, Brian
  Gaudet, Richard Linares, and Fabio Curti.
\newblock Safe lunar landing via images: A reinforcement meta-learning
  application to autonomous hazard avoidance and landing.
\newblock In {\em Proceedings of the 2020 AAS/AIAA Astrodynamics Specialist
  Conference, Virtual}, pages 9--12, 2020.

\bibitem{elkins2020autonomous}
J~Elkins, R~Sood, and C~Rumpf.
\newblock Autonomous spacecraft attitude control using deep reinforcement
  learning.
\newblock In {\em 71st International Astronautical Congress (IAC)}, volume
  2020, 2020.

\bibitem{singh2022stochastic}
Sandeep~K Singh and John~L Junkins.
\newblock Stochastic learning and extremal-field map based autonomous guidance
  of low-thrust spacecraft.
\newblock {\em Scientific Reports}, 12(1):17774, 2022.

\bibitem{hao2021intelligent}
Zhou Hao, RB~Ashith Shyam, Arunkumar Rathinam, and Yang Gao.
\newblock Intelligent spacecraft visual gnc architecture with the
  state-of-the-art ai components for on-orbit manipulation.
\newblock {\em Frontiers in Robotics and AI}, 8:639327, 2021.

\bibitem{yun2020multi}
Kyongsik Yun, Changrak Choi, Ryan Alimo, Anthony Davis, Linda Forster, Amir
  Rahmani, Muhammad Adil, and Ramtin Madani.
\newblock Multi-agent motion planning using deep learning for space
  applications.
\newblock In {\em ASCEND 2020}, page 4233. 2020.

\bibitem{zhou20213d}
Dong Zhou, Gunaghui Sun, and Xiaopeng Hong.
\newblock 3d visual tracking framework with deep learning for asteroid
  exploration.
\newblock {\em arXiv preprint arXiv:2111.10737}, 2021.

\bibitem{huan2020pose}
Wenxiu Huan, Mingmin Liu, and Qinglei Hu.
\newblock Pose estimation for non-cooperative spacecraft based on deep
  learning.
\newblock In {\em 2020 39th Chinese Control Conference (CCC)}, pages
  3339--3343. IEEE, 2020.

\bibitem{roberts2021deep}
Thomas~G Roberts, Peng~Mun Siew, Daniel Jang, and Richard Linares.
\newblock A deep reinforcement learning application to space-based sensor
  tasking for space situational awareness.
\newblock In {\em Proceedings of the 2021 Advanced Maui Optical and Space
  Surveillance Technologies Conference (AMOS), Wailea Beach Resort, Maui, HI,
  USA}, pages 14--17, 2021.

\bibitem{roberts2021geosynchronous}
Thomas~G Roberts and Richard Linares.
\newblock Geosynchronous satellite maneuver classification via supervised
  machine learning.
\newblock In {\em Advanced Maui Optical and Space Surveillance Technologies
  Conference. Maui, Hawaii}, 2021.

\bibitem{liu2022multi}
Yuan Liu, Ming Zhu, Jing Wang, Xiangji Guo, Yifan Yang, and Jiarong Wang.
\newblock Multi-scale deep neural network based on dilated convolution for
  spacecraft image segmentation.
\newblock {\em Sensors}, 22(11):4222, 2022.

\bibitem{rondao2022chinet}
Duarte Rondao, Nabil Aouf, and Mark~A Richardson.
\newblock Chinet: Deep recurrent convolutional learning for multimodal
  spacecraft pose estimation.
\newblock {\em IEEE Transactions on Aerospace and Electronic Systems}, 2022.

\bibitem{liu2023spacecraft}
Liang Liu, Ling Tian, Zhao Kang, and Tianqi Wan.
\newblock Spacecraft anomaly detection with attention temporal convolution
  networks.
\newblock {\em Neural Computing and Applications}, pages 1--9, 2023.

\bibitem{tian2020deep}
Chunwei Tian, Lunke Fei, Wenxian Zheng, Yong Xu, Wangmeng Zuo, and Chia-Wen
  Lin.
\newblock Deep learning on image denoising: An overview.
\newblock {\em Neural Networks}, 131:251--275, 2020.

\bibitem{jiang2018mems}
Changhui Jiang, Shuai Chen, Yuwei Chen, Boya Zhang, Ziyi Feng, Hui Zhou, and
  Yuming Bo.
\newblock A mems imu de-noising method using long short term memory recurrent
  neural networks (lstm-rnn).
\newblock {\em Sensors}, 18(10):3470, 2018.

\bibitem{han2021hybrid}
Shipeng Han, Zhen Meng, Xingcheng Zhang, and Yuepeng Yan.
\newblock Hybrid deep recurrent neural networks for noise reduction of mems-imu
  with static and dynamic conditions.
\newblock {\em Micromachines}, 12(2):214, 2021.

\bibitem{chen2022towards}
Hua Chen, Tarek~M Taha, and Vamsy~P Chodavarapu.
\newblock Towards improved inertial navigation by reducing errors using deep
  learning methodology.
\newblock {\em Applied Sciences}, 12(7):3645, 2022.

\bibitem{engelsman2022data}
Daniel Engelsman and Itzik Klein.
\newblock Data-driven denoising of accelerometer signals.
\newblock {\em arXiv preprint arXiv:2206.05937}, 2022.

\bibitem{hu2021novel}
Minghuan Hu, Jiandong Mao, Juan Li, Qiang Wang, and Yi~Zhang.
\newblock A novel lidar signal denoising method based on convolutional
  autoencoding deep learning neural network.
\newblock {\em Atmosphere}, 12(11):1403, 2021.

\bibitem{liu2023dlc}
Kangcheng Liu and Muqing Cao.
\newblock Dlc-slam: A robust lidar-slam system with learning-based denoising
  and loop closure.
\newblock {\em IEEE/ASME Transactions on Mechatronics}, 2023.

\bibitem{pang2021recorrupted}
Tongyao Pang, Huan Zheng, Yuhui Quan, and Hui Ji.
\newblock Recorrupted-to-recorrupted: unsupervised deep learning for image
  denoising.
\newblock In {\em Proceedings of the IEEE/CVF conference on computer vision and
  pattern recognition}, pages 2043--2052, 2021.

\bibitem{cui2019pet}
Jianan Cui, Kuang Gong, Ning Guo, Chenxi Wu, Xiaxia Meng, Kyungsang Kim, Kun
  Zheng, Zhifang Wu, Liping Fu, Baixuan Xu, et~al.
\newblock Pet image denoising using unsupervised deep learning.
\newblock {\em European journal of nuclear medicine and molecular imaging},
  46:2780--2789, 2019.

\bibitem{arsene2019deep}
Corneliu~TC Arsene, Richard Hankins, and Hujun Yin.
\newblock Deep learning models for denoising ecg signals.
\newblock In {\em 2019 27th European Signal Processing Conference (EUSIPCO)},
  pages 1--5. IEEE, 2019.

\bibitem{nurmaini2020deep}
Siti Nurmaini, Annisa Darmawahyuni, Akhmad~Noviar Sakti~Mukti, Muhammad~Naufal
  Rachmatullah, Firdaus Firdaus, and Bambang Tutuko.
\newblock Deep learning-based stacked denoising and autoencoder for ecg
  heartbeat classification.
\newblock {\em Electronics}, 9(1):135, 2020.

\bibitem{peng2020novel}
Zhiyun Peng, Sui Peng, Lidan Fu, Binchun Lu, Junjie Tang, Ke~Wang, and Wenyuan
  Li.
\newblock A novel deep learning ensemble model with data denoising for
  short-term wind speed forecasting.
\newblock {\em Energy Conversion and Management}, 207:112524, 2020.

\bibitem{zhang2023birdsoundsdenoising}
Youshan Zhang and Jialu Li.
\newblock Birdsoundsdenoising: Deep visual audio denoising for bird sounds.
\newblock In {\em Proceedings of the IEEE/CVF Winter Conference on Applications
  of Computer Vision}, pages 2248--2257, 2023.

\bibitem{xu2020listening}
Ruilin Xu, Rundi Wu, Yuko Ishiwaka, Carl Vondrick, and Changxi Zheng.
\newblock Listening to sounds of silence for speech denoising.
\newblock {\em Advances in Neural Information Processing Systems},
  33:9633--9648, 2020.

\bibitem{tian2022sdndti}
Qiyuan Tian, Ziyu Li, Qiuyun Fan, Jonathan~R Polimeni, Berkin Bilgic, David~H
  Salat, and Susie~Y Huang.
\newblock Sdndti: Self-supervised deep learning-based denoising for diffusion
  tensor mri.
\newblock {\em Neuroimage}, 253:119033, 2022.

\bibitem{thrun2002probabilistic}
Sebastian Thrun.
\newblock Probabilistic robotics.
\newblock {\em Communications of the ACM}, 45(3):52--57, 2002.

\bibitem{geiger2012we}
Andreas Geiger, Philip Lenz, and Raquel Urtasun.
\newblock Are we ready for autonomous driving? the kitti vision benchmark
  suite.
\newblock In {\em 2012 IEEE conference on computer vision and pattern
  recognition}, pages 3354--3361. IEEE, 2012.

\bibitem{burri2016euroc}
Michael Burri, Janosch Nikolic, Pascal Gohl, Thomas Schneider, Joern Rehder,
  Sammy Omari, Markus~W Achtelik, and Roland Siegwart.
\newblock The euroc micro aerial vehicle datasets.
\newblock {\em The International Journal of Robotics Research},
  35(10):1157--1163, 2016.

\bibitem{carlevaris2016university}
Nicholas Carlevaris-Bianco, Arash~K Ushani, and Ryan~M Eustice.
\newblock University of michigan north campus long-term vision and lidar
  dataset.
\newblock {\em The International Journal of Robotics Research},
  35(9):1023--1035, 2016.

\bibitem{song2015sun}
Shuran Song, Samuel~P Lichtenberg, and Jianxiong Xiao.
\newblock Sun rgb-d: A rgb-d scene understanding benchmark suite.
\newblock In {\em Proceedings of the IEEE conference on computer vision and
  pattern recognition}, pages 567--576, 2015.

\bibitem{cordts2016cityscapes}
Marius Cordts, Mohamed Omran, Sebastian Ramos, Timo Rehfeld, Markus Enzweiler,
  Rodrigo Benenson, Uwe Franke, Stefan Roth, and Bernt Schiele.
\newblock The cityscapes dataset for semantic urban scene understanding.
\newblock In {\em Proceedings of the IEEE conference on computer vision and
  pattern recognition}, pages 3213--3223, 2016.

\bibitem{jensen2016vision}
Morten~Born{\o} Jensen, Mark~Philip Philipsen, Andreas M{\o}gelmose,
  Thomas~Baltzer Moeslund, and Mohan~Manubhai Trivedi.
\newblock Vision for looking at traffic lights: Issues, survey, and
  perspectives.
\newblock {\em IEEE Transactions on Intelligent Transportation Systems},
  17(7):1800--1815, 2016.

\bibitem{robicquet2016learning}
Alexandre Robicquet, Amir Sadeghian, Alexandre Alahi, and Silvio Savarese.
\newblock Learning social etiquette: Human trajectory prediction.
\newblock In {\em European Conference on Computer Vision (ECCV)}, volume~3,
  2016.

\bibitem{mueller2016benchmark}
Matthias Mueller, Neil Smith, and Bernard Ghanem.
\newblock A benchmark and simulator for uav tracking.
\newblock In {\em Computer Vision--ECCV 2016: 14th European Conference,
  Amsterdam, The Netherlands, October 11--14, 2016, Proceedings, Part I 14},
  pages 445--461. Springer, 2016.

\bibitem{majdik2017zurich}
Andr{\'a}s~L Majdik, Charles Till, and Davide Scaramuzza.
\newblock The zurich urban micro aerial vehicle dataset.
\newblock {\em The International Journal of Robotics Research}, 36(3):269--273,
  2017.

\bibitem{maddern20171}
Will Maddern, Geoffrey Pascoe, Chris Linegar, and Paul Newman.
\newblock 1 year, 1000 km: The oxford robotcar dataset.
\newblock {\em The International Journal of Robotics Research}, 36(1):3--15,
  2017.

\bibitem{perazzi2016benchmark}
Federico Perazzi, Jordi Pont-Tuset, Brian McWilliams, Luc Van~Gool, Markus
  Gross, and Alexander Sorkine-Hornung.
\newblock A benchmark dataset and evaluation methodology for video object
  segmentation.
\newblock In {\em Proceedings of the IEEE conference on computer vision and
  pattern recognition}, pages 724--732, 2016.

\bibitem{yu2020bdd100k}
Fisher Yu, Haofeng Chen, Xin Wang, Wenqi Xian, Yingying Chen, Fangchen Liu,
  Vashisht Madhavan, and Trevor Darrell.
\newblock Bdd100k: A diverse driving dataset for heterogeneous multitask
  learning.
\newblock In {\em Proceedings of the IEEE/CVF conference on computer vision and
  pattern recognition}, pages 2636--2645, 2020.

\bibitem{huang2018apolloscape}
Xinyu Huang, Xinjing Cheng, Qichuan Geng, Binbin Cao, Dingfu Zhou, Peng Wang,
  Yuanqing Lin, and Ruigang Yang.
\newblock The apolloscape dataset for autonomous driving.
\newblock In {\em Proceedings of the IEEE conference on computer vision and
  pattern recognition workshops}, pages 954--960, 2018.

\bibitem{sun2018robust}
Ke~Sun, Kartik Mohta, Bernd Pfrommer, Michael Watterson, Sikang Liu, Yash
  Mulgaonkar, Camillo~J Taylor, and Vijay Kumar.
\newblock Robust stereo visual inertial odometry for fast autonomous flight.
\newblock {\em IEEE Robotics and Automation Letters}, 3(2):965--972, 2018.

\bibitem{dai2017scannet}
Angela Dai, Angel~X Chang, Manolis Savva, Maciej Halber, Thomas Funkhouser, and
  Matthias Nie{\ss}ner.
\newblock Scannet: Richly-annotated 3d reconstructions of indoor scenes.
\newblock In {\em Proceedings of the IEEE conference on computer vision and
  pattern recognition}, pages 5828--5839, 2017.

\bibitem{choi2018kaist}
Yukyung Choi, Namil Kim, Soonmin Hwang, Kibaek Park, Jae~Shin Yoon, Kyounghwan
  An, and In~So Kweon.
\newblock Kaist multi-spectral day/night data set for autonomous and assisted
  driving.
\newblock {\em IEEE Transactions on Intelligent Transportation Systems},
  19(3):934--948, 2018.

\bibitem{Delmerico19icra}
Jeffrey Delmerico, Titus Cieslewski, Henri Rebecq, Matthias Faessler, and
  Davide Scaramuzza.
\newblock Are we ready for autonomous drone racing? the {UZH-FPV} drone racing
  dataset.
\newblock In {\em {IEEE} Int. Conf. Robot. Autom. ({ICRA})}, 2019.

\bibitem{chang2019argoverse}
Ming-Fang Chang, John Lambert, Patsorn Sangkloy, Jagjeet Singh, Slawomir Bak,
  Andrew Hartnett, De~Wang, Peter Carr, Simon Lucey, Deva Ramanan, et~al.
\newblock Argoverse: 3d tracking and forecasting with rich maps.
\newblock In {\em Proceedings of the IEEE/CVF conference on computer vision and
  pattern recognition}, pages 8748--8757, 2019.

\bibitem{patil2019h3d}
Abhishek Patil, Srikanth Malla, Haiming Gang, and Yi-Ting Chen.
\newblock The h3d dataset for full-surround 3d multi-object detection and
  tracking in crowded urban scenes.
\newblock In {\em 2019 International Conference on Robotics and Automation
  (ICRA)}, pages 9552--9557. IEEE, 2019.

\bibitem{kesten2019lyft}
R~Kesten, M~Usman, J~Houston, T~Pandya, K~Nadhamuni, A~Ferreira, M~Yuan, B~Low,
  A~Jain, P~Ondruska, et~al.
\newblock Lyft level 5 av dataset 2019.
\newblock {\em urlhttps://level5. lyft. com/dataset}, 1:3, 2019.

\bibitem{geyer2020a2d2}
Jakob Geyer, Yohannes Kassahun, Mentar Mahmudi, Xavier Ricou, Rupesh Durgesh,
  Andrew~S Chung, Lorenz Hauswald, Viet~Hoang Pham, Maximilian M{\"u}hlegg,
  Sebastian Dorn, et~al.
\newblock A2d2: Audi autonomous driving dataset.
\newblock {\em arXiv preprint arXiv:2004.06320}, 2020.

\bibitem{pham20203d}
Quang-Hieu Pham, Pierre Sevestre, Ramanpreet~Singh Pahwa, Huijing Zhan, Chun~Ho
  Pang, Yuda Chen, Armin Mustafa, Vijay Chandrasekhar, and Jie Lin.
\newblock A 3d dataset: Towards autonomous driving in challenging environments.
\newblock In {\em 2020 IEEE International Conference on Robotics and Automation
  (ICRA)}, pages 2267--2273. IEEE, 2020.

\bibitem{deschaud2021kitti}
Jean-Emmanuel Deschaud.
\newblock Kitti-carla: a kitti-like dataset generated by carla simulator.
\newblock {\em arXiv preprint arXiv:2109.00892}, 2021.

\bibitem{caesar2020nuscenes}
Holger Caesar, Varun Bankiti, Alex~H Lang, Sourabh Vora, Venice~Erin Liong,
  Qiang Xu, Anush Krishnan, Yu~Pan, Giancarlo Baldan, and Oscar Beijbom.
\newblock nuscenes: A multimodal dataset for autonomous driving.
\newblock In {\em Proceedings of the IEEE/CVF conference on computer vision and
  pattern recognition}, pages 11621--11631, 2020.

\bibitem{sun2020scalability}
Pei Sun, Henrik Kretzschmar, Xerxes Dotiwalla, Aurelien Chouard, Vijaysai
  Patnaik, Paul Tsui, James Guo, Yin Zhou, Yuning Chai, Benjamin Caine, et~al.
\newblock Scalability in perception for autonomous driving: Waymo open dataset.
\newblock In {\em Proceedings of the IEEE/CVF conference on computer vision and
  pattern recognition}, pages 2446--2454, 2020.

\bibitem{antonini2020blackbird}
Amado Antonini, Winter Guerra, Varun Murali, Thomas Sayre-McCord, and Sertac
  Karaman.
\newblock The blackbird uav dataset.
\newblock {\em The International Journal of Robotics Research},
  39(10-11):1346--1364, 2020.

\bibitem{zhu2021detection}
Pengfei Zhu, Longyin Wen, Dawei Du, Xiao Bian, Heng Fan, Qinghua Hu, and Haibin
  Ling.
\newblock Detection and tracking meet drones challenge.
\newblock {\em IEEE Transactions on Pattern Analysis and Machine Intelligence},
  44(11):7380--7399, 2021.

\bibitem{chen2022whuvid}
Tianyang Chen, Fangling Pu, Hongjia Chen, and Zhihong Liu.
\newblock Whuvid: A large-scale stereo-imu dataset for visual-inertial odometry
  and autonomous driving in chinese urban scenarios.
\newblock {\em Remote Sensing}, 14(9):2033, 2022.

\bibitem{chang2022hdin}
Yingxiu Chang, Yongqiang Cheng, John Murray, Shi Huang, and Guangyi Shi.
\newblock The hdin dataset: A real-world indoor uav dataset with multi-task
  labels for visual-based navigation.
\newblock {\em Drones}, 6(8):202, 2022.

\bibitem{zou2023object}
Zhengxia Zou, Keyan Chen, Zhenwei Shi, Yuhong Guo, and Jieping Ye.
\newblock Object detection in 20 years: A survey.
\newblock {\em Proceedings of the IEEE}, 2023.

\bibitem{viola2004robust}
Paul Viola and Michael~J Jones.
\newblock Robust real-time face detection.
\newblock {\em International journal of computer vision}, 57:137--154, 2004.

\bibitem{dalal2005histograms}
Navneet Dalal and Bill Triggs.
\newblock Histograms of oriented gradients for human detection.
\newblock In {\em 2005 IEEE computer society conference on computer vision and
  pattern recognition (CVPR'05)}, volume~1, pages 886--893. Ieee, 2005.

\bibitem{felzenszwalb2008discriminatively}
Pedro Felzenszwalb, David McAllester, and Deva Ramanan.
\newblock A discriminatively trained, multiscale, deformable part model.
\newblock In {\em 2008 IEEE conference on computer vision and pattern
  recognition}, pages 1--8. Ieee, 2008.

\bibitem{girshick2014rich}
Ross Girshick, Jeff Donahue, Trevor Darrell, and Jitendra Malik.
\newblock Rich feature hierarchies for accurate object detection and semantic
  segmentation.
\newblock In {\em Proceedings of the IEEE conference on computer vision and
  pattern recognition}, pages 580--587, 2014.

\bibitem{he2015spatial}
Kaiming He, Xiangyu Zhang, Shaoqing Ren, and Jian Sun.
\newblock Spatial pyramid pooling in deep convolutional networks for visual
  recognition.
\newblock {\em IEEE transactions on pattern analysis and machine intelligence},
  37(9):1904--1916, 2015.

\bibitem{girshick2015fast}
Ross Girshick.
\newblock Fast r-cnn.
\newblock In {\em Proceedings of the IEEE international conference on computer
  vision}, pages 1440--1448, 2015.

\bibitem{ren2015faster}
Shaoqing Ren, Kaiming He, Ross Girshick, and Jian Sun.
\newblock Faster r-cnn: Towards real-time object detection with region proposal
  networks.
\newblock {\em Advances in neural information processing systems}, 28, 2015.

\bibitem{redmon2016you}
Joseph Redmon, Santosh Divvala, Ross Girshick, and Ali Farhadi.
\newblock You only look once: Unified, real-time object detection.
\newblock In {\em Proceedings of the IEEE conference on computer vision and
  pattern recognition}, pages 779--788, 2016.

\bibitem{redmon2017yolo9000}
Joseph Redmon and Ali Farhadi.
\newblock Yolo9000: better, faster, stronger.
\newblock In {\em Proceedings of the IEEE conference on computer vision and
  pattern recognition}, pages 7263--7271, 2017.

\bibitem{redmon2018yolov3}
Joseph Redmon and Ali Farhadi.
\newblock Yolov3: An incremental improvement.
\newblock {\em arXiv preprint arXiv:1804.02767}, 2018.

\bibitem{bochkovskiy2020yolov4}
Alexey Bochkovskiy, Chien-Yao Wang, and Hong-Yuan~Mark Liao.
\newblock Yolov4: Optimal speed and accuracy of object detection.
\newblock {\em arXiv preprint arXiv:2004.10934}, 2020.

\bibitem{liu2016ssd}
Wei Liu, Dragomir Anguelov, Dumitru Erhan, Christian Szegedy, Scott Reed,
  Cheng-Yang Fu, and Alexander~C Berg.
\newblock Ssd: Single shot multibox detector.
\newblock In {\em Computer Vision--ECCV 2016: 14th European Conference,
  Amsterdam, The Netherlands, October 11--14, 2016, Proceedings, Part I 14},
  pages 21--37. Springer, 2016.

\bibitem{lin2017feature}
Tsung-Yi Lin, Piotr Doll{\'a}r, Ross Girshick, Kaiming He, Bharath Hariharan,
  and Serge Belongie.
\newblock Feature pyramid networks for object detection.
\newblock In {\em Proceedings of the IEEE conference on computer vision and
  pattern recognition}, pages 2117--2125, 2017.

\bibitem{lin2017focal}
Tsung-Yi Lin, Priya Goyal, Ross Girshick, Kaiming He, and Piotr Doll{\'a}r.
\newblock Focal loss for dense object detection.
\newblock In {\em Proceedings of the IEEE international conference on computer
  vision}, pages 2980--2988, 2017.

\bibitem{law2018cornernet}
Hei Law and Jia Deng.
\newblock Cornernet: Detecting objects as paired keypoints.
\newblock In {\em Proceedings of the European conference on computer vision
  (ECCV)}, pages 734--750, 2018.

\bibitem{zhou2019objects}
Xingyi Zhou, Dequan Wang, and Philipp Kr{\"a}henb{\"u}hl.
\newblock Objects as points.
\newblock {\em arXiv preprint arXiv:1904.07850}, 2019.

\bibitem{carion2020end}
Nicolas Carion, Francisco Massa, Gabriel Synnaeve, Nicolas Usunier, Alexander
  Kirillov, and Sergey Zagoruyko.
\newblock End-to-end object detection with transformers.
\newblock In {\em Computer Vision--ECCV 2020: 16th European Conference,
  Glasgow, UK, August 23--28, 2020, Proceedings, Part I 16}, pages 213--229.
  Springer, 2020.

\bibitem{cai2018cascade}
Zhaowei Cai and Nuno Vasconcelos.
\newblock Cascade r-cnn: Delving into high quality object detection.
\newblock In {\em Proceedings of the IEEE conference on computer vision and
  pattern recognition}, pages 6154--6162, 2018.

\bibitem{tan2020efficientdet}
Mingxing Tan, Ruoming Pang, and Quoc~V Le.
\newblock Efficientdet: Scalable and efficient object detection.
\newblock In {\em Proceedings of the IEEE/CVF conference on computer vision and
  pattern recognition}, pages 10781--10790, 2020.

\bibitem{sun2021sparse}
Peize Sun, Rufeng Zhang, Yi~Jiang, Tao Kong, Chenfeng Xu, Wei Zhan, Masayoshi
  Tomizuka, Lei Li, Zehuan Yuan, Changhu Wang, et~al.
\newblock Sparse r-cnn: End-to-end object detection with learnable proposals.
\newblock In {\em Proceedings of the IEEE/CVF conference on computer vision and
  pattern recognition}, pages 14454--14463, 2021.

\bibitem{yang2019reppoints}
Ze~Yang, Shaohui Liu, Han Hu, Liwei Wang, and Stephen Lin.
\newblock Reppoints: Point set representation for object detection.
\newblock In {\em Proceedings of the IEEE/CVF international conference on
  computer vision}, pages 9657--9666, 2019.

\bibitem{tian2019fcos}
Zhi Tian, Chunhua Shen, Hao Chen, and Tong He.
\newblock Fcos: Fully convolutional one-stage object detection.
\newblock In {\em Proceedings of the IEEE/CVF international conference on
  computer vision}, pages 9627--9636, 2019.

\bibitem{lin2014microsoft}
Tsung-Yi Lin, Michael Maire, Serge Belongie, James Hays, Pietro Perona, Deva
  Ramanan, Piotr Doll{\'a}r, and C~Lawrence Zitnick.
\newblock Microsoft coco: Common objects in context.
\newblock In {\em Computer Vision--ECCV 2014: 13th European Conference, Zurich,
  Switzerland, September 6-12, 2014, Proceedings, Part V 13}, pages 740--755.
  Springer, 2014.

\bibitem{li2017lecture}
Fei-Fei Li, Justin Johnson, and Serena Yeung.
\newblock Lecture 11: Detection and segmentation, 2017.

\bibitem{long2015fully}
Jonathan Long, Evan Shelhamer, and Trevor Darrell.
\newblock Fully convolutional networks for semantic segmentation.
\newblock In {\em Proceedings of the IEEE conference on computer vision and
  pattern recognition}, pages 3431--3440, 2015.

\bibitem{ronneberger2015u}
Olaf Ronneberger, Philipp Fischer, and Thomas Brox.
\newblock U-net: Convolutional networks for biomedical image segmentation.
\newblock In {\em Medical Image Computing and Computer-Assisted
  Intervention--MICCAI 2015: 18th International Conference, Munich, Germany,
  October 5-9, 2015, Proceedings, Part III 18}, pages 234--241. Springer, 2015.

\bibitem{badrinarayanan2017segnet}
Vijay Badrinarayanan, Alex Kendall, and Roberto Cipolla.
\newblock Segnet: A deep convolutional encoder-decoder architecture for image
  segmentation.
\newblock {\em IEEE transactions on pattern analysis and machine intelligence},
  39(12):2481--2495, 2017.

\bibitem{chen2017deeplab}
Liang-Chieh Chen, George Papandreou, Iasonas Kokkinos, Kevin Murphy, and Alan~L
  Yuille.
\newblock Deeplab: Semantic image segmentation with deep convolutional nets,
  atrous convolution, and fully connected crfs.
\newblock {\em IEEE transactions on pattern analysis and machine intelligence},
  40(4):834--848, 2017.

\bibitem{zhao2017pyramid}
Hengshuang Zhao, Jianping Shi, Xiaojuan Qi, Xiaogang Wang, and Jiaya Jia.
\newblock Pyramid scene parsing network.
\newblock In {\em Proceedings of the IEEE conference on computer vision and
  pattern recognition}, pages 2881--2890, 2017.

\bibitem{zhao2018icnet}
Hengshuang Zhao, Xiaojuan Qi, Xiaoyong Shen, Jianping Shi, and Jiaya Jia.
\newblock Icnet for real-time semantic segmentation on high-resolution images.
\newblock In {\em Proceedings of the European conference on computer vision
  (ECCV)}, pages 405--420, 2018.

\bibitem{paszke2016enet}
Adam Paszke, Abhishek Chaurasia, Sangpil Kim, and Eugenio Culurciello.
\newblock Enet: A deep neural network architecture for real-time semantic
  segmentation.
\newblock {\em arXiv preprint arXiv:1606.02147}, 2016.

\bibitem{chen2017rethinking}
Liang-Chieh Chen, George Papandreou, Florian Schroff, and Hartwig Adam.
\newblock Rethinking atrous convolution for semantic image segmentation.
\newblock {\em arXiv preprint arXiv:1706.05587}, 2017.

\bibitem{wang2020deep}
Jingdong Wang, Ke~Sun, Tianheng Cheng, Borui Jiang, Chaorui Deng, Yang Zhao,
  Dong Liu, Yadong Mu, Mingkui Tan, Xinggang Wang, et~al.
\newblock Deep high-resolution representation learning for visual recognition.
\newblock {\em IEEE transactions on pattern analysis and machine intelligence},
  43(10):3349--3364, 2020.

\bibitem{he2017mask}
Kaiming He, Georgia Gkioxari, Piotr Doll{\'a}r, and Ross Girshick.
\newblock Mask r-cnn.
\newblock In {\em Proceedings of the IEEE international conference on computer
  vision}, pages 2961--2969, 2017.

\bibitem{chaurasia2017linknet}
Abhishek Chaurasia and Eugenio Culurciello.
\newblock Linknet: Exploiting encoder representations for efficient semantic
  segmentation.
\newblock In {\em 2017 IEEE visual communications and image processing (VCIP)},
  pages 1--4. IEEE, 2017.

\bibitem{lin2017refinenet}
Guosheng Lin, Anton Milan, Chunhua Shen, and Ian Reid.
\newblock Refinenet: Multi-path refinement networks for high-resolution
  semantic segmentation.
\newblock In {\em Proceedings of the IEEE conference on computer vision and
  pattern recognition}, pages 1925--1934, 2017.

\bibitem{yu2018bisenet}
Changqian Yu, Jingbo Wang, Chao Peng, Changxin Gao, Gang Yu, and Nong Sang.
\newblock Bisenet: Bilateral segmentation network for real-time semantic
  segmentation.
\newblock In {\em Proceedings of the European conference on computer vision
  (ECCV)}, pages 325--341, 2018.

\bibitem{mahendrakar2022performance}
Trupti Mahendrakar, Andrew Ekblad, Nathan Fischer, Ryan White, Markus Wilde,
  Brian Kish, and Isaac Silver.
\newblock Performance study of yolov5 and faster r-cnn for autonomous
  navigation around non-cooperative targets.
\newblock In {\em 2022 IEEE Aerospace Conference (AERO)}, pages 1--12. IEEE,
  2022.

\bibitem{sadeghi2022deep}
Shabnam Sadeghi~Esfahlani, Alireza Sanaei, Mohammad Ghorabian, and Hassan
  Shirvani.
\newblock The deep convolutional neural network role in the autonomous
  navigation of mobile robots (srobo).
\newblock {\em Remote Sensing}, 14(14):3324, 2022.

\bibitem{shin2022environment}
Donghun Shin, Joongho Cho, and Jaeho Kim.
\newblock Environment-adaptive object detection framework for autonomous mobile
  robots.
\newblock {\em Sensors}, 22(19):7647, 2022.

\bibitem{jocher2020ultralytics}
Glenn Jocher, Alex Stoken, Jirka Borovec, Ayush Chaurasia, and L~Changyu.
\newblock ultralytics/yolov5.
\newblock {\em Github Repository, YOLOv5}, 2020.

\bibitem{shao2019objects365}
Shuai Shao, Zeming Li, Tianyuan Zhang, Chao Peng, Gang Yu, Xiangyu Zhang, Jing
  Li, and Jian Sun.
\newblock Objects365: A large-scale, high-quality dataset for object detection.
\newblock In {\em Proceedings of the IEEE/CVF international conference on
  computer vision}, pages 8430--8439, 2019.

\bibitem{he2016deep}
Kaiming He, Xiangyu Zhang, Shaoqing Ren, and Jian Sun.
\newblock Deep residual learning for image recognition.
\newblock In {\em Proceedings of the IEEE conference on computer vision and
  pattern recognition}, pages 770--778, 2016.

\bibitem{zhou2017places}
Bolei Zhou, Agata Lapedriza, Aditya Khosla, Aude Oliva, and Antonio Torralba.
\newblock Places: A 10 million image database for scene recognition.
\newblock {\em IEEE transactions on pattern analysis and machine intelligence},
  40(6):1452--1464, 2017.

\bibitem{zhou2022sgm3d}
Zheyuan Zhou, Liang Du, Xiaoqing Ye, Zhikang Zou, Xiao Tan, Li~Zhang, Xiangyang
  Xue, and Jianfeng Feng.
\newblock Sgm3d: stereo guided monocular 3d object detection.
\newblock {\em IEEE Robotics and Automation Letters}, 7(4):10478--10485, 2022.

\bibitem{houston2021one}
John Houston, Guido Zuidhof, Luca Bergamini, Yawei Ye, Long Chen, Ashesh Jain,
  Sammy Omari, Vladimir Iglovikov, and Peter Ondruska.
\newblock One thousand and one hours: Self-driving motion prediction dataset.
\newblock In {\em Conference on Robot Learning}, pages 409--418. PMLR, 2021.

\bibitem{afif2022evaluation}
Mouna Afif, Riadh Ayachi, Yahia Said, and Mohamed Atri.
\newblock An evaluation of efficientdet for object detection used for indoor
  robots assistance navigation.
\newblock {\em Journal of Real-Time Image Processing}, 19(3):651--661, 2022.

\bibitem{talele2019detection}
Ajay Talele, Aseem Patil, and Bhushan Barse.
\newblock Detection of real time objects using tensorflow and opencv.
\newblock {\em Asian Journal For Convergence In Technology (AJCT)
  ISSN-2350-1146}, 2019.

\bibitem{abadi2016tensorflow}
Mart{\'\i}n Abadi, Paul Barham, Jianmin Chen, Zhifeng Chen, Andy Davis, Jeffrey
  Dean, Matthieu Devin, Sanjay Ghemawat, Geoffrey Irving, Michael Isard, et~al.
\newblock Tensorflow: a system for large-scale machine learning.
\newblock In {\em Osdi}, volume~16, pages 265--283. Savannah, GA, USA, 2016.

\bibitem{bradski2000opencv}
Gary Bradski.
\newblock The opencv library.
\newblock {\em Dr. Dobb's Journal: Software Tools for the Professional
  Programmer}, 25(11):120--123, 2000.

\bibitem{jiao2021road}
Shuangjian Jiao and Lingling Wang.
\newblock Road obstacle detection in bad weather based on deep learning.
\newblock In {\em Journal of Physics: Conference Series}, volume 1881, page
  042041. IOP Publishing, 2021.

\bibitem{haris2020obstacle}
Malik Haris and Jin Hou.
\newblock Obstacle detection and safely navigate the autonomous vehicle from
  unexpected obstacles on the driving lane.
\newblock {\em Sensors}, 20(17):4719, 2020.

\bibitem{luu2020traditional}
Van-Tin Luu, Viet-Cuong Huynh, Vu-Hoang Tran, Trung-Hieu Nguyen, et~al.
\newblock Traditional method meets deep learning in an adaptive lane and
  obstacle detection system.
\newblock In {\em 2020 5th International Conference on Green Technology and
  Sustainable Development (GTSD)}, pages 148--152. IEEE, 2020.

\bibitem{qiu2020vision}
Zhengjun Qiu, Nan Zhao, Lei Zhou, Mengcen Wang, Liangliang Yang, Hui Fang, Yong
  He, and Yufei Liu.
\newblock Vision-based moving obstacle detection and tracking in paddy field
  using improved yolov3 and deep sort.
\newblock {\em Sensors}, 20(15):4082, 2020.

\bibitem{cervera2022u19}
Albert~Aar{\'o}n Cervera-Uribe and Paul~Erick Mendez-Monroy.
\newblock U19-net: a deep learning approach for obstacle detection in
  self-driving cars.
\newblock {\em Soft Computing}, 26(11):5195--5207, 2022.

\bibitem{ci2022novel}
Wenyan Ci, Tianxiang Xu, Runze Lin, and Shan Lu.
\newblock A novel method for unexpected obstacle detection in the traffic
  environment based on computer vision.
\newblock {\em Applied Sciences}, 12(18):8937, 2022.

\bibitem{pinggera2016lost}
Peter Pinggera, Sebastian Ramos, Stefan Gehrig, Uwe Franke, Carsten Rother, and
  Rudolf Mester.
\newblock Lost and found: detecting small road hazards for self-driving
  vehicles.
\newblock In {\em 2016 IEEE/RSJ International Conference on Intelligent Robots
  and Systems (IROS)}, pages 1099--1106. IEEE, 2016.

\bibitem{wang2022farmland}
Dashuai Wang, Zhuolin Li, Xiaoqiang Du, Zenghong Ma, and Xiaoguang Liu.
\newblock Farmland obstacle detection from the perspective of uavs based on
  non-local deformable detr.
\newblock {\em Agriculture}, 12(12):1983, 2022.

\bibitem{li2022stereovoxelnet}
Hongyu Li, Zhengang Li, Neset~Unver Akmandor, Huaizu Jiang, Yanzhi Wang, and
  Taskin Padir.
\newblock Stereovoxelnet: Real-time obstacle detection based on occupancy
  voxels from a stereo camera using deep neural networks.
\newblock {\em arXiv preprint arXiv:2209.08459}, 2022.

\bibitem{cortes2022dali}
Irene Cort{\'e}s, Jorge Beltr{\'a}n, Arturo De~La~Escalera, and Fernando
  Garc{\'\i}a.
\newblock Dali: Domain adaptation in lidar point clouds for 3d obstacle
  detection.
\newblock In {\em 2022 IEEE 25th International Conference on Intelligent
  Transportation Systems (ITSC)}, pages 3837--3842. IEEE, 2022.

\bibitem{wang2022obstacle}
Wenshan Wang, Shuang Wang, Yongcun Guo, and Yanqiu Zhao.
\newblock Obstacle detection method of unmanned electric locomotive in coal
  mine based on yolov3-4l.
\newblock {\em Journal of Electronic Imaging}, 31(2):023032--023032, 2022.

\bibitem{franke2022towards}
Marten Franke, Chaitra Reddy, Danijela Risti{\'c}-Durrant, Jehan Jayawardana,
  Kai Michels, Milan Bani{\'c}, and Milo{\v{s}} Simonovi{\'c}.
\newblock Towards holistic autonomous obstacle detection in railways by
  complementing of on-board vision with uav-based object localization.
\newblock In {\em 2022 IEEE/RSJ International Conference on Intelligent Robots
  and Systems (IROS)}, pages 7012--7019. IEEE, 2022.

\bibitem{devi2022vision}
Y~Sarada Devi, S~Sathvik, P~Ananya, P~Tharuni, and N~Naga~Krishna Vamsi.
\newblock Vision-based obstacle detection and collision prevention in
  self-driving cars.
\newblock In {\em Journal of Physics: Conference Series}, volume 2335, page
  012019. IOP Publishing, 2022.

\bibitem{hayashida2022obstacle}
Naoya Hayashida and Hiroyoshi Miwa.
\newblock Obstacle detection support system using monocular camera.
\newblock In {\em Advances in Intelligent Networking and Collaborative Systems:
  The 14th International Conference on Intelligent Networking and Collaborative
  Systems (INCoS-2022)}, pages 62--73. Springer, 2022.

\bibitem{byun2022design}
Sung-Woo Byun, Donghee Noh, and Hea-Min Lee.
\newblock Design of obstacle detection method for autonomous driving in
  agricultural environments.
\newblock In {\em 2022 Thirteenth International Conference on Ubiquitous and
  Future Networks (ICUFN)}, pages 494--496. IEEE, 2022.

\bibitem{huu2022proposing}
Phat~Nguyen Huu, Quyen Pham~Thi, and Phuong Tong Thi~Quynh.
\newblock Proposing lane and obstacle detection algorithm using yolo to control
  self-driving cars on advanced networks.
\newblock {\em Advances in Multimedia}, 2022, 2022.

\bibitem{chen2022real}
Guotong Chen, Jiangtao Qi, and Zeyu Dai.
\newblock Real-time maritime obstacle detection based on yolov5 for autonomous
  berthing.
\newblock In {\em Bio-Inspired Computing: Theories and Applications: 16th
  International Conference, BIC-TA 2021, Taiyuan, China, December 17--19, 2021,
  Revised Selected Papers, Part II}, pages 412--427. Springer, 2022.

\bibitem{ma2022fast}
Wenhua Ma, Xin Wang, and Baohua Wang.
\newblock Fast obstacle detection platform based on yolo.
\newblock In {\em Proceedings of the 5th International Conference on Electrical
  Engineering and Information Technologies for Rail Transportation (EITRT)
  2021: Rail Transportation System Safety and Maintenance Technologies}, pages
  218--226. Springer, 2022.

\bibitem{gasperini2021certainnet}
Stefano Gasperini, Jan Haug, Mohammad-Ali~Nikouei Mahani, Alvaro Marcos-Ramiro,
  Nassir Navab, Benjamin Busam, and Federico Tombari.
\newblock Certainnet: Sampling-free uncertainty estimation for object
  detection.
\newblock {\em IEEE Robotics and Automation Letters}, 7(2):698--705, 2021.

\bibitem{hu2022investigating}
Hanjiang Hu, Zuxin Liu, Sharad Chitlangia, Akhil Agnihotri, and Ding Zhao.
\newblock Investigating the impact of multi-lidar placement on object detection
  for autonomous driving.
\newblock In {\em Proceedings of the IEEE/CVF Conference on Computer Vision and
  Pattern Recognition}, pages 2550--2559, 2022.

\bibitem{chang2020spatial}
Shuo Chang, Yifan Zhang, Fan Zhang, Xiaotong Zhao, Sai Huang, Zhiyong Feng, and
  Zhiqing Wei.
\newblock Spatial attention fusion for obstacle detection using mmwave radar
  and vision sensor.
\newblock {\em Sensors}, 20(4):956, 2020.

\bibitem{pranav2022deeprecog}
MV~Pranav, AV~Shreyas~Madhav, and Janaki Meena.
\newblock Deeprecog: Threefold underwater image deblurring and object
  recognition framework for auv vision systems.
\newblock {\em Multimedia Systems}, pages 1--11, 2022.

\bibitem{zhang2023automatic}
Qiang Zhang, Fei Yan, Weina Song, Rui Wang, and Gen Li.
\newblock Automatic obstacle detection method for the train based on deep
  learning.
\newblock {\em Sustainability}, 15(2):1184, 2023.

\bibitem{ullah2019localization}
Inam Ullah, Yu~Shen, Xin Su, Christian Esposito, and Chang Choi.
\newblock A localization based on unscented kalman filter and particle filter
  localization algorithms.
\newblock {\em IEEE Access}, 8:2233--2246, 2019.

\bibitem{raja2021pfin}
Gunasekaran Raja, Sailakshmi Suresh, Sudha Anbalagan, Aishwarya
  Ganapathisubramaniyan, and Neeraj Kumar.
\newblock Pfin: An efficient particle filter-based indoor navigation framework
  for uavs.
\newblock {\em IEEE Transactions on Vehicular Technology}, 70(5):4984--4992,
  2021.

\bibitem{grisetti2010tutorial}
Giorgio Grisetti, Rainer K{\"u}mmerle, Cyrill Stachniss, and Wolfram Burgard.
\newblock A tutorial on graph-based slam.
\newblock {\em IEEE Intelligent Transportation Systems Magazine}, 2(4):31--43,
  2010.

\bibitem{qin2020avp}
Tong Qin, Tongqing Chen, Yilun Chen, and Qing Su.
\newblock Avp-slam: Semantic visual mapping and localization for autonomous
  vehicles in the parking lot.
\newblock In {\em 2020 IEEE/RSJ International Conference on Intelligent Robots
  and Systems (IROS)}, pages 5939--5945. IEEE, 2020.

\bibitem{bavle2020vps}
Hriday Bavle, Paloma De~La~Puente, Jonathan~P How, and Pascual Campoy.
\newblock Vps-slam: Visual planar semantic slam for aerial robotic systems.
\newblock {\em IEEE Access}, 8:60704--60718, 2020.

\bibitem{celik2013monocular}
Koray Celik and Arun~K Somani.
\newblock Monocular vision slam for indoor aerial vehicles.
\newblock {\em Journal of electrical and computer engineering}, 2013:4--4,
  2013.

\bibitem{geromichalos2020slam}
Dimitrios Geromichalos, Martin Azkarate, Emmanouil Tsardoulias, Levin Gerdes,
  Loukas Petrou, and Carlos Perez Del~Pulgar.
\newblock Slam for autonomous planetary rovers with global localization.
\newblock {\em Journal of Field Robotics}, 37(5):830--847, 2020.

\bibitem{baldini2018autonomous}
Francesca Baldini, Alexei Harvard, Soon-Jo Chung, Issa Nesnas, and Shyamkumar
  Bhaskaran.
\newblock Autonomous small body mapping and spacecraft navigation via real-time
  spc-slam.
\newblock 2018.

\bibitem{tian2022kimera}
Yulun Tian, Yun Chang, Fernando~Herrera Arias, Carlos Nieto-Granda, Jonathan~P
  How, and Luca Carlone.
\newblock Kimera-multi: Robust, distributed, dense metric-semantic slam for
  multi-robot systems.
\newblock {\em IEEE Transactions on Robotics}, 38(4), 2022.

\bibitem{zhang2022deep}
Haoyang Zhang.
\newblock Deep learning applications in simultaneous localization and mapping.
\newblock In {\em Journal of Physics: Conference Series}, volume 2181, page
  012012. IOP Publishing, 2022.

\bibitem{beghdadi2022comprehensive}
Ayman Beghdadi and Malik Mallem.
\newblock A comprehensive overview of dynamic visual slam and deep learning:
  concepts, methods and challenges.
\newblock {\em Machine Vision and Applications}, 33(4):54, 2022.

\bibitem{amiri2019semi}
Ali~Jahani Amiri, Shing~Yan Loo, and Hong Zhang.
\newblock Semi-supervised monocular depth estimation with left-right
  consistency using deep neural network.
\newblock In {\em 2019 IEEE International Conference on Robotics and
  Biomimetics (ROBIO)}, pages 602--607. IEEE, 2019.

\bibitem{li2019depth}
Fu~Li, Quanlu Li, Tianjiao Zhang, Yi~Niu, and Guangming Shi.
\newblock Depth acquisition with the combination of structured light and deep
  learning stereo matching.
\newblock {\em Signal Processing: Image Communication}, 75:111--117, 2019.

\bibitem{xiao2019dynamic}
Linhui Xiao, Jinge Wang, Xiaosong Qiu, Zheng Rong, and Xudong Zou.
\newblock Dynamic-slam: Semantic monocular visual localization and mapping
  based on deep learning in dynamic environment.
\newblock {\em Robotics and Autonomous Systems}, 117:1--16, 2019.

\bibitem{wang2021lidar}
Weiqi Wang, Xiong You, Xin Zhang, Lingyu Chen, Lantian Zhang, and Xu~Liu.
\newblock Lidar-based slam under semantic constraints in dynamic environments.
\newblock {\em Remote Sensing}, 13(18):3651, 2021.

\bibitem{yu2020machining}
Chun-Yen Yu and Chao-Chung Peng.
\newblock Machining learning for 2d-slam object classification and recognition.
\newblock In {\em 2020 IEEE International Conference on Consumer
  Electronics-Taiwan (ICCE-Taiwan)}, pages 1--2. IEEE, 2020.

\bibitem{langer2020domain}
Ferdinand Langer, Andres Milioto, Alexandre Haag, Jens Behley, and Cyrill
  Stachniss.
\newblock Domain transfer for semantic segmentation of lidar data using deep
  neural networks.
\newblock In {\em 2020 IEEE/RSJ International Conference on Intelligent Robots
  and Systems (IROS)}, pages 8263--8270. IEEE, 2020.

\bibitem{song2022dynavins}
Seungwon Song, Hyungtae Lim, Alex~Junho Lee, and Hyun Myung.
\newblock Dynavins: A visual-inertial slam for dynamic environments.
\newblock {\em IEEE Robotics and Automation Letters}, 7(4):11523--11530, 2022.

\bibitem{sun2022multi}
Ying Sun, Jun Hu, Juntong Yun, Ying Liu, Dongxu Bai, Xin Liu, Guojun Zhao,
  Guozhang Jiang, Jianyi Kong, and Baojia Chen.
\newblock Multi-objective location and mapping based on deep learning and
  visual slam.
\newblock {\em Sensors}, 22(19):7576, 2022.

\bibitem{mur2017orb}
Raul Mur-Artal and Juan~D Tard{\'o}s.
\newblock Orb-slam2: An open-source slam system for monocular, stereo, and
  rgb-d cameras.
\newblock {\em IEEE transactions on robotics}, 33(5):1255--1262, 2017.

\bibitem{sturm2012benchmark}
J{\"u}rgen Sturm, Nikolas Engelhard, Felix Endres, Wolfram Burgard, and Daniel
  Cremers.
\newblock A benchmark for the evaluation of rgb-d slam systems.
\newblock In {\em 2012 IEEE/RSJ international conference on intelligent robots
  and systems}, pages 573--580. IEEE, 2012.

\bibitem{hong2021visual}
Sungchul Hong, Antyanta Bangunharcana, Jae-Min Park, Minseong Choi, and
  Hyu-Soung Shin.
\newblock Visual slam-based robotic mapping method for planetary construction.
\newblock {\em Sensors}, 21(22):7715, 2021.

\bibitem{lang2023svr}
Rongling Lang, Ya~Fan, and Qing Chang.
\newblock Svr-net: A sparse voxelized recurrent network for robust monocular
  slam with direct tsdf mapping.
\newblock {\em Sensors}, 23(8):3942, 2023.

\bibitem{lian2023point}
Yuanfeng Lian, Hao Sun, and Shaohua Dong.
\newblock Point--line-aware heterogeneous graph attention network for visual
  slam system.
\newblock {\em Applied Sciences}, 13(6):3816, 2023.

\bibitem{jeong2023cnn}
Hyein Jeong and Heoncheol Lee.
\newblock Cnn-based fault detection of scan matching for accurate slam in
  dynamic environments.
\newblock {\em Sensors}, 23(6):2940, 2023.

\bibitem{asgharpoor2022design}
Arman Asgharpoor~Golroudbari.
\newblock {\em Design and Simulation of Heading Estimation Algorithm}.
\newblock PhD thesis, University of Tehran, 2022.

\bibitem{asgharpoor4387238generalizable}
Arman Asgharpoor~Golroudbari and Mohammad~Hossein Sabour.
\newblock Generalizable end-to-end deep learning frameworks for real-time
  attitude estimation using 6dof inertial measurement units.
\newblock {\em Available at SSRN 4387238}, 2023.

\bibitem{lee2012factorized}
Guo~Xiong Lee and Kay-Soon Low.
\newblock A factorized quaternion approach to determine the arm motions using
  triaxial accelerometers with anatomical and sensor constraints.
\newblock {\em IEEE Transactions on Instrumentation and Measurement},
  61(6):1793--1802, 2012.

\bibitem{fauske2007estimation}
Kjell~Magne Fauske, Fredrik Gustafsson, and Oyvind Hegrenaes.
\newblock Estimation of auv dynamics for sensor fusion.
\newblock In {\em 2007 10th International Conference on Information Fusion},
  pages 1--6. IEEE, 2007.

\bibitem{hoang2022yaw}
Minh~Long Hoang and Antonio Pietrosanto.
\newblock Yaw/heading optimization by machine learning model based on mems
  magnetometer under harsh conditions.
\newblock {\em Measurement}, 193:111013, 2022.

\bibitem{zhao2022attitude}
Donghua Zhao, Yueze Liu, Xindong Wu, Hao Dong, Chenguang Wang, Jun Tang, Chong
  Shen, and Jun Liu.
\newblock Attitude-induced error modeling and compensation with gru networks
  for the polarization compass during uav orientation.
\newblock {\em Measurement}, 190:110734, 2022.

\bibitem{wang2020recent}
Xizhao Wang, Yanxia Zhao, and Farhad Pourpanah.
\newblock Recent advances in deep learning.
\newblock {\em International Journal of Machine Learning and Cybernetics},
  11(4):747--750, 2020.

\bibitem{xiao2018opportunities}
Cao Xiao, Edward Choi, and Jimeng Sun.
\newblock Opportunities and challenges in developing deep learning models using
  electronic health records data: a systematic review.
\newblock {\em Journal of the American Medical Informatics Association},
  25(10):1419--1428, 2018.

\bibitem{zulqarnain2020comparative}
Muhammad Zulqarnain, Rozaida Ghazali, Yana Mazwin~Mohmad Hassim, and Muhammad
  Rehan.
\newblock A comparative review on deep learning models for text classification.
\newblock {\em Indones. J. Electr. Eng. Comput. Sci}, 19(1):325--335, 2020.

\bibitem{nevavuori2020crop}
Petteri Nevavuori, Nathaniel Narra, Petri Linna, and Tarmo Lipping.
\newblock Crop yield prediction using multitemporal uav data and
  spatio-temporal deep learning models.
\newblock {\em Remote Sensing}, 12(23):4000, 2020.

\bibitem{bouktif2019single}
Salah Bouktif, Ali Fiaz, Ali Ouni, and Mohamed~Adel Serhani.
\newblock Single and multi-sequence deep learning models for short and medium
  term electric load forecasting.
\newblock {\em Energies}, 12(1):149, 2019.

\bibitem{weber2021riann}
Daniel Weber, Clemens G{\"u}hmann, and Thomas Seel.
\newblock Riann—a robust neural network outperforms attitude estimation
  filters.
\newblock {\em AI}, 2:444--463, 2021.

\bibitem{narkhede2021incremental}
Parag Narkhede, Rahee Walambe, Shashi Poddar, and Ketan Kotecha.
\newblock Incremental learning of lstm framework for sensor fusion in attitude
  estimation.
\newblock {\em PeerJ Computer Science}, 7:e662, 2021.

\bibitem{brotchie2022leveraging}
James Brotchie, Wei Shao, Wenchao Li, and Allison Kealy.
\newblock Leveraging self-attention mechanism for attitude estimation in
  smartphones.
\newblock {\em Sensors}, 22(22):9011, 2022.

\bibitem{chen2018oxiod}
Changhao Chen, Peijun Zhao, Chris~Xiaoxuan Lu, Wei Wang, Andrew Markham, and
  Niki Trigoni.
\newblock Oxiod: The dataset for deep inertial odometry.
\newblock {\em arXiv preprint arXiv:1809.07491}, 2018.

\bibitem{aqel2016review}
Mohammad~OA Aqel, Mohammad~H Marhaban, M~Iqbal Saripan, and Napsiah~Bt Ismail.
\newblock Review of visual odometry: types, approaches, challenges, and
  applications.
\newblock {\em SpringerPlus}, 5:1--26, 2016.

\bibitem{nister2004visual}
David Nist{\'e}r, Oleg Naroditsky, and James Bergen.
\newblock Visual odometry.
\newblock In {\em Proceedings of the 2004 IEEE Computer Society Conference on
  Computer Vision and Pattern Recognition, 2004. CVPR 2004.}, volume~1, pages
  I--I. Ieee, 2004.

\bibitem{solin2018inertial}
Arno Solin, Santiago Cortes, Esa Rahtu, and Juho Kannala.
\newblock Inertial odometry on handheld smartphones.
\newblock In {\em 2018 21st International Conference on Information Fusion
  (FUSION)}, pages 1--5. IEEE, 2018.

\bibitem{brossard2019learning}
Martin Brossard and Silvere Bonnabel.
\newblock Learning wheel odometry and imu errors for localization.
\newblock In {\em 2019 International Conference on Robotics and Automation
  (ICRA)}, pages 291--297. IEEE, 2019.

\bibitem{muller2017flowdometry}
Peter Muller and Andreas Savakis.
\newblock Flowdometry: An optical flow and deep learning based approach to
  visual odometry.
\newblock In {\em 2017 IEEE Winter Conference on Applications of Computer
  Vision (WACV)}, pages 624--631. IEEE, 2017.

\bibitem{ohno2004differential}
Kazunori Ohno, Takashi Tsubouchi, Bunji Shigematsu, and Shin'ichi Yuta.
\newblock Differential gps and odometry-based outdoor navigation of a mobile
  robot.
\newblock {\em Advanced Robotics}, 18(6):611--635, 2004.

\bibitem{chen2020deep}
Changhao Chen, Peijun Zhao, Chris~Xiaoxuan Lu, Wei Wang, Andrew Markham, and
  Niki Trigoni.
\newblock Deep-learning-based pedestrian inertial navigation: Methods, data
  set, and on-device inference.
\newblock {\em IEEE Internet of Things Journal}, 7(5):4431--4441, 2020.

\bibitem{han2019deepvio}
Liming Han, Yimin Lin, Guoguang Du, and Shiguo Lian.
\newblock Deepvio: Self-supervised deep learning of monocular visual inertial
  odometry using 3d geometric constraints.
\newblock In {\em 2019 IEEE/RSJ International Conference on Intelligent Robots
  and Systems (IROS)}, pages 6906--6913. IEEE, 2019.

\bibitem{wang2017deepvo}
Sen Wang, Ronald Clark, Hongkai Wen, and Niki Trigoni.
\newblock Deepvo: Towards end-to-end visual odometry with deep recurrent
  convolutional neural networks.
\newblock In {\em 2017 IEEE international conference on robotics and automation
  (ICRA)}, pages 2043--2050. IEEE, 2017.

\bibitem{konda2015learning}
Kishore~Reddy Konda and Roland Memisevic.
\newblock Learning visual odometry with a convolutional network.
\newblock {\em VISAPP (1)}, 2015:486--490, 2015.

\bibitem{geiger2013vision}
Andreas Geiger, Philip Lenz, Christoph Stiller, and Raquel Urtasun.
\newblock Vision meets robotics: The kitti dataset.
\newblock {\em The International Journal of Robotics Research},
  32(11):1231--1237, 2013.

\bibitem{jiao2019magicvo}
Jichao Jiao, Jian Jiao, Yaokai Mo, Weilun Liu, and Zhongliang Deng.
\newblock Magicvo: An end-to-end hybrid cnn and bi-lstm method for monocular
  visual odometry.
\newblock {\em IEEE Access}, 7:94118--94127, 2019.

\bibitem{li2018undeepvo}
Ruihao Li, Sen Wang, Zhiqiang Long, and Dongbing Gu.
\newblock Undeepvo: Monocular visual odometry through unsupervised deep
  learning.
\newblock In {\em 2018 IEEE international conference on robotics and automation
  (ICRA)}, pages 7286--7291. IEEE, 2018.

\bibitem{simonyan2014very}
Karen Simonyan and Andrew Zisserman.
\newblock Very deep convolutional networks for large-scale image recognition.
\newblock {\em arXiv preprint arXiv:1409.1556}, 2014.

\bibitem{almalioglu2019ganvo}
Yasin Almalioglu, Muhamad Risqi~U Saputra, Pedro~PB De~Gusmao, Andrew Markham,
  and Niki Trigoni.
\newblock Ganvo: Unsupervised deep monocular visual odometry and depth
  estimation with generative adversarial networks.
\newblock In {\em 2019 International conference on robotics and automation
  (ICRA)}, pages 5474--5480. IEEE, 2019.

\bibitem{wang2018end}
Sen Wang, Ronald Clark, Hongkai Wen, and Niki Trigoni.
\newblock End-to-end, sequence-to-sequence probabilistic visual odometry
  through deep neural networks.
\newblock {\em The International Journal of Robotics Research},
  37(4-5):513--542, 2018.

\bibitem{clark2017vinet}
Ronald Clark, Sen Wang, Hongkai Wen, Andrew Markham, and Niki Trigoni.
\newblock Vinet: Visual-inertial odometry as a sequence-to-sequence learning
  problem.
\newblock In {\em Proceedings of the AAAI Conference on Artificial
  Intelligence}, volume~31, pages~,, 2017.

\bibitem{aslan2022hvionet}
Muhammet~Fatih Aslan, Akif Durdu, Abdullah Yusefi, and Alper Yilmaz.
\newblock Hvionet: A deep learning based hybrid visual--inertial odometry
  approach for unmanned aerial system position estimation.
\newblock {\em Neural Networks}, 155:461--474, 2022.

\bibitem{yang2022efficient}
Mingyu Yang, Yu~Chen, and Hun-Seok Kim.
\newblock Efficient deep visual and inertial odometry with adaptive visual
  modality selection.
\newblock In {\em Computer Vision--ECCV 2022: 17th European Conference, Tel
  Aviv, Israel, October 23--27, 2022, Proceedings, Part XXXVIII}, pages
  233--250. Springer, 2022.

\bibitem{george2023visual}
Anand George, Niko Koivum{\"a}ki, Teemu Hakala, Juha Suomalainen, and Eija
  Honkavaara.
\newblock Visual-inertial odometry using high flying altitude drone datasets.
\newblock {\em Drones}, 7(1):36, 2023.

\bibitem{fgi}
Anand George, Niko Koivumäki, Teemu Hakala, Juha Suomalainen, and Eija
  Honkavaara.
\newblock Fgi masala stereo-visual-inertial dataset 2021.
\newblock \url{https://doi.org/10.23729/f4c5d2d0-eddb-40a1-89ce-601c252dab35},
  11 2022.
\newblock National Land Survey of Finland, FGI Dept. of Remote sensing and
  photogrammetry.

\bibitem{tu2022ema}
Zheming Tu, Changhao Chen, Xianfei Pan, Ruochen Liu, Jiarui Cui, and Jun Mao.
\newblock Ema-vio: Deep visual--inertial odometry with external memory
  attention.
\newblock {\em IEEE Sensors Journal}, 22(21):20877--20885, 2022.

\bibitem{almalioglu2022selfvio}
Yasin Almalioglu, Mehmet Turan, Muhamad Risqi~U Saputra, Pedro~PB
  de~Gusm{\~a}o, Andrew Markham, and Niki Trigoni.
\newblock Selfvio: Self-supervised deep monocular visual--inertial odometry and
  depth estimation.
\newblock {\em Neural Networks}, 150:119--136, 2022.

\bibitem{cortes2018deep}
Santiago Cort{\'e}s, Arno Solin, and Juho Kannala.
\newblock Deep learning based speed estimation for constraining strapdown
  inertial navigation on smartphones.
\newblock In {\em 2018 IEEE 28th International Workshop on Machine Learning for
  Signal Processing (MLSP)}, pages 1--6. IEEE, 2018.

\bibitem{chen2018ionet}
Changhao Chen, Xiaoxuan Lu, Andrew Markham, and Niki Trigoni.
\newblock Ionet: Learning to cure the curse of drift in inertial odometry.
\newblock In {\em Proceedings of the AAAI Conference on Artificial
  Intelligence}, volume~32, pages~,, 2018.

\bibitem{chen2019deep}
Changhao Chen, Chris~Xiaoxuan Lu, Johan Wahlstr{\"o}m, Andrew Markham, and Niki
  Trigoni.
\newblock Deep neural network based inertial odometry using low-cost inertial
  measurement units.
\newblock {\em IEEE Transactions on Mobile Computing}, 20(4):1351--1364, 2019.

\bibitem{esfahani2019aboldeepio}
Mahdi~Abolfazli Esfahani, Han Wang, Keyu Wu, and Shenghai Yuan.
\newblock Aboldeepio: A novel deep inertial odometry network for autonomous
  vehicles.
\newblock {\em IEEE Transactions on Intelligent Transportation Systems},
  21(5):1941--1950, 2019.

\bibitem{silva2019end}
Jo{\~a}o~Paulo Silva~do Monte~Lima, Hideaki Uchiyama, and Rin-ichiro Taniguchi.
\newblock End-to-end learning framework for imu-based 6-dof odometry.
\newblock {\em Sensors}, 19(17):3777, 2019.

\bibitem{liu2020tlio}
Wenxin Liu, David Caruso, Eddy Ilg, Jing Dong, Anastasios~I Mourikis, Kostas
  Daniilidis, Vijay Kumar, and Jakob Engel.
\newblock Tlio: Tight learned inertial odometry.
\newblock {\em IEEE Robotics and Automation Letters}, 5(4):5653--5660, 2020.

\bibitem{guimaraes2021deep}
V{\^a}nia Guimar{\~a}es, In{\^e}s Sousa, and Miguel~Velhote Correia.
\newblock A deep learning approach for foot trajectory estimation in gait
  analysis using inertial sensors.
\newblock {\em Sensors}, 21(22):7517, 2021.

\bibitem{soyer2021efficient}
M~Serhat Soyer, A~Abdel-Qader, and Mehmet~Cengiz Onba{\c{s}}l{\i}.
\newblock An efficient and low-latency deep inertial odometer for smartphone
  positioning.
\newblock {\em IEEE Sensors Journal}, 21(24):27676--27685, 2021.

\bibitem{uno2022deep}
Takuma Uno, Naoya Isoyama, Hideaki Uchiyama, Nobuchika Sakata, and Kiyoshi
  Kiyokawa.
\newblock Deep inertial underwater odometry system.
\newblock {\em CEUR Workshop Proceedings}, 2022.

\bibitem{chen2022deep}
Boxuan Chen, Ruifeng Zhang, Shaochu Wang, Liqiang Zhang, and Yu~Liu.
\newblock Deep-learning-based inertial odometry for pedestrian tracking using
  attention mechanism and res2net module.
\newblock {\em IEEE Sensors Letters}, 6(11):1--4, 2022.

\bibitem{deraz2022deep}
Ashraf~A Deraz, Osama Badawy, Mostafa~A Elhosseini, Mostafa Mostafa, Hesham~A
  Ali, and Ali~I El-Desouky.
\newblock Deep learning based on lstm model for enhanced visual odometry
  navigation system.
\newblock {\em Ain Shams Engineering Journal}, page 102050, 2022.

\bibitem{wang2022a2dio}
Yingying Wang, Hu~Cheng, and Max Q-H Meng.
\newblock A2dio: Attention-driven deep inertial odometry for pedestrian
  localization based on 6d imu.
\newblock In {\em 2022 International Conference on Robotics and Automation
  (ICRA)}, pages 819--825. IEEE, 2022.

\bibitem{shoushtari2022l5in+}
Hossein Shoushtari, Firas Kassawat, Dorian Harder, Korvin Venzke, J{\"o}rg
  M{\"u}ller-Lietzkow, and Harald Sternberg.
\newblock L5in+: From an analytical platform to optimization of deep inertial
  odometry.
\newblock In {\em 12th International Conference on Indoor Positioning and
  Indoor Navigation-Work-in-Progress Papers (IPIN-WiP 2022), 5-7 September
  2022, Beijing, China.}, pages~,. CEUR-WS, 2022.

\bibitem{cioffi2022learned}
Giovanni Cioffi, Leonard Bauersfeld, Elia Kaufmann, and Davide Scaramuzza.
\newblock Learned inertial odometry for autonomous drone racing.
\newblock {\em arXiv preprint arXiv:2210.15287}, 2022.

\bibitem{zhang2022dido}
Kunyi Zhang, Chenxing Jiang, Jinghang Li, Sheng Yang, Teng Ma, Chao Xu, and Fei
  Gao.
\newblock Dido: Deep inertial quadrotor dynamical odometry.
\newblock {\em IEEE Robotics and Automation Letters}, 7(4):9083--9090, 2022.

\bibitem{buchanan2022learning}
Russell Buchanan, Marco Camurri, Frank Dellaert, and Maurice Fallon.
\newblock Learning inertial odometry for dynamic legged robot state estimation.
\newblock In {\em Conference on Robot Learning}, pages 1575--1584. PMLR, 2022.

\bibitem{wang2022magnetic}
Yan Wang, Jian Kuang, You Li, and Xiaoji Niu.
\newblock Magnetic field-enhanced learning-based inertial odometry for indoor
  pedestrian.
\newblock {\em IEEE Transactions on Instrumentation and Measurement}, 71:1--13,
  2022.

\bibitem{saha2022tinyodom}
Swapnil~Sayan Saha, Sandeep~Singh Sandha, Luis~Antonio Garcia, and Mani
  Srivastava.
\newblock Tinyodom: Hardware-aware efficient neural inertial navigation.
\newblock {\em Proceedings of the ACM on Interactive, Mobile, Wearable and
  Ubiquitous Technologies}, 6(2):1--32, 2022.

\bibitem{visca2022meta}
Marco Visca, Roger Powell, Yang Gao, and Saber Fallah.
\newblock Meta-conv1d energy-aware path planner for mobile robots in
  unstructured terrains.
\newblock 2022.

\end{thebibliography}

\end{document}